%% file: main.tex
\documentclass{article} 
\usepackage{iclr2027_conference,times}

\input{math_commands.tex}

\usepackage{hyperref}
\usepackage{url}
\usepackage{amsmath}
\usepackage{graphicx}
\usepackage{subcaption}
\usepackage{wrapfig}
\usepackage{booktabs}  
\usepackage{array}  
\usepackage{arydshln}  

\usepackage{amssymb}
\usepackage{enumitem}
\usepackage{tikz}
\usetikzlibrary{positioning,arrows.meta,calc}
\usepackage{pgfplots}
\pgfplotsset{compat=1.16}
\usepackage[most]{tcolorbox}

\title{Strategically Diverse Sampling for \\ Self-Training}

\iclrfinalcopy

\author{Alexander Gurung, Esmeralda S. Whitammer$^\dagger$ \ \& Mirella Lapata \\
School of Informatics, University of Edinburgh \\
\texttt{\{alex.gurung,esmeralda.whitammer\}@ed.ac.uk}, \texttt{mlap@inf.ed.ac.uk}}

\begin{document}

\maketitle
\lhead{Preprint. Under review.}
\begingroup
\renewcommand\thefootnote{$\dagger$}%
\footnotetext{CIFAR Fellow.}
\endgroup

\begin{abstract}
Many LLM training and inference methods, including  RL and test-time scaling,  
depend on repeated sampling,
but benefit only when the responses meaningfully differ. Self-training faces the same challenge: training data is typically constructed by 
sampling IID responses and filtering primarily for correctness,
thereby over-representing strategies a model already favours. We
investigate \emph{strategic diversity}, or substantive variation among
approaches to a problem, as an alternative principle for constructing
self-training data. We generate strategically diverse data with two sampling methods: \textsc{Groot}, a new method which
constructs a hierarchical tree of approaches and samples distinct
paths, and Verbalized Sampling (\textsc{VS}), adapted to produce an
unstructured set of approaches. Across competitive programming and
Next-Chapter Prediction domains, models trained on strategically sampled data
outperform IID-trained counterparts on difficult tasks and 
provide strong initializations for RL and test-time scaling. Most
strikingly, self-training on strategically diverse but incorrect traces from Qwen3-4B outperforms IID
distillation from a 235B teacher. These results challenge prevailing assumptions
about what makes useful self-training data and show that diversity of approaches can matter more than correctness or teacher scale. 
\end{abstract}

\section{Introduction}

Many LLM training and inference methods rely on sampling multiple
responses to the same input. Common examples include sampling many candidates
under a verifier \citep{brown2024repeated}, estimating
policy gradients from rollout groups with reinforcement learning
(RL) methods like GRPO \citep{shao_deepseekmath_2024}, and aggregating
populations of candidate answers
\citep{venkatraman2026recursive}. Self-training similarly samples
outputs from a model and fine-tunes on those that pass a correctness
filter, most commonly through Rejection Sampling Fine-Tuning
(RFT). 
These methods all benefit when repeated samples are meaningfully different, a property termed functional diversity \citep{jain2026homogenization}.

However, responses sampled independently from an LLM are often 
surface variations of the same underlying strategy, reflecting the
model's tendency to concentrate probability on a single approach. 
Consequently, the value of additional samples
diminishes rapidly at larger sampling budgets
\citep{yue2025reinforcement}.  In RL, this causes rollout groups to
earn identical rewards and provide no gradient; in test-time scaling,
aggregators see near-identical populations with little to
combine \citep{venkatraman2026recursive}. These failures are particularly acute near a model's
capability boundary, where success is rare or never observed at
feasible sampling budgets \citep{noukhovitch2026learningsolvehardproblems}. 

This mode-collapsed behavior is encouraged during post-training
\citep{karouzos2026collapse,zhang2026verbalized}, and standard
self-training pipelines may further concentrate the distribution \citep{li2025preserving}. Task-specific training data is typically sampled independently from the
student or a larger teacher model, and filtered primarily for
correctness.  We argue that performance can be improved by
changing the sampling procedure  to encourage \emph{strategic
diversity}: substantive variation in the approaches taken to solve a
task, so that the search space of reasonable decisions is meaningfully
covered. This differs from lexical or embedding-based diversity,
which emphasizes surface form, and is orthogonal to correctness.  Our
intuition is that IID self-training reinforces
strategies the model already prefers, strategically sampled data preserves diversity in the fine-tuned model, giving repeated sampling, RL, and test-time scaling a more useful distribution of samples.

We instantiate this idea through two methods for generating
strategically diverse self-training data. We apply \textsc{Verbalized Sampling} (VS;
\citealt{zhang2026verbalized}) to predict several
approaches to a given question together with their estimated probabilities. \textsc{Groot}
(named for its tree structure) instead asks the model to construct a
decision tree over possible approaches and selects $n$~distinct
root-to-leaf paths. Each approach is passed to the
model as a hidden instruction for generating a complete response
(\autoref{fig:pipeline}). We finetune models on this data and evaluate via test pass@$k$, test-time scaling and RL. We compare these strategically-diverse sampling methods against IID
sampling from both the student and a much larger teacher, at standard
and high temperatures and across substantially larger sampling
budgets.

\input{figures/fig_pipeline}

We evaluate in two domains that differ along two important dimensions:
the structure of their approach spaces and the availability of a
verifier. \textbf{Competitive programming} requires hierarchical decisions over
algorithms, data structures, and implementation choices, giving its
approach space a natural tree structure. We train on Cobalt
\citep{chen2026bridging,li2023taco} and evaluate on Cobalt,
LiveCodeBench \citep{jain2024livecodebench}, and OJBench
\citep{wang2026ojbench}. We focus particularly on frontier problems,
operationally defined as problems solved in at most two of 64 IID
attempts. In contrast, \textbf{Next-Chapter Prediction} (NCP;
\citealt{gurung_learning_2025}) requires planning  the next chapter of
a story. Its approach space consists of many interdependent narrative
decisions without an obvious hierarchy, and quality is measured
continuously rather than through binary correctness. 

Across both domains, models trained on strategically sampled data
consistently outperform IID-based training. On frontier code problems,
\textsc{Groot}- and VS-trained models more than triple the base
model's held-out pass@8, whereas IID training provides little
improvement even with 16 times more samples. On NCP, training on
IID-sampled high-quality plans doesn't improve the model's base capabilities, while strategically diverse training
significantly increases the likelihood of high-quality plans. Strategically trained models also provide stronger
initializations for RL and test-time aggregation. 

Strikingly,
training on strategically diverse but incorrect responses outperforms
training on correct IID responses, and strategically diverse samples
generated by the student outperform IID samples distilled from a 235B
teacher. These results show that the composition of the sampled
strategy set can matter more than correctness, sampling budget, or
teacher scale. 

Our contributions are the following: (1) We identify \emph{strategic diversity} as an important dimension of self-training data and develop methods for eliciting it directly from the student model: \textsc{Groot}, a new hierarchical sampling method, and an approach-level adaptation of Verbalized Sampling. (2) We show that strategically diverse data improves performance under repeated sampling and provides a stronger initialization for downstream RL and test-time aggregation, in competitive programming and story-planning. This holds true even when IID sampling has a larger budget, finds more correct answers, or uses a stronger teacher model.


%
%

\section{Related Work}

\textbf{Data Generation for Reasoning Post-Training.} Many reasoning-focused post-training methods build corpora by sampling solutions from a model and retaining those that are correct. STaR \citep{zelikman2022star} iteratively finetunes on successful self-generated rationales, while \citet{yuan2023scaling} formalize IID sampling followed by training on correct solutions as Rejection Sampling Fine-Tuning (RFT). RFT is now widely used \citep{yang2025qwen3,kimiteam2026k2,glm2026glm5}, including as a warmup before RL \citep{deepseekai2026r1,rrv2026midtraining}. When a stronger model is
available, the same sample-and-filter procedure can instead be used to
distill from that `teacher'
\citep{he2025skywork,mistralai2025magistral,liu2025acereason,5team2025glm}. However, small students do not always benefit from strong-teacher traces \citep{li2025small}, suggesting that training-corpus utility depends on more than correctness.

\citet{yuan2023scaling} show that distinct reasoning paths matter beyond correctness, while \citet{guha2025openthoughts} find that sampling more solutions per problem can improve performance as much as adding new problems. \citet{rrv2026midtraining} condition generations on human-written heuristics and select candidates with a reward model. HDPO \citep{cao2026hdpo} generates structured hint trajectories with a 235B teacher and filters them with a reliability model, while prefix-conditioned SFT \citep{fan-etal-2026-learning} prepends random tokens so that different responses to the same problem are not trained under identical inputs. In creative writing, DPWriter \citep{cao2026dpwriter} branches RL rollouts at planning steps and scores continuations using a learned quality reward and n-gram diversity bonus. Structure has also been imposed at inference time: Tree-of-Thoughts \citep{yao2023tot} searches over intermediate reasoning states, evaluating and pruning branches to produce a better answer.

In contrast, our work generates training data from the student itself, focuses on the student's own conception of the problem approach-space, and tries to sample distinct approaches to difficult tasks.



\textbf{Diversity Collapse in Post-Training.} Despite improving performance, post-training can concentrate LLM responses on a narrow set of approaches. \citet{yue2025reinforcement} show that RL with verifiable rewards increases pass@1 but decreases pass@$k$, with the solvable set remaining largely within the base model's capabilities, though prolonged training can partially expand this boundary \citep{liu2025prorl}. \citet{cui2025entropy} attribute this to entropy collapse induced by the RL objective, while \citet{karouzos2026collapse} trace it to post-training stages (SFT or preference tuning depending on the pipeline) and show that decoding changes alone cannot reverse it.
This diversity collapse  harms downstream methods. Group-based RL requires rollouts with differing rewards; when solutions follow the same strategy, reward variance and the learning signal diminish. Test-time scaling likewise depends on diverse populations, whether selecting responses with a verifier \citep{brown2024repeated}, majority voting over reasoning chains \citep{wang2023selfconsistency}, or recursively combining solutions \citep{venkatraman2026recursive}. We use the latter, Recursive Self-Aggregation (RSA), in  experiments.

\textbf{Preserving and Measuring Diversity.} Most work counteracts diversity collapse by modifying the training objective, either in RL \citep{cui2025entropy,li2026divergence,chen2025passk,li2026setpo,lochab2026uniform,gai2025differential,peng2026sampled,li2025darling} or in SFT through entropy-preserving or token-reweighting objectives \citep{li2025preserving,klypa2026diversity,wang2026gradients,li2026loglikelihood,wu2026generalization}. Other work intervenes at inference time: Verbalized Sampling \citep{zhang2026verbalized} prompts a model to produce distinct responses with associated probabilities, Diversified Sampling perturbs best-of-$n$ prompts using sampled roles or generated ideas \citep{wang2025divsampling}, and DIVE \citep{xiong2026dive} uses verifier feedback to evolve populations of natural-language skills for a frozen model. We instead aim to increase diversity through the training \textit{data} during the warmup stage, keeping the training objective and test-time inference fixed.

Diversity is commonly measured using surface statistics such as distinct n-grams \citep{li2016diversity}, response-embedding similarity \citep{friedman2023vendi,chen2026posttraining,li2026synthesizing}, or semantic equivalence among valid outputs \citep{shypula2026evaluating}. Closest to our notion, \citet{lee2026strategy} define approach-level diversity as variation in strategy across correct solutions, assessed by a calibrated LLM judge. 
In contrast, strategic diversity does not require correct responses and aims to reflect the model's own conception of reasonable decision points. We measure the effect of diverse training data through pass@$k$ on difficult problems, downstream RL, and test-time scaling.

%
%

\section{Strategically Diverse Sampling Methods}

We seek a sampling procedure that returns $n$ responses meaningfully differing in \emph{approach}. 
We would also like this procedure to only require the model being trained, add little cost to training-data generation, and require no additional training. 
Standard IID sampling does not reliably provide approach-level diversity: samples from the same conditional distribution concentrate on a small number of strategies, while higher temperature perturbs tokens without necessarily changing the underlying approach. We instead want the model to commit to a different approach \emph{before} generating a solution. We consider two such methods that differ in how they represent the
space of possible approaches and encourage separation
among responses: one elicits an unstructured set of
alternatives, whereas the other organizes the space hierarchically and
selects distinct decision paths.

\textbf{\textsc{Verbalized Sampling (VS)}} Proposed by
\citet{zhang2026verbalized}, \textsc{VS} prompts an LLM for a list of
possible answers paired with their hypothetical probability (see
Figure~\ref{fig:pipeline}, bottom). They find that this simple,
training-free method produces substantially more diverse responses
than direct prompting while maintaining their quality. In contrast to
the original work which applied \textsc{VS} to full answers, we apply
\textsc{VS} to predict approaches which are extended into full answers
in a \emph{separate} call. Separation is whatever the model's own
enumeration supplies;  the reported probabilities are estimates rather
than constraints, and nothing requires two listed approaches to differ
in any particular respect.

\textbf{\textsc{Groot}} We also introduce \textsc{Groot}, as a more
structured way of sampling diverse responses.  \textsc{Groot} asks the
model to build a reasoning tree of the ways the problem could be
approached, and then to pick $n$ distinct paths over this tree. The
model constructs an approach for each of these paths, yielding
\textit{strategically diverse} responses (Figure~\ref{fig:pipeline}, middle).
Diversity here is enforced by the fact that two root-to-leaf paths
diverge at some point, resulting in approaches that differ at a 
decision the model marks as important. 
Note that the tree in \textsc{Groot} serves a different purpose
than in tree-structured inference methods such as Tree-of-Thoughts
\citep{yao2023tot}: we only create this tree once during dataset creation, with standard unstructured generation at test-time.

Our two strategic-sampling instantiations,
\textbf{\textsc{VS}} and \textbf{\textsc{Groot}}, differ only in how much
explicit structure they impose on the approach space. We hypothesize that which method is preferable
depends on the domain and downstream objectives.
Competitive programming may involve naturally hierarchical
decisions, such as the choice of algorithm constraining the
relevant data structures, whereas in next-chapter prediction
the search space is broad and less obviously structured. Imposing structure might also limit the breadth of exploration, while unstructured search may miss critical decision points. We compare methods empirically in Section~\ref{sec:results}.

Both \textsc{VS} and \textsc{Groot} elicit multiple approaches in a
single planning call, but the resulting training examples must
retain the standard reasoning-then-answer format. We therefore
separate approach elicitation from approach execution. First the model
produces $n$ distinct approaches to the problem, represented as either
a probabilistic list or distinct tree paths. Each approach is then
passed to the model in a separate prompt alongside the original
question, framed as a hidden instruction. The model is instructed to
reason as if it arrived at the approach on its own, yielding valid
reasoning traces to the original question. We automatically discard
responses that refer explicitly to the approach. This two stage approach 
allows us to focus on improving diversity in the semantic planning space. 
All prompts are provided in
Appendix~\ref{app:prompts}, and Appendix~\ref{app:examples} shows
example outputs.

One limitation of these methods is that they  both add one generation call per 
problem (or per list of plans generated). In order to compare the effectiveness of the resulting datasets at comparable sizes, we do not include this additional cost in the budgets described later, but we do compare our diverse datasets against IID baselines generated with sampling budgets up to $16\times$ larger.

\section{Experimental Setup}
\label{sec:setup}

We run all experiments with Qwen3-4B-Instruct \citep{yang2025qwen3},
and replicate primary results on Nemotron3 Nano-4B
\citep{nvidia2025nemotron3}. All data generation, training, and
evaluation use the same model within an experiment (e.g., we train the
Qwen model on Qwen-generated data, and vice-versa for Nemotron). We
primarily report Qwen3-4B-Instruct in the main text, but ablations
with Nemotron3-Nano-4B show the same trends
(Appendix~\ref{app:n3n}).

\subsection{Tasks and Benchmarks}

\textbf{Competitive Programming} A model is given a problem statement
and must produce a program that passes a held-out test suite, so
correctness is binary and automatically verifiable. We train on Cobalt
\citep{chen2026bridging,li2023taco} and evaluate on its test set as
well as on the held-out LiveCodeBench and OJBench benchmarks
\citep{jain2024livecodebench,wang2026ojbench}. 

As we expect diverse sampling to be most helpful on difficult problems where the correct approach is not already obvious to the model, we filter these datasets for model-specific \textit{frontier} problems that are difficult for a given model. We create \textit{frontier} datasets by sampling 64 solutions IID and keep problems solved $\leq 2$ times. We use only this filtered dataset for training, but report performance on both the full and frontier sets (as well as dataset-defined difficulty sets).

\textbf{Next-Chapter Prediction (NCP)} \citet{gurung_learning_2025}
ask a model to plan the next chapter of a book,  given plot and
character information established so far, organized in a Story
Information document. We adapt the task to plan $\sim$200-word chunks
instead of the whole chapter, hypothesizing that a narrow focus will
encourage more specific plans. Plans are evaluated by how much they
reduce the perplexity of the true section, expressed as a percentage
of the no-plan perplexity, following the reward formulation of
\citet{gurung_learning_2025}. As this task does not have binary
`correctness', we use percent-improvement thresholds as a means of
acquiring binary scores. For example, we report the number of unique
plans that achieve $\geq 15\%$ perplexity improvement (see next
section). 

\subsection{Training Data and Fine-Tuning}

\paragraph{Training Data} We collect datasets with the two strategic-sampling methods (\textsc{Groot} and \textsc{VS}), and with standard \textsc{IID}. 

For code, we use sample-budgets of 4 and 8 (denoted by $-4$ and $-8$),  plus a high-budget IID variant, IID-64. This allows us to control for the number of correct samples when comparing strategic  
\input{tables/mining_v2_table}
 methods with IID, as with matched budgets \textsc{Groot} and \textsc{VS} consistently produce more correct datapoints.

For NCP, we report results with budgets 8 and 16 as high-quality plans ($\geq 15\%$) are very rare. 
Table~\ref{tab:mining} reports the solved problem and threshold-clearing counts. 

We also compare against naively increasing temperature to increase diversity. We sweep temperatures across sampling methods (Appendix~\ref{app:temperature}) and find $T=1.5$ consistently increases the number of unique problems solved without causing severe degradation in answer quality.


In addition to self-training, a standard approach for dataset creation is to use a stronger model (when within budget constraints). We repeat IID, \textsc{VS}, and \textsc{Groot} data generation with Qwen3-235B-A22B-Instruct, which shows better
performance on these tasks while having a similar reasoning style, a
compatibility that prior work finds important for distillation
\citep{li2026rethinking}.

\paragraph{Fine-tuning} We train our base model with SFT and RFT on these datasets, with the latter filtering out incorrect
answers. To estimate whether improvements come from diversity or
simply from sampling more correct answers, we also test an `ANTI'
setting where we filter out all \textit{correct} answers.  We
use the same hyperparameters (Appendix~\ref{app:hparams}) for
every dataset, and select the best checkpoint by validation loss on a dataset created and filtered with the same corresponding pipeline as for training (e.g. IID sampled with budget 4 and RFT filtered). To reduce training cost on NCP, we match SFT and ANTI dataset sizes to RFT, selecting a random $d$ datapoints and the worst $d$ respectively. We report means over three seeds.


\subsection{Evaluating Trained Models}

We evaluate the models trained on these datasets in three ways (Figure~\ref{fig:pipeline}, right): direct test-time performance, test-time aggregation, and as initialization for RL. At test time all models receive the original prompts, so differences result only from training's effect on their output distributions.

\textbf{Test-time performance:} For code we evaluate performance with
pass@$k$. To construct a comparable binary metric for NCP we use
coverage: the fraction of test sections for which at least one of $k$
sampled plans clears a perplexity-improvement threshold. Scores are estimated from 128 samples.

\textbf{Test-time scaling:} We hypothesize more diverse training
datasets will encourage the model to produce more useful populations
for test-time scaling. We run Recursive Self-Aggregation (RSA;
\citealt{venkatraman2026recursive}) on top of each trained model: an
initial population of 16 candidate solutions is refined through an
evolutionary process over ten iterations, where each child is
generated by aggregating four members of the previous population. The
trained model is used for all stages. We report pass@1, averaged over the three aggregation runs.

\textbf{Reinforcement learning:} High asymptotic pass@$k$ may accelerate RL by reducing zero-advantage groups. We initialize RL from each trained model, and use the MaxRL advantage estimator \citep{tajwar2026maxrl} for code (as it targets higher pass@$k$) and GRPO estimator for NCP. MaxRL, which requires a binary reward, divides the binary reward by the sampled group's mean reward, and is reported to improve pass@$k$ at comparable pass@$1$. We hypothesize that this objective will better maintain diversity in the models used for initialization. We select the best performing checkpoint by pass@8 on the validation set, and report test-set pass@$k$.

%
%

\input{tables/code_passk_table}

\section{Results}
\label{sec:results}

\label{sec:results-mining}
\textbf{Strategically-diverse sampling solves more distinct problems.}
Although the goal of our strategic-diversity sampling methods is not
simply to maximize correctness, we find that they do solve significantly more
unique problems than IID sampling with a matched budget. 
On the frontier Cobalt train set with a four sample budget,
\textsc{Groot} solves 155 problems against 42 for IID sampling and 60 for IID sampling
at $T=1.5$ (Table~\ref{tab:mining}). 
Based on the high-budget high-temperature IID-64 variant, we estimate that IID sampling would need about $\frac{64 \times 155}{379}\approx 26$ samples to match \textsc{Groot-$4$}. NCP
shows a similar pattern: with a budget of 16 samples and a
high-quality threshold of 15\% improvement, \textsc{Groot} found high-quality plans for 286 sections. In
contrast, IID sampling only found 156. Appendix~\ref{app:ncp} gives
dataset and scoring details, as well as full coverage curves.


\label{sec:results-passk}
\input{figures/fig_ncp_rl_pair}

\textbf{Training on strategically diverse data produces better
  asymptotic performance.} Table~\ref{tab:code_passk} reports pass@$k$
for Qwen3-4B-Instruct trained with RFT on each dataset, on the
frontier subsets of the in-distribution test set (Cobalt) and on the held-out benchmarks. At high-$k$, both \textsc{Groot} and \textsc{VS} dramatically outperform the IID-based approaches. For example, \textsc{Groot}-4 and \textsc{VS}-4 reach a
pass@64 of 15.5 and 17.5 on the held-out benchmarks, compared to only 5.2 for the budget matched IID-4. Even at higher budgets IID remains limited, with IID-64 achieving only~6.5.
Appendix~\ref{app:native}
reports results on the full test sets, as well as split by the benchmarks'
own easy/medium/hard difficulty tiers rather than by our
model-relative frontier definition. We find that gains are concentrated on the medium and hard classes of problems, although in the most difficult classes performance remains stable regardless of training. 

Contrary to
\citet{guha2025openthoughts} who find that more responses per question
is an important scaling axis, our results show that scaling the
training data further has little downstream effect for both IID and strategic sampling methods (Table~\ref{tab:code_passk_all},
Appendix~\ref{app:ablations}). One potential explanation is that the
model's ability to identify and execute reasonable diverse responses plateaus early.

The gap is larger on NCP, where training on \textsc{Groot} and \textsc{VS}
data substantially improves the base model while IID data has little
effect.  Training on plans with $\geq 20\%$ improvement and
evaluating at the 15\% bar, \textsc{VS} reaches
coverage@32 of 4.08\%, compared to 1.97\% for the base model and
2.08\% for IID.  The ordering is unchanged at the 10\% and 5\%
bars and at every~$k$ (Figure~\ref{fig:ncp_curves_primary}), and in all settings the IID-setting fails to significantly improve from the base model. Coverage at all three thresholds is reported in
Appendix~\ref{app:ncp}, where we also find that increasing the sampling budget from 8 to 16 improves \textsc{VS} but leaves \textsc{Groot} and IID unchanged.

\label{sec:results-hightemp}
\textbf{Temperature-based diversity is less effective than strategic diversity.} Raising the temperature during sampling improves IID-based models, but they still perform worse than strategic sampling-based models. On frontier code, IID-64 at $T{=}1.5$ remains substantially below \textsc{Groot}-$4$ and \mbox{\textsc{VS}-$4$}, despite using $16\times$ the sampling budget (Table~\ref{tab:code_passk}). NCP shows the same: high-temperature IID improves over standard IID but remains below \textsc{Groot} and \textsc{VS} at every threshold (Table~\ref{tab:ncp_t15}). Higher temperature can also improve strategic sampling itself, suggesting that temperature and approach-level diversification are complementary rather than interchangeable (Appendix~\ref{app:temperature}).


\label{sec:results-filters}
\textbf{Gains are driven by diversity, not correctness.} 
As described previously, \textsc{Groot} and \textsc{VS} produce training sets that are both more approach-diverse
\textit{and} contain more correct answers than IID sampling
(Table~\ref{tab:mining}). To isolate the effect of training on more diverse approaches from the effect of training on more correct ones, we train in our ANTI setting that filters for only \textit{incorrect} responses.

As Table~\ref{tab:code_passk} shows, \textbf{training on incorrect but strategically diverse responses outperforms RFT on IID-based samples}. This is true even when IID has a larger sampling budget, and even when IID uses the tuned temperature $T=1.5$. ANTI \textsc{Groot}-4 and \textsc{VS}-4 achieve held-out frontier pass@64 of 12.8 and 14.8 respectively, versus 7.9 for IID-64 at $T=1.5$. Correctness filtering still helps within strategically diverse data: RFT \textsc{Groot}-4 reaches 15.5 versus 12.8 for ANTI, and RFT \textsc{VS}-4 reaches 17.5 versus 14.8.  Further filter and budget ablations are in Appendix~\ref{app:ablations}. 

We find similar trends on NCP. Even ANTI, which uses the worst-performing plans, and SFT, which uses randomly selected plans, outperform all variants of IID-sampled data. With a threshold of 10\% improvement, ANTI \textsc{Groot}-16, for example, achieves a coverage@8 of 7.5\%, below the RFT performance of 10.6\% but still above IID-16's 6.80\%. \autoref{tab:ncpstudents} shows coverage at different thresholds and across settings.

\label{sec:results-teacher}
\input{tables/teacher_table}

\textbf{Self-generated strategic diversity beats a 235B teacher.}
The larger Qwen3-235B teacher substantially improves IID training, but still underperforms strategically diverse self-training. The teacher raises IID-4 held-out frontier pass@64 from 5.8 to 13.4, while self-trained \textsc{Groot}-4 and \textsc{VS}-4 reach 20.1 and 22.8 (Table~\ref{tab:teacher}). Combining strategic sampling with the teacher models improves over teacher IID but remains below self-training, possibly due to greater teacher-student distribution mismatch. As a result, it is both more efficient and more performant to use strategic sampling with the student. Teacher-trained models use a 16k generation limit because of longer reasoning traces; standard 8k results appear in Appendix~\ref{app:teacher}. For Nemotron models the self-generated strategic diversity methods still outperform IID sampling with the large teacher, but the strategic diversity methods also benefit significantly from the large teacher (Appendix~\ref{app:n3n}).


\input{tables/rsa_pair}

\textbf{Strategically diverse models are a better basis for test-time
  scaling.} We next test whether trained models benefit more from test-time scaling using Recursive Self-Aggregation (RSA; \citealt{venkatraman2026recursive}) in both domains. For code, Table~\ref{tab:rsa} reports pass@1 before and after aggregation. On the Cobalt frontier, \textsc{Groot} and \textsc{VS} start higher than IID and the base model, and end with an even larger gap. For example, IID-4 improves from 0.9 to 3.8, whereas \textsc{VS}-4 starts at 4.2 and reaches 10.6. Following our other results, high-temperature, high-budget IID-64 performs better (final pass@1 of 7.4) than lower budget IID models but still worse than the much cheaper strategically diverse models. More results and ablations are in Table~\ref{tab:rsa_full}.

Strategic sampling methods also outperform IID and base models when applying RSA to the NCP task, measured with coverage@1 at different thresholds. \autoref{tab:ncp_rsa} shows the IID model performing even worse than the base model, while \textsc{Groot} and \textsc{VS} perform better at each threshold. For example, with a quality threshold of $5\%$,
\textsc{Groot} improves during test-time scaling from 15.7 to 24.6, while IID only improves from 10.7 to 18.1. \textsc{Groot} and \textsc{VS} show a larger \textit{improvement} during RSA than IID, indicating that the performance is not solely due to a better starting population.

\input{tables/postrl_table}

\label{sec:results-rl}
\textbf{Strategically diverse models are a better basis for RL.}
Figure~\ref{fig:rl_performance} shows best-so-far validation pass@8
over RL training from each initialization, averaged over three
seeds. \textsc{Groot}-4 and \textsc{VS}-4 begin at a pass@8 that RL
from the base model fails to reach over six epochs, and continue
improving. In contrast, RL from IID-4 starts below even the base model
and needs 64 steps to reach the strategic models' starting pass@8. RL
from IID-64, the largest IID dataset, starts higher but plateaus and
finishes below both strategic models.

We evaluate the models post-RL with the same pass@$k$ setup as before, and find that strategically diverse models converged to better performance than IID and base models (Table~\ref{tab:postrl}). On the Cobalt frontier,
\textsc{Groot}-4 reaches a pass@64 of 39.0 and \textsc{VS}-4
reaches 37.8, compared to 36.6 for IID-4 and 29.3 for base model. Similarly,
on the held-out frontier benchmarks the average pass@64 is 18.0 and 16.0 for \textsc{Groot}-4 and \textsc{VS}-4 versus
10.9 for IID-4. For the Nemotron experiments, we find that models converged to similar performance regardless of initialization, but strategic sampling methods reached that performance faster (Appendix~\ref{app:n3n}).

Applying RL to NCP, \textsc{Groot} significantly outperforms other methods and similar to our code experiments is followed by \textsc{VS}, IID and the base model. \autoref{tab:ncprl} shows that with a threshold of 10\%, \textsc{Groot} reaches a coverage@8 of
18.8\% while \textsc{VS} and IID reach 5.5\% and 4.8\% respectively. As the reward term encourages \textit{uniquely positive} plans, we also report coverage with additional foil thresholds and find \textsc{Groot} is also significantly better than other initializations (\autoref{tab:ncprl_foils}).

\providecolor{paperBlue}{HTML}{0072B2}
\providecolor{paperOrange}{HTML}{D55E00}
\providecolor{paperGreen}{HTML}{009E73}
\providecolor{paperSlate}{HTML}{6B7280}
\newcommand{\lesson}[1]{\textbf{\textcolor{paperOrange}{#1}}}
\begin{tcolorbox}[colback=paperSlate!7, colframe=paperSlate!75!black,
  colbacktitle=paperSlate!75!black, coltitle=white, halign title=flush center,
  boxrule=0.6pt, arc=2pt, left=2pt, right=5pt, top=3pt, bottom=3pt, boxsep=1pt,
    title=\textbf{Key Takeaways}]
\small
\begin{enumerate}[label=\arabic*., leftmargin=1.6em, itemsep=1pt, topsep=0pt, parsep=0pt]
\item \lesson{Elicit distinct approaches rather than sampling more.} Strategically diverse samples consistently beat IID samples; even with 64 samples and $T{=}1.5$ IID trails \textsc{Groot} with only 4 samples.
\item \lesson{Diversity matters more than correctness.} Even training on the \textit{incorrect} strategic
  samples beats rejection sampling on IID data; the best combination is diverse \textit{and} correct.
\item \lesson{Sample from the student, not from a larger teacher.} Self-generated diversity
  from a 4B model beats IID distillation from a 235B teacher.
\item \lesson{Use \textsc{VS} for pass@$k$, \textsc{Groot} to initialize RL.} On code tasks, \textsc{VS}
  leads before RL (held-out pass@64 of $17.5$ vs.\ $15.5$) and \textsc{Groot} after it
  ($18.0$ vs.\ $16.0$), with a wider gap on NCP.
\end{enumerate}
\end{tcolorbox}
\vspace{-8pt}

%
%

\section{Conclusion}

Our results challenge the prevailing wisdom that correctness is the most important axis for
post-training data. Instead we identify \textit{strategic diversity}, or the diversity of reasonable
approaches according to a given LLM being improved, as a driving force in model improvement at difficult
tasks. We propose a structured (\textsc{Groot}) and an unstructured (using \textsc{VS}) way of acquiring
these diverse responses, and perform experiments on frontier coding and narrative tasks. Models
trained on diverse data consistently outperform those trained on IID-sampled data, even at high
temperature. This trend extends to when a larger teacher model guides improvement: teacher-generated data at larger sampling
budgets still fails to match the small but diverse self-generated sets.

In the future, we would like to investigate whether strategic
diversity can be induced at inference time instead of fine-tuning, and
whether it can be optimized for directly during online RL rather than
inherited from the initialization. 



\bibliography{diversity}
\bibliographystyle{iclr2027_conference}

\appendix \section{Additional experiments and analysis}

\subsection{Second base model: Nemotron3-Nano-4B} \label{app:n3n}

To confirm that our results are not specific to one base model, we repeat the dataset construction,
training, and evaluation with Nemotron3-Nano-4B. We use the same definition of frontier difficulty
as before ($\leq 2$ correct out of 64 samples), but defined by this model's capabilities.

\textbf{Repeated sampling.} Table~\ref{tab:n3n_passk} contains the test-set pass@$k$ and shows the same trends as
Qwen3-4B-Instruct (Table~\ref{tab:code_passk}). At a sampling budget of 4, \textsc{Groot} and
\textsc{VS} more than double the base model's held-out frontier pass@64 of 9.8, reaching 19.9 and
20.8, while the best IID model reaches 14.9 and IID-4 falls slightly below the base model at 8.8.
These students are trained once rather than three times, so this table carries no training-seed
spread.

\input{tables/n3n_passk_table}

\textbf{Reinforcement Learning.} Figure~\ref{fig:rl_n3n} shows the RL curves: as on Qwen, the \textsc{Groot}- and \textsc{VS}-trained
models start far above the base and IID-trained models. On this base model the benefit is primarily
one of speed: \textsc{VS}-4 reaches a validation pass@8 of 40\% after 28 RL steps and
\textsc{Groot}-4 after 48, while IID-4 needs 64, about the same as RL from the untrained base
model, which takes 68. All three trained models then converge to the same validation level. After
RL (Table~\ref{tab:n3n_postrl}) the three finish very close to each other on the held-out
frontier, at a macro pass@64 of 33.6 for \textsc{VS}-4, 33.3 for IID-4 and 31.0 for
\textsc{Groot}-4.

\input{figures/fig_rl_n3n} \input{tables/n3n_postrl_table}

\textbf{Test-time aggregation.} Table~\ref{tab:n3n_rsa} contains the
RSA results. The Nemotron model appears to be a less capable
aggregator in general, often only slightly increasing pass@1, and
the ordering differs by benchmark: IID-8 is best on Cobalt
  and OJBench, on both the full test sets and the frontier subsets,
  while \textsc{VS}-4 is best on LiveCodeBench in both.

\input{tables/n3n_rsa_table}

\textbf{Data creation.} The dataset creation process with
Nemotron3-Nano-4B reproduces the ordering we see for Qwen3-4B
(Table~\ref{tab:n3ncp}). At matched budgets the structured
  methods solve roughly twice as many training-frontier problems as
  IID sampling: 115 for \textsc{Groot} and 137 for \textsc{VS} against
  52 at budget 4, and 197 and 233 against 97 at budget 8. On NCP the
  same ordering holds with smaller margins, since plans are scored by
  a continuous improvement threshold rather than by test execution: at
  the 20\% threshold, 533 and 535 sections against 472 for IID.

\input{tables/n3ncp_table}

\textbf{NCP.} We repeat the full NCP experiment with Nemotron3-Nano-4B as the base model: creating the training
data and training models on plans that clear a $20\%$ improvement threshold.
Table~\ref{tab:n3ncp_students} reports coverage at the $10\%$ threshold at $k=16$: the \textsc{Groot} and \textsc{VS} models reach a coverage of 45.3\% and 43.6\% respectively, compared to the base
model's 37.4\% and IID's 34.6\%.

\input{tables/n3ncp_students_table}

\textbf{Larger teacher.} We repeat the dataset construction with a larger teacher (Nemotron3-Super-120B-A12B-BF16), holding the
generation budget and every other setting fixed so that only the teacher changes
(Table~\ref{tab:n3L}). We find that using the large teacher to generate training data improves performance of each sampling method, but strategic sampling methods are still clearly the best. On the
held-out frontier, \textsc{VS}-4's macro pass@64 increases from 20.8 to 27.4, and \textsc{Groot}-4's from 19.9 to 25.7. IID-4 also increases, from 8.8 to 13.8, but notably remains below both self-training strategic sampling methods.

\input{tables/n3L_table}

\subsection{Sampling budget, data mixtures, and data filters} \label{app:ablations}

Although our main results use a budget of four samples per problem, we investigate the utility of
increasing the sampling budget. We also experiment with combining samples from \textsc{Groot} and
\textsc{VS}, as they often solve distinct sets of problems. Table~\ref{tab:code_passk_all} extends
Table~\ref{tab:code_passk} to budget 8 and to MIX-8, which trains on half \textsc{Groot}-8 and half
\textsc{VS}-8 data. Doubling the budget has little effect: held-out frontier pass@64 changes by less
than two points for every method, and MIX-8 performs comparably to either set individually. The
training sets themselves are small: at budget 4 they hold 220 \textsc{Groot} and 194 \textsc{VS}
examples against 44 for IID, and at budget 8, 419 and 373 against 83. The largest IID set, IID-64,
holds 749 examples, or 1,080 at $T=1.5$.

\input{tables/code_passk_all_table}

Section~\ref{sec:results-filters} compares RFT, SFT, and ANTI for the budget-4 datasets.
Table~\ref{tab:filters_all} gives the same comparison for every Qwen3-4B code dataset across multiple
budgets and data mixtures. The results are the same: on these benchmarks every strategic dataset
beats the corresponding IID-64 dataset with the same filter (e.g., \textsc{VS} and \textsc{Groot} with RFT beat IID-64 RFT). RFT is
the best filter within each strategic method, and SFT and ANTI perform nearly identically. These
results add further evidence that a small amount of strategically diverse data is more useful than
large amounts of correct-but-nondiverse data.

\input{tables/filter_all_table}

\subsection{Sampling temperature} \label{app:temperature}

Section~\ref{sec:results-hightemp} compares strategic diversity against raising the sampling
temperature. Here we describe our temperature sweeps that lead to our hyperparameter choices.

For IID, we swept the sampling temperature over $\{0.1, 0.5, 0.85, 1.0, 1.5, 2.0, 2.5\}$ on the
training frontier and count unique problems solved (mean over three seeds) at a sampling budget
of four (Figure~\ref{fig:temp_grid}, left). Unique problems solved rise from 42 at the standard
$T=0.85$ to 64 at $T=1.5$, plateauing and then collapsing to 34 at $T=2.5$ as outputs degrade.
We use $T=1.5$.

For \textsc{Groot} and \textsc{VS}, we sweep both planner and solver temperatures over $\{0.1, 0.5,
0.85, 1.0, 1.5, 2.0\}$. Figure~\ref{fig:temp_grid} shows unique frontier training problems solved at
a sampling budget of four, mean over three seeds. Raising the planner temperature gives the
largest and most consistent gain: averaged over solver temperatures, \textsc{Groot} goes from 146
problems at $T=0.1$ to 169 at $T=2.0$, and \textsc{VS} from 143 to 162. Raising the solver
temperature helps up to $T=1.0$ and then hurts, with both methods falling to 141 at $T=2.0$. Every
cell of the grid solves more unique problems than IID sampling at any temperature. These
counts average over solver temperatures at a fixed budget, so they are not directly comparable with
the single planner--solver settings of Table~\ref{tab:mining}.
Table~\ref{tab:hightemp} gives the downstream comparison for the IID budgets at both the
default temperature and $T=1.5$, including the budget-8 students that
Table~\ref{tab:code_passk} omits.

The ability to vary the diversity of approaches independently of how they are implemented
is a key benefit of these strategic sampling methods, especially in domains that require precise
answers, such as code. We experiment with a planner temperature of 1.0 or 1.5 and a solver temperature
of 1.0 for the high-temperature \textsc{Groot} and \textsc{VS} students. Raising the planner
temperature slightly improves \textsc{Groot} on all three code benchmarks, by up to
4.0 points of frontier Cobalt pass@64. \textsc{VS} is less stable, gaining slightly on Cobalt and performing worse on
LiveCodeBench (Table~\ref{tab:hightemp}).

\input{tables/hightemp_table} \input{figures/fig_temp}
\input{tables/ncp_t15_trunc_pair}

\subsection{Teacher comparison, all benchmarks} \label{app:teacher}

Due to the high truncation rate from training on traces from Qwen3-235B-A22B-Instruct (Table~\ref{tab:teacher_truncation}),
Section~\ref{sec:results-teacher} compares teacher- and self-generated
data at a 16k-token generation limit.  Tables~\ref{tab:teacher16k}
and~\ref{tab:teacher8k} give the full comparison at both the 16k and
the standard 8k limit. We also include IID-8 rows, since the
  two-call planner-solver pipeline costs roughly 1.25$\times$ as much per
  problem as IID sampling, to ensure that the extra computational
cost of \textsc{Groot} and \textsc{VS} does not explain their
  advantage. Teacher traces help IID substantially (frontier
  Macro-Avg pass@64 5.8 to 13.4 at 16k) but hurt the strategic methods
  (\textsc{Groot} 20.1 to 17.8, \textsc{VS} 22.8 to 17.7), so the
  two sources of improvement do not combine: for \textsc{Groot} and
  \textsc{VS}, self-generated traces are a  better training signal.
  
  However, on Nemotron the larger-teacher has a positive effect on \textsc{Groot} and \textsc{VS} (Appendix~\ref{app:n3n}). The same overall trend still holds, with strategic sampling methods outperforming IID; this teacher-effect inconsistency may be due to the greater reasoning similarity between Nemotron3 4B and its 120B teacher. 
  
Predictably, at an 8k-token limit performance on the teacher rows is
consistently worse than at 16k. The teacher gain concentrates
  on Cobalt, the in-domain benchmark, where teacher-trained IID
  reaches a frontier pass@64 of 36.5 at 8k against 30.5 for
  \textsc{Groot}-4; on the held-out benchmarks strategic diversity
  remains ahead (Macro-Avg 12.4 against 17.5).

\input{tables/teacher_16k_table} \input{tables/teacher_8k_table}

\subsection{RL ablations} \label{app:rl}

Section~\ref{sec:results-rl} reports post-RL performance from the
budget-4 RFT models. We further ablate the sampling budget,
temperature, and correctness filter in our post-training datasets and
investigate their effect on downstream RL. Table~\ref{tab:postrl_all}
gives RL results from every RFT model and
Table~\ref{tab:postrl_sft_anti} from the SFT and ANTI models. Our
results highlight strategic diversity's role as the primary predictor
of downstream performance. After RL, every Qwen3-4B \textsc{Groot},
\textsc{VS}, and MIX model reaches a held-out frontier pass@64 between
15.4 and 18.0, while the strongest IID model we train, IID-8 with high
temperature $T=1.5$, only reaches 12.3. RL from the untrained
  base model reaches 11.4, above IID-4 (10.9), IID-8 (8.3) and IID-4
  at $T=1.5$ (10.2), so IID fine-tuning before RL gains little over no
  fine-tuning at all. RL also does most of its work on the weaker
  initializations: IID-4 roughly doubles (5.2 to 10.9), whereas among
  the strategic models only \textsc{Groot}-4 improves substantially
  (15.5 to 18.0) and \textsc{VS}-4 declines (17.5 to
  16.0). In contrast to \textsc{Groot}, \textsc{VS}-initialized RL sometimes concentrates learning on in-distribution problems, learning
  less generalizable reasoning behaviour. This relationship between warmup data and generalization after RL should be further explored in future work.
  
Although models
trained with high-temperature IID samples start RL ahead of
  the other IID models, by the end of RL the IID models end
  within a few points of each other and still below every strategic
  model. The same holds when RL starts from SFT or ANTI models. For
example, the SFT \textsc{Groot}-4 and \textsc{VS}-4 models end at 14.4
and 13.2 respectively, while the IID-4 models peak at 10.8.

\input{tables/postrl_all_table} \input{tables/postrl_sft_anti_table}

\subsection{Performance by native difficulty} \label{app:native}

Although we define frontier performance relative to each model, we
could also use external definitions of problem
difficulty. Tables~\ref{tab:native_cobalt}, \ref{tab:native_lcb},
and~\ref{tab:native_ojbench} split the full test sets by the
benchmarks' own easy/medium/hard difficulty tiers.
\input{tables/native_cobalt_table} \input{tables/native_lcb_table}
\input{tables/native_ojbench_table} We find that performance gains
predominantly occur in categories where the base model had neither
high ($>80\%$ pass@1) nor extremely low ($<5\%$ pass@1)
performance. For example, \textsc{Groot}-based models increase the
base model's pass@64 from 57.7 to 65.4 and from 20.0 to 28.0 on medium
and hard LiveCodeBench problems respectively, and from 18.5 to 29.3 on
medium OJBench problems. On hard OJBench problems however, where the
base model only has 2.0\% pass@64, performance does not change
(2.4\%), and the same is true of easy LiveCodeBench problems, where
the base model is already at 97.7\%. The trend holds the same for
test-time scaling: Tables~\ref{tab:native_rsa_cobalt}
and~\ref{tab:native_rsa_lcb} give the same split for RSA, and
the strategic models aggregate best on the middle tiers
  (Cobalt-medium: \textsc{VS}-4 44.0 and \textsc{Groot}-4 39.5,
  against 27.3 for the base model; LiveCodeBench-hard: \textsc{VS}-4
  15.1 and \textsc{Groot}-4 14.5, against 10.9). On the hardest Cobalt
  tier every model ends comparable to the base model (9.1 to
  12.8).

\input{tables/rsa_full_table} 
\input{tables/native_rsa_cobalt_table}
\input{tables/native_rsa_pair}

\subsection{NCP details} \label{app:ncp}

\textbf{Data.} We use the book dataset from
\citet{gurung_learning_2025}: 30 books split into train, validation,
and test by book (22, 4, and 4 books). We further split each chapter
into sections of $\sim$200 words, giving 7,349 training, 1,147
validation, and 1,457 test sections. We re-generate the Story
Information document for each book-chapter combination with
gpt-oss-120B, and use it to edit the existing chapter summaries into
per-section guidance that ensures the plans we generate are specific
to the next chunk of the chapter. Each prompt contains the Story
Information, the per-section guidance, and the target length. We
exclude sections that are outside a 200-420 word range, or that only
contain chapter header information. Formally, the
improvement of a plan is $100\times\left(1 -
\mathrm{PPL}_{\mathrm{plan}}/\mathrm{PPL}_{\mathrm{no\text{-}plan}}\right)$,
where both perplexities are token-level means over the true section
under the frozen base model, with and without the plan in context.

\input{figures/fig_ncp_curves} \textbf{Sampling.} \textsc{Groot} and
\textsc{VS} make two planner calls per section, each producing eight
approaches, and one solver call per approach, giving 16 candidate
plans; IID makes 16 independent plan calls. By default, sampling is at
$T=0.85$.
Each answer is a short reasoning trace followed by the plan;
the solver prompt hides the approach, and we drop the 0.4\% of
candidates that mention it.

\textbf{Scoring.} Following the VR-CLI reward of
\citet{gurung_learning_2025}, a plan is scored by the mean per-token
log-likelihood of the true section under the base model with the plan
present, minus the same quantity without a plan; we report this as a
percent improvement in perplexity. The scorer is always the base
model. A candidate succeeds at threshold $X$ if its improvement is at
least $X\%$. Table~\ref{tab:mining} gives, for each
threshold, the number of training sections with at least one of their
16 candidate plans clearing it.

\textbf{Training and evaluation.} A training set keeps every candidate
clearing its threshold; the sampling methods are compared at equal
budget (at 20\%, budget 16: 1,349 \textsc{Groot}, 1,388 \textsc{VS},
and 952 IID examples). Training follows the code experiments (LoRA,
four epochs, best checkpoint by validation loss), and we train each
model three times. At test time we score the same way and estimate
coverage at $k$ using the same estimator of \citet{chen2021evaluating}
on the binary thresholded scores; coverage is over all 1,430 test
sections. Table~\ref{tab:ncp} gives coverage@32 at the three
thresholds and Figure~\ref{fig:ncp_curves} the full curves;
  the suffix on each model name is its sampling budget, 8 or 16
  candidate plans per training section. The ordering is consistent
across all $k$ and thresholds: \textsc{VS} and \textsc{Groot} are
close to each other and clearly above both the base model and IID,
which are themselves separated by at most roughly one seed
  standard deviation. For example, models trained at the
  20\% threshold reach a coverage@32 at the 10\% bar of 15.0 for
  \textsc{Groot}-16 and 16.6 for \textsc{VS}-16, against 9.7 for
  IID-16 and 9.5 for the base model. Raising the sampling budget from 8
  to 16 helps \textsc{VS} (39.8 to 43.1 at the 5\% bar) but leaves
  \textsc{Groot} (39.9 to 39.1) and IID (27.2 to 28.3) within seed
  noise.

\textbf{Filters and RL.} Table~\ref{tab:ncpstudents} compares the three dataset-creation filters: RFT, SFT, and ANTI. RFT filters for only datapoints over an improvement threshold of 20\%, and we use the size of the RFT set, $d$ for each sampling method to set the size of the other filtered datasets. SFT randomly samples $d$ responses, and ANTI selects the $d$ worst responses by the percent-improvement metric. On
IID data all filters perform similarly: RFT, SFT and ANTI are all within 1\% of each other at coverage@32 with a 10\%
threshold. On \textsc{Groot} and \textsc{VS} we find much larger variance, although performance across filters still remains above the best IID performance. This matches our findings on the competitive programming tasks, where we found that training on strategically sampled data outperforms IID data even when incorrect (Table~\ref{tab:filters_all}).

Table~\ref{tab:ncprl} gives coverage before and after RL from each RFT dataset, and the base model. Due to the high cost of RL on this setting we only report results from one RL seed. Our reward optimizes a combination of terms: high percent-improvement on the goal chapter section, and contrastive low percent-improvement on ``foil'' sections taken from another book and another chapter in the same book. This combination is meant to encourage \textit{uniquely} positive plans, that do not collapse to generic writing advice. To best represent the goal of RL training we therefore report coverage@k results with and without contrastive thresholds (in addition to percent-improvement $\geq$ threshold, both foils $<$ 5\%).

Only \textsc{Groot}-16 improves on the positive threshold-only metric after RL, from 15.0 to 26.3 at the
10\% threshold, while every other initialization declines. Adding the contrastive terms shows improvement for all models after RL, but \textsc{Groot} still performs the best. Similar to our competitive programming domain, \textsc{VS} performs better than \textsc{Groot} before RL, but not after.

\input{tables/ncp_v2_table}
\input{tables/ncpstudents_table}
\input{tables/ncprl_ksweep_table}
\input{tables/ncprl_foils_table}

\subsection{Hyperparameters} \label{app:hparams}

Table~\ref{tab:hparams} lists the sampling, training, evaluation, RSA,
and RL settings.  They are shared across datasets apart from
  the places the table marks a code/NCP split: the default
  sampling budget (4 per problem for code, 16 per section for NCP) and
  the number of sequences per step and maximum sequence length used in
  training. Nemotron3-Nano-4B uses the same settings except for the
LoRA target modules, since its Mamba layers do not accept LoRA on
their convolution and output projections. We run Nemotron models in their non-thinking mode at every stage due to extremely high truncation rates (41.9\% of samples with a 16k limit), with one line added to code prompts to keep an explicit reasoning step: ``Before writing code think step by step through the problem.''

\input{tables/hparams_table}

\clearpage
\input{sections/appendix_prompts}
\input{sections/appendix_examples}


\end{document}

%% file: math_commands.tex
\usepackage{amsmath,amsfonts,bm}

\def\eqref#1{equation~\ref{#1}}

\def\1{\bm{1}}

\DeclareMathAlphabet{\mathsfit}{\encodingdefault}{\sfdefault}{m}{sl}
\SetMathAlphabet{\mathsfit}{bold}{\encodingdefault}{\sfdefault}{bx}{n}



%% file: figures/fig_pipeline.tex
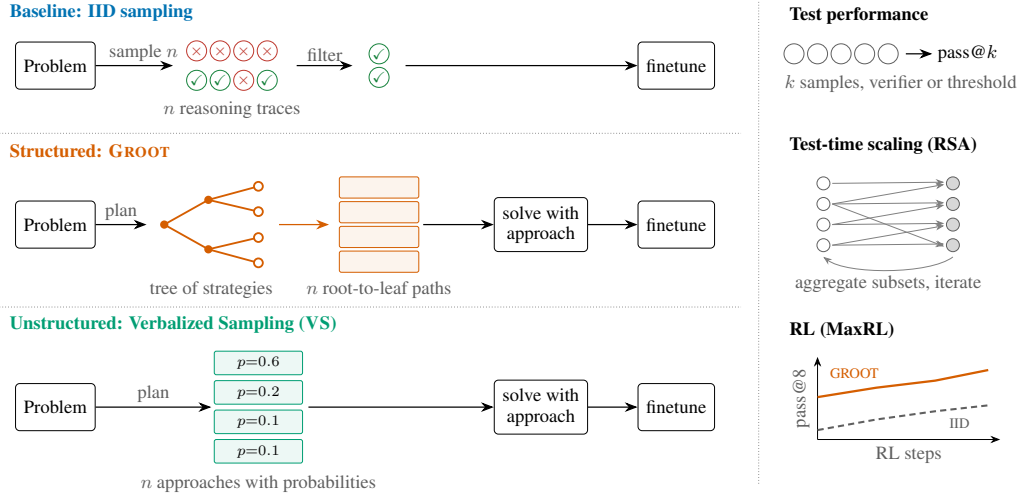
\begin{figure}[t]
\centering
\resizebox{\textwidth}{!}{\input{figures/fig1_pipeline_body.tex}}
\caption{Generating strategically diverse training data.  All 
  methods sample from the base model 
  at the same budget. IID
  sampling draws $n$ independent solutions and keeps the correct
  ones. \textsc{Groot} generates a tree of strategies and samples $n$
  root-to-leaf paths; \textsc{VS}  directly proposes $n$~approaches
  with their probabilities.  Each approach is  passed back as a
  hidden instruction to produce a solution. Trained models are
  evaluated with pass@$k$, test-time scaling, and RL.} \vspace{-30pt}
\label{fig:pipeline}
\end{figure}

%% file: figures/fig1_pipeline_body.tex
\definecolor{iidC}{RGB}{0,114,178}
\definecolor{grootC}{RGB}{213,94,0}
\definecolor{vsC}{RGB}{0,158,115}
\definecolor{okG}{RGB}{22,130,60}
\definecolor{badR}{RGB}{190,55,45}
\definecolor{dimN}{RGB}{95,95,95}

\begin{tikzpicture}[
  x=1cm,y=1cm,
  every node/.style={font=\small},
  box/.style={draw, rounded corners=2pt, minimum height=0.8cm, inner sep=4pt, align=center},
  flow/.style={-{Stealth[length=2.2mm]}, semithick},
  lbl/.style={font=\small, inner sep=1pt},
  sol/.style={circle, draw=dimN, minimum size=0.34cm, inner sep=0pt, font=\scriptsize},
  card/.style={draw, rounded corners=1pt, minimum width=1.35cm, minimum height=0.4cm, inner sep=1pt},
]

\draw[densely dotted, gray] (0,5.95) -- (12.6,5.95);
\draw[densely dotted, gray] (0,3.0) -- (12.6,3.0);
\draw[densely dotted, gray] (12.9,0.0) -- (12.9,8.2);

\node[font=\small\bfseries, anchor=west, text=iidC] at (0.05,8.0) {Baseline: IID sampling};
\node[box] (prob1) at (0.95,7.1) {Problem};
\draw[flow] (prob1.east) -- (2.95,7.1) node[pos=0.62, above=1pt, lbl, text=dimN] {sample $n$};
\foreach \xx in {3.35,3.75,4.15,4.55} {
  \node[sol, draw=badR] at (\xx,7.35) {\textcolor{badR}{$\times$}};
}
\foreach \xx in {3.35,3.75,4.55} {
  \node[sol, draw=okG] at (\xx,6.85) {\textcolor{okG}{\checkmark}};
}
\node[sol, draw=badR] at (4.15,6.85) {\textcolor{badR}{$\times$}};
\node[lbl, text=dimN] at (3.95,6.35) {$n$ reasoning traces};
\draw[flow] (5.05,7.1) -- (6.0,7.1) node[midway, above=1pt, lbl, text=dimN] {filter};
\node[sol, draw=okG] at (6.45,7.3) {\textcolor{okG}{\checkmark}};
\node[sol, draw=okG] at (6.45,6.9) {\textcolor{okG}{\checkmark}};
\node[box] (ft1) at (11.5,7.1) {finetune};
\draw[flow] (6.9,7.1) -- (ft1.west);

\node[font=\small\bfseries, anchor=west, text=grootC] at (0.05,5.65) {Structured: \textsc{Groot}};
\node[box] (prob2) at (0.95,4.4) {Problem};
\draw[flow] (prob2.east) -- (2.5,4.4) node[midway, above=1pt, lbl, text=dimN] {plan};
\begin{scope}[line width=0.9pt]
  \coordinate (r) at (2.8,4.4);
  \coordinate (n1) at (3.55,4.82); \coordinate (n2) at (3.55,3.98);
  \coordinate (l1) at (4.4,5.05);  \coordinate (l2) at (4.4,4.62);
  \coordinate (l3) at (4.4,4.18);  \coordinate (l4) at (4.4,3.75);
  \draw[grootC] (r) -- (n1) -- (l1); \draw[grootC] (n1) -- (l2);
  \draw[grootC] (r) -- (n2) -- (l3); \draw[grootC] (n2) -- (l4);
  \fill[grootC] (r) circle (0.07); \fill[grootC] (n1) circle (0.07); \fill[grootC] (n2) circle (0.07);
  \foreach \p in {l1,l2,l3,l4} \draw[grootC, fill=white] (\p) circle (0.075);
\end{scope}
\node[lbl, text=dimN] at (3.6,3.3) {tree of strategies};
\draw[flow, grootC] (4.75,4.4) -- (5.6,4.4);
\foreach \yy in {5.03,4.61,4.19,3.77}
  \node[card, draw=grootC, fill=grootC!7, minimum height=0.36cm] at (6.45,\yy) {};
\node[lbl, text=dimN] at (6.45,3.3) {$n$ root-to-leaf paths};
\node[box] (solve2) at (9.2,4.4) {solve with\\approach};
\draw[flow] (7.2,4.4) -- (solve2.west);
\node[box] (ft2) at (11.5,4.4) {finetune};
\draw[flow] (solve2.east) -- (ft2.west);

\node[font=\small\bfseries, anchor=west, text=vsC] at (0.05,2.7) {Unstructured: Verbalized Sampling (\textsc{VS})};
\node[box] (prob3) at (0.95,1.3) {Problem};
\draw[flow] (prob3.east) -- (3.6,1.3) node[pos=0.5, above=1pt, lbl, text=dimN] {plan};
\foreach \yy/\pp in {2.05/0.6, 1.55/0.2, 1.05/0.1, 0.55/0.1}
  \node[card, draw=vsC, fill=vsC!7, minimum width=1.5cm] at (4.4,\yy) {\scriptsize $p{=}\pp$};
\node[lbl, text=dimN] at (4.4,0.0) {$n$ approaches with probabilities};
\node[box] (solve3) at (9.2,1.3) {solve with\\approach};
\draw[flow] (5.25,1.3) -- (solve3.west);
\node[box] (ft3) at (11.5,1.3) {finetune};
\draw[flow] (solve3.east) -- (ft3.west);

\node[font=\small\bfseries, anchor=west] at (13.3,7.95) {Test performance};
\foreach \xx in {13.5,13.9,14.3,14.7,15.1} \node[sol, draw=dimN] at (\xx,7.3) {};
\draw[flow] (15.4,7.3) -- (15.85,7.3);
\node[lbl, anchor=west] at (15.9,7.3) {pass@$k$};
\node[lbl, text=dimN, anchor=west] at (13.3,6.8) {$k$ samples, verifier or threshold};

\node[font=\small\bfseries, anchor=west] at (13.3,5.75) {Test-time scaling (RSA)};
\tikzset{rsadot/.style={circle, draw=dimN, minimum size=0.22cm, inner sep=0pt}}
\foreach \yy in {4.05,4.4,4.75,5.1} \node[rsadot] at (14.0,\yy) {};
\foreach \yy in {4.05,4.4,4.75,5.1} \node[rsadot, fill=dimN!25] at (16.2,\yy) {};
\foreach \a/\b in {5.1/5.12, 4.75/5.08, 4.75/4.77, 4.4/4.73, 4.4/4.42, 4.05/4.38, 4.05/4.07, 4.75/4.03}
  \draw[-{Stealth[length=1.4mm]}, thin, gray] (14.15,\a) -- (16.05,\b);
\draw[-{Stealth[length=1.4mm]}, thin, gray] (16.2,3.85) .. controls (15.8,3.55) and (14.4,3.55) .. (14.0,3.85);
\node[lbl, text=dimN] at (15.1,3.4) {aggregate subsets, iterate};

\node[font=\small\bfseries, anchor=west] at (13.3,2.6) {RL (MaxRL)};
\begin{scope}[shift={(13.9,0.75)}]
  \draw[-{Stealth[length=1.6mm]}, thin] (0,0) -- (3.1,0) node[below=2pt, pos=0.5, lbl, text=dimN] {RL steps};
  \draw[-{Stealth[length=1.6mm]}, thin] (0,0) -- (0,1.4);
  \node[lbl, text=dimN, rotate=90] at (-0.28,0.7) {pass@8};
  \draw[grootC, line width=1.1pt] (0,0.72) -- (1.0,0.88) -- (2.0,1.0) -- (2.9,1.18);
  \draw[dimN, densely dashed, line width=1.0pt] (0,0.16) -- (1.0,0.34) -- (2.0,0.48) -- (2.9,0.58);
  \node[lbl, text=grootC, anchor=west] at (0.17,1.12) {\scriptsize GROOT};
  \node[lbl, text=dimN, anchor=west] at (2.2,0.28) {\scriptsize IID};
\end{scope}
\end{tikzpicture}

%% file: tables/mining_v2_table.tex
\begin{wraptable}[18]{r}{0.53\textwidth}
\vspace{-.1cm}
\centering\small
\footnotesize\setlength{\tabcolsep}{3pt}
\caption{Qwen3-4B sampling results. Code: number of 1,833 Cobalt training-frontier problems solved at least once at each sampling budget. NCP: number of 6,295 training sections with $>0$ of 16 plans reaching each perplexity-improvement threshold. Every row is over the same sections, the ones all six sampling runs cover.}
\label{tab:mining}
\vspace{-.1cm}
\begin{tabular}{lrrr@{\hskip 8pt}rrr}
\toprule
 & \multicolumn{3}{c@{\hskip 8pt}}{Code} & \multicolumn{3}{c}{NCP} \\
\cmidrule(lr){2-4}\cmidrule(lr){5-7}
Method & 4 & 8 & 64 & 5\% & 10\% & 15\% \\
\midrule
IID & \hphantom{0}42 & \hphantom{0}77 & 312 & 1,921 & 631 & 156 \\
IID ($T{=}1.5$) & \hphantom{0}60 & 106 & 379 & 5,326 & 3,545 & 1,495 \\
\addlinespace
\textsc{Groot} & 155 & 240 & --- & 2,775 & 1,077 & 286 \\
\textsc{Groot} ($T{=}1.5$) & \textbf{158} & \textbf{243} & --- & 5,366 & 3,474 & 1,505 \\
\addlinespace
\textsc{VS} & 137 & 217 & --- & 2,835 & 1,127 & 321 \\
\textsc{VS} ($T{=}1.5$) & 142 & 222 & --- & 5,310 & 3,493 & 1,491 \\
\bottomrule
\end{tabular}
\end{wraptable}

%% file: tables/code_passk_table.tex
\begin{table*}[t]
\centering
\footnotesize
\caption{\label{tab:code_passk} Code pass@$k$ for Qwen3-4B-Instruct on frontier subsets (problems solved by the base model $\leq 2$ of 64 times). Cobalt is in-distribution; LiveCodeBench and OJBench are held out, with Macro-Avg averaging them. Suffix~-$n$ denotes responses sampled per problem (e.g., \textsc{Groot}-4 uses four). Models use RFT unless marked ANTI, which retains only \emph{incorrect} responses. Bold marks the best value per column; full test-set results, additional datasets, and budgets appear in Appendix~\ref{app:ablations}.}
\begin{tabular}{l@{\hskip 10pt}c@{\hskip 13pt}c@{\hskip 8pt}c@{\hskip 8pt}c}
\toprule
Method & Cobalt & LiveCodeBench & OJBench & Macro-Avg \\
 & $p@1/8/64$ & $p@1/8/64$ & $p@1/8/64$ & $p@1/8/64$ \\
\midrule
Base & 0.5/3.6/21.9 & 0.2/1.3/7.4 & 0.1/0.5/2.9 & 0.2/0.9/5.2 \\
\addlinespace[2pt]\hdashline\addlinespace[2pt]
IID-4 & 0.7/4.9/19.8 & 0.2/1.5/7.0 & 0.1/0.7/3.5 & 0.2/1.1/5.2 \\
IID-4 ($T{=}1.5$) & 1.0/6.5/22.4 & 0.5/3.2/10.8 & 0.1/1.0/5.2 & 0.3/2.1/8.0 \\
IID-64 & 1.0/6.3/21.5 & 0.4/2.5/8.7 & 0.1/1.0/4.4 & 0.2/1.8/6.5 \\
IID-64 ($T{=}1.5$) & 1.5/7.4/23.0 & 0.4/2.8/10.7 & 0.2/1.4/5.2 & 0.3/2.1/7.9 \\
\addlinespace[2pt]\hdashline\addlinespace[2pt]
\textsc{Groot}-4 & 3.5/14.1/\textbf{30.5} & 1.5/8.0/19.5 & 1.2/5.3/11.5 & 1.4/6.7/15.5 \\
\textsc{Groot}-4 (ANTI) & 3.1/11.8/26.4 & 1.2/6.3/17.1 & 1.0/3.7/8.4 & 1.1/5.0/12.8 \\
\addlinespace[3.5pt]
\textsc{VS}-4 & \textbf{4.1}/\textbf{14.3}/30.2 & \textbf{2.0}/\textbf{9.6}/\textbf{23.1} & \textbf{1.5}/\textbf{5.6}/\textbf{11.9} & \textbf{1.8}/\textbf{7.6}/\textbf{17.5} \\
\textsc{VS}-4 (ANTI) & 3.3/12.5/29.1 & 1.2/7.0/20.3 & 0.9/3.8/9.2 & 1.1/5.4/14.8 \\
\bottomrule
\end{tabular}
\vspace{-.5cm}
\end{table*}

%% file: figures/fig_ncp_rl_pair.tex
\begin{figure}[t]
\centering
\begin{subfigure}[b]{0.531648\linewidth}
\centering
\resizebox{!}{126pt}{\begingroup\setlength{\linewidth}{177pt}\input{figures/fig_ncp_primary_tikz}\endgroup}%
\phantomsubcaption\label{fig:ncp_curves_primary}
\end{subfigure}\hfill
\begin{subfigure}[b]{0.4378\linewidth}
\centering
\includegraphics[height=126pt,trim={7.488 2.911 6.8465 6.8465},clip]{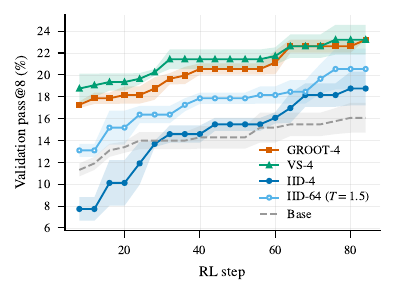}%
\phantomsubcaption\label{fig:rl_performance}
\end{subfigure}
\vspace{-5pt}
\caption{Qwen3-4B. \textbf{(a)} NCP coverage of test sections at the 15\% improvement threshold as
  the number of sampled plans $k$ increases. \textbf{(b)} Best-so-far pass@8 on the Cobalt
  validation frontier during RL from each budget-4 RFT model, mean and standard error over three
  seeds; IID-based models need many steps to reach where the strategic models (\textsc{Groot},
  \textsc{VS}) begin.}
\label{fig:ncp_rl_pair}
\vspace{-20pt}
\end{figure}

%% file: figures/fig_ncp_primary_tikz.tex
\definecolor{ncpIID}{RGB}{0,114,178}
\definecolor{ncpGroot}{RGB}{213,94,0}
\definecolor{ncpVS}{RGB}{0,158,115}
\definecolor{ncpBase}{RGB}{153,153,153}
\begin{tikzpicture}
\begin{axis}[
  width=\linewidth, height=0.68\linewidth,
  scale only axis=false,
  xmode=log, log basis x=2,
  xmin=0.84, xmax=38.1, ymin=0.11, ymax=4.27,
  xtick={1,2,4,8,16,32}, xticklabels={1,2,4,8,16,32},
  ytick={0.5,1.0,1.5,2.0,2.5,3.0,3.5,4.0},
  yticklabel style={/pgf/number format/fixed,/pgf/number format/fixed zerofill,/pgf/number format/precision=1},
  xlabel={samples $k$}, ylabel={coverage (\%)},
  label style={font=\footnotesize}, tick label style={font=\footnotesize},
  grid=both, major grid style={line width=0.4pt,draw=black!22},
  axis line style={line width=0.5pt}, tick style={black,line width=0.5pt},
  axis on top=false,
  legend style={font=\footnotesize, at={(0.02,0.98)}, anchor=north west,
                draw=none, fill=none, row sep=-1.5pt, inner sep=1pt},
  legend cell align=left,
  mark size=1.7pt, line width=0.9pt,
]
\addplot[ncpBase, densely dashed, mark=none]
  coordinates {(1,0.297) (2,0.503) (4,0.783) (8,1.132) (16,1.533) (32,1.970)};
\addlegendentry{Base}
\addplot[ncpIID, mark=*, mark options={fill=ncpIID}]
  coordinates {(1,0.308) (2,0.519) (4,0.805) (8,1.169) (16,1.601) (32,2.081)};
\addlegendentry{IID}
\addplot[ncpGroot, mark=square*, mark options={fill=ncpGroot}]
  coordinates {(1,0.565) (2,0.926) (4,1.419) (8,2.039) (16,2.762) (32,3.602)};
\addlegendentry{\textsc{Groot}}
\addplot[ncpVS, mark=triangle*, mark options={fill=ncpVS,scale=1.15}]
  coordinates {(1,0.669) (2,1.071) (4,1.609) (8,2.290) (16,3.113) (32,4.080)};
\addlegendentry{\textsc{VS}}
\end{axis}
\end{tikzpicture}

%% file: tables/teacher_table.tex
\begin{table*}[t]
\centering
\footnotesize
\caption{\label{tab:teacher} Frontier-set pass@$k$ with teacher- versus self-generated training data. Teacher data uses Qwen3-235B-A22B-Instruct for both approaches and solutions; self-generated data uses the 4B student. In all cases the trained model is Qwen3-4B-Instruct, with sampling budget 4 and a 16k token cap (increased from 8k due to truncation in teacher-trained models). Full test sets, budget 8, and 8k-cap results are in Appendix~\ref{app:teacher}.}
\begin{tabular}{l@{\hskip 8pt}c|c@{\hskip 8pt}c@{\hskip 8pt}c}
\hline
Method & Cobalt & LiveCodeBench & OJBench & Macro-Avg \\
 & $p@1/8/64$ & $p@1/8/64$ & $p@1/8/64$ & $p@1/8/64$ \\
\hline
IID-4 (self) & 0.7/5.2/20.5 & 0.3/1.9/8.2 & 0.1/0.7/3.3 & 0.2/1.3/5.8 \\
IID-4 (teacher) & 4.6/16.7/39.3 & 1.1/6.9/19.1 & 0.3/2.1/7.6 & 0.7/4.5/13.4 \\
\hdashline
GROOT-4 (self) & 4.8/18.5/39.5 & 2.4/11.3/25.0 & 1.8/7.3/\textbf{15.2} & 2.1/9.3/20.1 \\
GROOT-4 (teacher) & 5.0/19.4/\textbf{40.3} & 1.8/9.3/20.5 & 1.4/6.7/15.0 & 1.6/8.0/17.8 \\
VS-4 (self) & \textbf{5.5}/\textbf{19.6}/\textbf{40.3} & \textbf{3.1}/\textbf{13.6}/\textbf{30.8} & \textbf{2.1}/\textbf{7.6}/14.7 & \textbf{2.6}/\textbf{10.6}/\textbf{22.8} \\
VS-4 (teacher) & 4.6/18.6/39.4 & 1.8/9.4/21.6 & 1.1/6.1/13.8 & 1.5/7.8/17.7 \\
\hline
\end{tabular}
\vspace{-10pt}
\end{table*}


%% file: tables/rsa_pair.tex
\begin{table}[t]
\centering
\begin{minipage}[t]{0.49\linewidth}
\vspace{0pt}
\centering
\caption{Qwen3-4B RSA pass@1 on code frontier subsets before and after 10 aggregation iterations (population 16). Mean over three runs; bold marks column bests. IID-64 uses $T=1.5$. Full test-set results are in Appendix~\ref{app:native}.}
\label{tab:rsa}
\input{tables/rsa_body}
\end{minipage}\hfill
\begin{minipage}[t]{0.49\linewidth}
\vspace{-1pt}
\centering
\caption{Qwen3-4B NCP RSA results, coverage@1 at each perplexity-improvement threshold for 16-plan populations, before~$\rightarrow$~after ten aggregation iterations. Values are means over three aggregation runs; bold marks the best final result in each column.}
\label{tab:ncp_rsa}
\input{tables/ncp_rsa_v2_body}
\end{minipage}
\vspace{-10pt}
\end{table}

%% file: tables/rsa_body.tex
\footnotesize\setlength{\tabcolsep}{0pt}
\begin{tabular}{lccc}
\toprule
Method & Cobalt & LCB & OJB \\
\midrule
Base & \hphantom{0}0.8 $\rightarrow$ \hphantom{0}3.1 & \hphantom{0}0.5 $\rightarrow$ \hphantom{0}2.3 & \hphantom{0}0.1 $\rightarrow$ \hphantom{0}0.8 \\
IID-4 & \hphantom{0}0.9 $\rightarrow$ \hphantom{0}3.8 & \hphantom{0}0.6 $\rightarrow$ \hphantom{0}2.1 & \hphantom{0}0.1 $\rightarrow$ \hphantom{0}0.7 \\
IID-64 & \hphantom{0}1.4 $\rightarrow$ \hphantom{0}7.4 & \hphantom{0}0.6 $\rightarrow$ \hphantom{0}4.0 & \hphantom{0}0.2 $\rightarrow$ \hphantom{0}1.8 \\
\addlinespace[2pt]\hdashline\addlinespace[2pt]
\textsc{Groot}-4 & \hphantom{0}3.5 $\rightarrow$ 10.4 & \hphantom{0}2.1 $\rightarrow$ \hphantom{0}7.0 & \hphantom{0}1.5 $\rightarrow$ \hphantom{0}4.5 \\
\textsc{VS}-4 & \hphantom{0}4.2 $\rightarrow$ \textbf{10.6} & \hphantom{0}2.3 $\rightarrow$ \textbf{\hphantom{0}7.9} & \hphantom{0}1.5 $\rightarrow$ \textbf{\hphantom{0}5.3} \\
\bottomrule
\end{tabular}

%% file: tables/ncp_rsa_v2_body.tex
\footnotesize
\setlength{\tabcolsep}{3pt}
\begin{tabular}{@{}lccc@{}}
\toprule
Method & $\geq$5\% & $\geq$10\% & $\geq$20\% \\
\midrule
Base & 11.6 $\rightarrow$ 22.3 & 3.1 $\rightarrow$ 6.9 & 0.4 $\rightarrow$ 0.6 \\
IID-16 & 10.7 $\rightarrow$ 18.1 & 2.9 $\rightarrow$ 5.4 & 0.4 $\rightarrow$ 0.5 \\
\addlinespace[2pt]\hdashline\addlinespace[2pt]
\textsc{Groot}-16 & 15.7 $\rightarrow$ \textbf{24.6} & 4.4 $\rightarrow$ 7.9 & 0.5 $\rightarrow$ 0.7 \\
\textsc{VS}-16 & 16.3 $\rightarrow$ \textbf{24.6} & 4.9 $\rightarrow$ \textbf{8.5} & 0.6 $\rightarrow$ \textbf{0.9} \\
\bottomrule
\end{tabular}

%% file: tables/postrl_table.tex
\begin{table*}[t]
\centering
\footnotesize
\caption{Qwen3-4B frontier-subset performance after RL from each budget-4 initialization; the full test sets are in Appendix~\ref{app:rl}. For each arm the first row is the student before RL and the second the mean over three RL runs of the checkpoints selected by validation pass@8. We report pass@1\,/\,pass@8\,/\,pass@64; Macro-Avg is the mean of the two held-out benchmarks, LiveCodeBench and OJBench. Bold marks the best value in each column.}
\label{tab:postrl}
\setlength{\tabcolsep}{4pt}
\newcolumntype{Y}{r@{\,/\,}r@{\,/\,}r}
\begin{tabular}{l@{\hskip 8pt}Y@{\hskip 6pt}|@{\hskip 6pt}Y@{\hskip 8pt}Y@{\hskip 8pt}Y}
\toprule
Method & \multicolumn{3}{c|}{Cobalt} & \multicolumn{3}{c}{LiveCodeBench} & \multicolumn{3}{c}{OJBench} & \multicolumn{3}{c}{Macro-Avg} \\
 & \multicolumn{3}{c|}{$p@1/8/64$} & \multicolumn{3}{c}{$p@1/8/64$} & \multicolumn{3}{c}{$p@1/8/64$} & \multicolumn{3}{c}{$p@1/8/64$} \\
\midrule
Base & \hphantom{0}0.5 & \hphantom{0}3.6 & 21.9 & \hphantom{0}0.2 & \hphantom{0}1.3 & \hphantom{0}7.4 & \hphantom{0}0.1 & \hphantom{0}0.5 & \hphantom{0}2.9 & \hphantom{0}0.2 & \hphantom{0}0.9 & \hphantom{0}5.2 \\
\quad +RL & \hphantom{0}4.5 & 13.9 & 29.3 & \hphantom{0}1.1 & \hphantom{0}5.9 & 15.3 & \hphantom{0}0.4 & \hphantom{0}2.4 & \hphantom{0}7.5 & \hphantom{0}0.8 & \hphantom{0}4.1 & 11.4 \\
\addlinespace[2pt]\hdashline\addlinespace[2pt]
IID-4 & \hphantom{0}0.7 & \hphantom{0}4.9 & 19.8 & \hphantom{0}0.2 & \hphantom{0}1.5 & \hphantom{0}7.0 & \hphantom{0}0.1 & \hphantom{0}0.7 & \hphantom{0}3.5 & \hphantom{0}0.2 & \hphantom{0}1.1 & \hphantom{0}5.2 \\
\quad +RL & \hphantom{0}5.5 & 17.3 & 36.6 & \hphantom{0}1.1 & \hphantom{0}6.0 & 15.4 & \hphantom{0}0.4 & \hphantom{0}2.0 & \hphantom{0}6.4 & \hphantom{0}0.7 & \hphantom{0}4.0 & 10.9 \\
\addlinespace[2pt]\hdashline\addlinespace[2pt]
GROOT-4 & \hphantom{0}3.5 & 14.1 & 30.5 & \hphantom{0}1.5 & \hphantom{0}8.0 & 19.5 & \hphantom{0}1.2 & \hphantom{0}5.3 & 11.5 & \hphantom{0}1.4 & \hphantom{0}6.7 & 15.5 \\
\quad +RL & \textbf{\hphantom{0}7.2} & \textbf{20.9} & \textbf{39.0} & \textbf{\hphantom{0}2.2} & \textbf{\hphantom{0}9.5} & \textbf{22.2} & \textbf{\hphantom{0}1.8} & \textbf{\hphantom{0}6.8} & \textbf{13.9} & \textbf{\hphantom{0}2.0} & \textbf{\hphantom{0}8.1} & \textbf{18.0} \\
\addlinespace[3.5pt]
VS-4 & \hphantom{0}4.1 & 14.3 & 30.2 & \hphantom{0}2.0 & \hphantom{0}9.6 & 23.1 & \hphantom{0}1.5 & \hphantom{0}5.6 & 11.9 & \hphantom{0}1.8 & \hphantom{0}7.6 & 17.5 \\
\quad +RL & \hphantom{0}6.2 & 19.4 & 37.8 & \hphantom{0}2.1 & \hphantom{0}8.9 & 20.5 & \hphantom{0}1.4 & \hphantom{0}5.6 & 11.5 & \hphantom{0}1.7 & \hphantom{0}7.2 & 16.0 \\
\bottomrule
\end{tabular}
\vspace{-15pt}
\end{table*}

%% file: tables/n3n_passk_table.tex
\begin{table*}[t]
\centering
\footnotesize
\setlength{\aboverulesep}{0pt}\setlength{\belowrulesep}{0pt}
\setlength{\extrarowheight}{2pt}
\caption{Nemotron3-Nano-4B: code pass@$k$ on the full test sets and on the Nemotron base model's own frontier (519 problems, 428 shared with the Qwen frontier); Nemotron and Qwen numbers are not comparable in absolute terms. We report pass@1/8/64; Macro-Avg is the mean of the two held-out benchmarks, LiveCodeBench and OJBench. Bold marks the best value in each column.}
\label{tab:n3n_passk}
\begin{tabular}{l@{\hskip 8pt}c|c@{\hskip 8pt}c@{\hskip 8pt}c}
\hline
Method & Cobalt & LiveCodeBench & OJBench & Macro-Avg \\
 & $p@1/8/64$ & $p@1/8/64$ & $p@1/8/64$ & $p@1/8/64$ \\
\hline
\multicolumn{2}{l@{\hskip 6pt}|@{\hskip 6pt}}{\textit{Full dataset}} & \multicolumn{3}{l}{} \\
Base & 11.4/32.4/52.2 & 23.1/38.2/50.4 & 3.2/6.6/10.5 & 13.2/22.4/30.4 \\[3.5pt]
IID-4 & 12.9/33.7/51.8 & 23.4/37.5/49.0 & 3.3/7.0/11.5 & 13.3/22.2/30.2 \\
IID-8 & \textbf{15.4}/38.0/55.2 & \textbf{26.7}/40.8/52.5 & 3.9/8.8/16.5 & \textbf{15.3}/24.8/34.5 \\
IID-64 & 13.1/33.9/52.5 & 24.1/38.9/51.9 & 3.5/7.5/12.5 & 13.8/23.2/32.2 \\[3.5pt]
GROOT-4 & 13.2/36.0/56.7 & 25.6/41.9/\textbf{55.8} & 5.0/12.5/22.0 & \textbf{15.3}/27.2/38.9 \\
GROOT-8 & 13.3/35.5/57.0 & 24.9/40.0/51.3 & 4.6/11.5/20.6 & 14.8/25.8/36.0 \\[3.5pt]
VS-4 & 14.1/\textbf{39.0}/\textbf{59.2} & 25.0/\textbf{43.4}/55.5 & \textbf{5.3}/\textbf{14.5}/\textbf{24.3} & 15.2/\textbf{28.9}/\textbf{39.9} \\
VS-8 & 14.3/37.2/57.5 & 25.5/41.0/53.3 & 4.9/12.0/20.7 & 15.2/26.5/37.0 \\
\hline\hline
\multicolumn{2}{l@{\hskip 6pt}|@{\hskip 6pt}}{\textit{Frontier ($\leq 2/64$)}} & \multicolumn{3}{l}{} \\
Base & 0.8/5.3/22.4 & 0.5/3.5/15.7 & 0.1/0.6/3.9 & 0.3/2.0/9.8 \\[3.5pt]
IID-4 & 0.9/6.0/23.0 & 0.5/3.2/12.7 & 0.1/1.0/4.9 & 0.3/2.1/8.8 \\
IID-8 & 1.5/8.8/27.1 & 0.8/5.1/19.3 & 0.4/2.8/10.4 & 0.6/3.9/14.9 \\
IID-64 & 0.9/5.9/22.6 & 0.6/4.2/17.8 & 0.2/1.2/6.0 & 0.4/2.7/11.9 \\[3.5pt]
GROOT-4 & 1.5/9.5/29.8 & 0.9/6.1/\textbf{23.5} & 1.4/6.7/16.3 & 1.1/6.4/19.9 \\
GROOT-8 & 1.3/8.5/29.9 & 0.9/5.5/16.0 & 1.1/5.6/14.8 & 1.0/5.5/15.4 \\
VS-4 & \textbf{2.1}/\textbf{12.5}/\textbf{34.2} & \textbf{1.3}/\textbf{7.7}/22.9 & \textbf{1.8}/\textbf{8.7}/\textbf{18.7} & \textbf{1.6}/\textbf{8.2}/\textbf{20.8} \\
VS-8 & 1.6/9.8/30.9 & 0.9/5.7/19.3 & 1.3/6.0/14.8 & 1.1/5.8/17.1 \\
\hline
\end{tabular}
\end{table*}

%% file: figures/fig_rl_n3n.tex
\begin{figure}[t]
\centering
\includegraphics[width=0.56\linewidth]{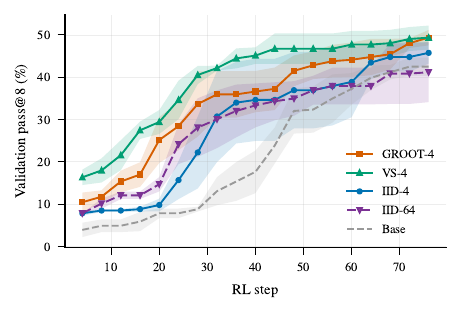}
\caption{Nemotron3 Nano-4B: best-so-far pass@8 on its validation frontier set during RL
training from each budget-4 model, mean and standard error over three seeds.}
\label{fig:rl_n3n}
\end{figure}

%% file: tables/n3n_postrl_table.tex
\begin{table*}[t]
\centering
\footnotesize
\setlength{\aboverulesep}{0pt}\setlength{\belowrulesep}{0pt}
\setlength{\extrarowheight}{2pt}
\caption{Nemotron3-Nano-4B competitive programming: test performance after RL from each initialization, mean over three RL runs. We report pass@1/8/64; Macro-Avg is the mean of the two held-out benchmarks, LiveCodeBench and OJBench. Bold marks the best value in each column.}
\label{tab:n3n_postrl}
\begin{tabular}{l@{\hskip 8pt}c|c@{\hskip 8pt}c@{\hskip 8pt}c}
\hline
Method & Cobalt & LiveCodeBench & OJBench & Macro-Avg \\
 & $p@1/8/64$ & $p@1/8/64$ & $p@1/8/64$ & $p@1/8/64$ \\
\hline
\multicolumn{2}{l@{\hskip 6pt}|@{\hskip 6pt}}{\textit{Full dataset}} & \multicolumn{3}{l}{} \\
Base & 11.4/32.4/52.2 & 23.1/38.2/50.4 & 3.2/6.6/10.5 & 13.2/22.4/30.4 \\
\quad +RL & 27.7/57.5/73.6 & 35.4/51.9/62.4 & 9.9/21.0/30.1 & 22.6/36.5/46.3 \\[3.5pt]
IID-4 & 12.9/33.7/51.8 & 23.4/37.5/49.0 & 3.3/7.0/11.5 & 13.3/22.2/30.2 \\
\quad +RL & 30.2/58.7/73.7 & \textbf{38.9}/\textbf{55.6}/65.6 & \textbf{11.0}/\textbf{22.8}/\textbf{31.5} & \textbf{25.0}/\textbf{39.3}/\textbf{48.5} \\[3.5pt]
IID-8 & 15.4/38.0/55.2 & 26.7/40.8/52.5 & 3.9/8.8/16.5 & 15.3/24.8/34.5 \\
\quad +RL & 28.8/56.9/72.3 & 36.8/52.9/62.7 & 8.1/17.5/25.3 & 22.4/35.2/44.0 \\[3.5pt]
IID-64 & 13.1/33.9/52.5 & 24.1/38.9/51.9 & 3.5/7.5/12.5 & 13.8/23.2/32.2 \\
\quad +RL & 26.9/55.1/70.8 & 35.4/50.5/60.0 & 8.8/18.3/27.3 & 22.1/34.4/43.7 \\[3.5pt]
GROOT-4 & 13.2/36.0/56.7 & 25.6/41.9/55.8 & 5.0/12.5/22.0 & 15.3/27.2/38.9 \\
\quad +RL & 28.7/57.7/73.5 & 36.6/53.6/63.4 & 10.7/22.0/30.5 & 23.6/37.8/47.0 \\[3.5pt]
GROOT-8 & 13.3/35.5/57.0 & 24.9/40.0/51.3 & 4.6/11.5/20.6 & 14.8/25.8/36.0 \\
\quad +RL & 29.8/58.5/73.5 & 37.8/54.9/64.7 & 9.8/21.1/29.2 & 23.8/38.0/46.9 \\[7pt]
VS-4 & 14.1/39.0/59.2 & 25.0/43.4/55.5 & 5.3/14.5/24.3 & 15.2/28.9/39.9 \\
\quad +RL & 29.0/57.5/73.4 & 37.5/55.5/\textbf{66.1} & 10.2/22.2/30.9 & 23.9/38.9/\textbf{48.5} \\[3.5pt]
VS-8 & 14.3/37.2/57.5 & 25.5/41.0/53.3 & 4.9/12.0/20.7 & 15.2/26.5/37.0 \\
\quad +RL & \textbf{30.6}/\textbf{59.3}/\textbf{74.2} & 37.0/53.8/63.6 & 10.7/21.8/29.5 & 23.9/37.8/46.6 \\
\hline\hline
\multicolumn{2}{l@{\hskip 6pt}|@{\hskip 6pt}}{\textit{Frontier ($\leq 2/64$)}} & \multicolumn{3}{l}{} \\
Base & 0.8/5.3/22.4 & 0.5/3.5/15.7 & 0.1/0.6/3.9 & 0.3/2.0/9.8 \\
\quad +RL & 11.7/35.9/57.4 & 7.0/20.3/35.7 & 5.3/15.4/25.0 & 6.1/17.8/30.3 \\[3.5pt]
IID-4 & 0.9/6.0/23.0 & 0.5/3.2/12.7 & 0.1/1.0/4.9 & 0.3/2.1/8.8 \\
\quad +RL & 13.0/37.0/57.4 & \textbf{8.9}/\textbf{24.5}/40.3 & \textbf{6.1}/\textbf{17.2}/\textbf{26.3} & \textbf{7.5}/\textbf{20.9}/33.3 \\[3.5pt]
IID-8 & 1.5/8.8/27.1 & 0.8/5.1/19.3 & 0.4/2.8/10.4 & 0.6/3.9/14.9 \\
\quad +RL & 12.3/35.1/55.7 & 7.9/21.1/35.5 & 3.6/11.7/19.9 & 5.7/16.4/27.7 \\[3.5pt]
IID-64 & 0.9/5.9/22.6 & 0.6/4.2/17.8 & 0.2/1.2/6.0 & 0.4/2.7/11.9 \\
\quad +RL & 10.7/32.1/53.0 & 6.7/18.0/31.1 & 4.2/12.5/22.0 & 5.5/15.3/26.6 \\[3.5pt]
GROOT-4 & 1.5/9.5/29.8 & 0.9/6.1/23.5 & 1.4/6.7/16.3 & 1.1/6.4/19.9 \\
\quad +RL & 12.2/36.3/57.6 & 7.1/21.9/36.6 & 5.9/16.4/25.3 & 6.5/19.1/31.0 \\[3.5pt]
GROOT-8 & 1.3/8.5/29.9 & 0.9/5.5/16.0 & 1.1/5.6/14.8 & 1.0/5.5/15.4 \\
\quad +RL & 12.9/37.4/57.3 & 8.4/23.9/38.8 & 5.1/15.4/24.0 & 6.7/19.6/31.4 \\[7pt]
VS-4 & 2.1/12.5/34.2 & 1.3/7.7/22.9 & 1.8/8.7/18.7 & 1.6/8.2/20.8 \\
\quad +RL & 12.3/35.1/56.8 & 8.2/24.2/\textbf{41.3} & 5.7/16.6/25.8 & 7.0/20.4/\textbf{33.6} \\[3.5pt]
VS-8 & 1.6/9.8/30.9 & 0.9/5.7/19.3 & 1.3/6.0/14.8 & 1.1/5.8/17.1 \\
\quad +RL & \textbf{13.6}/\textbf{38.0}/\textbf{58.7} & 8.4/22.7/37.3 & 6.0/16.2/24.3 & 7.2/19.5/30.8 \\
\hline
\end{tabular}
\end{table*}

%% file: tables/n3n_rsa_table.tex
\begin{table}[t]
\centering
\footnotesize\setlength{\tabcolsep}{4pt}
\caption{Nemotron3-Nano-4B competitive programming: RSA pass@1 before and after ten aggregation iterations (population size 16), mean over three runs; bold marks the best final value in each column.}
\label{tab:n3n_rsa}
\begin{tabular}{lccc|ccc}
\toprule
 & \multicolumn{3}{c|}{Full test set} & \multicolumn{3}{c}{Frontier ($\leq 2/64$)} \\
Method & Cobalt & LiveCodeBench & OJBench & Cobalt & LiveCodeBench & OJBench \\
\midrule
Base & \hphantom{0}8.0 $\rightarrow$ 21.9 & 21.0 $\rightarrow$ 28.2 & \hphantom{0}2.6 $\rightarrow$ \hphantom{0}7.7 & \hphantom{0}0.3 $\rightarrow$ \hphantom{0}3.9 & \hphantom{0}0.3 $\rightarrow$ \hphantom{0}2.2 & \hphantom{0}0.1 $\rightarrow$ \hphantom{0}3.5 \\
IID-4 & 10.7 $\rightarrow$ 23.8 & 21.9 $\rightarrow$ 29.9 & \hphantom{0}2.9 $\rightarrow$ \hphantom{0}9.1 & \hphantom{0}0.6 $\rightarrow$ \hphantom{0}4.8 & \hphantom{0}0.3 $\rightarrow$ \hphantom{0}0.9 & \hphantom{0}0.1 $\rightarrow$ \hphantom{0}4.1 \\
IID-8 & 13.6 $\rightarrow$ \textbf{26.7} & 25.9 $\rightarrow$ 32.5 & \hphantom{0}3.6 $\rightarrow$ \textbf{\hphantom{0}9.3} & \hphantom{0}1.0 $\rightarrow$ \textbf{\hphantom{0}5.5} & \hphantom{0}0.6 $\rightarrow$ \hphantom{0}2.8 & \hphantom{0}0.3 $\rightarrow$ \textbf{\hphantom{0}4.4} \\
\addlinespace[2pt]\hdashline\addlinespace[2pt]
\textsc{Groot}-4 & 11.5 $\rightarrow$ 22.8 & 23.4 $\rightarrow$ 28.8 & \hphantom{0}4.4 $\rightarrow$ \hphantom{0}8.0 & \hphantom{0}1.4 $\rightarrow$ \hphantom{0}4.3 & \hphantom{0}0.6 $\rightarrow$ \hphantom{0}2.4 & \hphantom{0}1.1 $\rightarrow$ \hphantom{0}3.8 \\
\textsc{Groot}-8 & 11.3 $\rightarrow$ 22.7 & 22.7 $\rightarrow$ 28.1 & \hphantom{0}3.7 $\rightarrow$ \hphantom{0}7.3 & \hphantom{0}1.0 $\rightarrow$ \hphantom{0}4.2 & \hphantom{0}0.6 $\rightarrow$ \hphantom{0}2.3 & \hphantom{0}0.7 $\rightarrow$ \hphantom{0}3.0 \\
\textsc{VS}-4 & 11.8 $\rightarrow$ 24.4 & 22.8 $\rightarrow$ \textbf{32.8} & \hphantom{0}4.4 $\rightarrow$ \hphantom{0}8.4 & \hphantom{0}1.6 $\rightarrow$ \hphantom{0}5.1 & \hphantom{0}1.1 $\rightarrow$ \textbf{\hphantom{0}3.9} & \hphantom{0}1.2 $\rightarrow$ \hphantom{0}3.8 \\
\textsc{VS}-8 & 12.3 $\rightarrow$ 22.8 & 23.8 $\rightarrow$ 30.3 & \hphantom{0}3.9 $\rightarrow$ \hphantom{0}8.2 & \hphantom{0}1.2 $\rightarrow$ \hphantom{0}4.2 & \hphantom{0}0.8 $\rightarrow$ \hphantom{0}2.0 & \hphantom{0}0.8 $\rightarrow$ \hphantom{0}4.0 \\
\bottomrule
\end{tabular}
\end{table}

%% file: tables/n3ncp_table.tex
\begin{table}[t]
\centering\footnotesize\setlength{\tabcolsep}{3pt}
\caption{Nemotron3-Nano-4B sampling results. Code: Cobalt training-frontier problems solved at least once at each sampling budget; the frontier is defined relative to Nemotron3-Nano-4B, so it is a different problem set from the one in Table~\ref{tab:mining}. NCP: number of 6,754 training sections with $>0$ of 16 plans reaching each perplexity-improvement threshold. Every row is over the same sections, the ones all three sampling runs cover.}
\label{tab:n3ncp}
\begin{tabular}{lrrr@{\hskip 8pt}rrr}
\toprule
 & \multicolumn{3}{c@{\hskip 8pt}}{Code} & \multicolumn{3}{c}{NCP} \\
\cmidrule(lr){2-4}\cmidrule(lr){5-7}
Method & 4 & 8 & 64 & 5\% & 10\% & 15\% \\
\midrule
IID & 52 & 97 & 424 & 5,375 & 2,807 & \hphantom{0}915 \\
\textsc{Groot} & 115 & 197 & --- & \textbf{5,725} & \textbf{3,340} & 1,140 \\
\textsc{VS} & \textbf{137} & \textbf{233} & --- & 5,707 & 3,286 & \textbf{1,172} \\
\bottomrule
\end{tabular}
\end{table}

%% file: tables/n3ncp_students_table.tex
\begin{table}[t]
\centering
\footnotesize
\setlength{\tabcolsep}{5pt}
\caption{Nemotron3-Nano-4B NCP coverage (\%) of test sections at three perplexity-improvement thresholds, for Nemotron3-Nano-4B trained on plans clearing a $20\%$ threshold at a sampling budget of 8. Coverage is over the same 1,430 test sections for every row. Bold marks the best value in each column.}
\label{tab:n3ncp_students}
\begin{tabular}{lcccccc}
\toprule
Method & \multicolumn{2}{c}{$5\%$} & \multicolumn{2}{c}{$10\%$} & \multicolumn{2}{c}{$15\%$} \\
 & $k=16$ & $k=64$ & $k=16$ & $k=64$ & $k=16$ & $k=64$ \\
\midrule
Base (no distill) & 74.9 & 86.7 & 37.4 & 51.4 & 11.1 & 18.6 \\
IID & 72.3 & 83.9 & 34.6 & 48.0 & 10.4 & 17.1 \\
\textsc{Groot} & \textbf{82.8} & \textbf{91.0} & \textbf{45.3} & \textbf{59.5} & \textbf{14.6} & \textbf{23.4} \\
\textsc{VS} & 81.4 & 90.8 & 43.6 & 57.7 & 14.2 & 22.6 \\
\bottomrule
\end{tabular}
\end{table}

%% file: tables/n3L_table.tex
\begin{table*}[t]
\centering
\footnotesize
\setlength{\aboverulesep}{0pt}\setlength{\belowrulesep}{0pt}
\setlength{\extrarowheight}{2pt}
\caption{Nemotron3-Nano-4B distilled from a larger teacher: code pass@$k$ on the full test sets and on the frontier subsets, which are defined relative to Nemotron3-Nano-4B. Each method appears twice, distilled from the 4B teacher and from the larger teacher (Nemotron3-Super-120B-A12B-BF16), with the generation budget and all other settings held fixed so that only the teacher changes. We report pass@1/8/64; Macro-Avg is the mean of the two held-out benchmarks, LiveCodeBench and OJBench. Bold marks the best value in each column.}
\label{tab:n3L}
\begin{tabular}{l@{\hskip 8pt}c|c@{\hskip 8pt}c@{\hskip 8pt}c}
\hline
Method & Cobalt & LiveCodeBench & OJBench & Macro-Avg \\
 & $p@1/8/64$ & $p@1/8/64$ & $p@1/8/64$ & $p@1/8/64$ \\
\hline
\multicolumn{2}{l@{\hskip 6pt}|@{\hskip 6pt}}{\textit{Full dataset}} & \multicolumn{3}{l}{} \\
Base & 11.4/32.4/52.2 & 23.1/38.2/50.4 & 3.2/6.6/10.5 & 13.2/22.4/30.4 \\[3.5pt]
IID-4 (self) & 12.9/33.7/51.8 & 23.4/37.5/49.0 & 3.3/7.0/11.5 & 13.3/22.2/30.2 \\
IID-4 (teacher) & 13.0/35.6/55.8 & 23.6/38.7/51.4 & 3.9/9.1/16.7 & 13.8/23.9/34.0 \\[3.5pt]
GROOT-4 (self) & 13.2/36.0/56.7 & 25.6/41.9/55.8 & 5.0/12.5/22.0 & 15.3/27.2/38.9 \\
GROOT-4 (teacher) & 16.4/43.3/65.6 & 27.7/46.8/59.9 & 6.2/16.0/26.4 & 16.9/31.4/43.1 \\[3.5pt]
VS-4 (self) & 14.1/39.0/59.2 & 25.0/43.4/55.5 & 5.3/14.5/24.3 & 15.2/28.9/39.9 \\
VS-4 (teacher) & \textbf{18.5}/\textbf{46.9}/\textbf{66.8} & \textbf{29.0}/\textbf{48.0}/\textbf{61.1} & \textbf{6.6}/\textbf{17.2}/\textbf{27.6} & \textbf{17.8}/\textbf{32.6}/\textbf{44.4} \\
\hline\hline
\multicolumn{2}{l@{\hskip 6pt}|@{\hskip 6pt}}{\textit{Frontier ($\leq 2/64$)}} & \multicolumn{3}{l}{} \\
Base & 0.8/5.3/22.4 & 0.5/3.5/15.7 & 0.1/0.6/3.9 & 0.3/2.0/9.8 \\[3.5pt]
IID-4 (self) & 0.9/6.0/23.0 & 0.5/3.2/12.7 & 0.1/1.0/4.9 & 0.3/2.1/8.8 \\
IID-4 (teacher) & 1.4/8.7/28.4 & 0.6/3.9/17.0 & 0.4/3.0/10.5 & 0.5/3.5/13.8 \\[3.5pt]
GROOT-4 (self) & 1.5/9.5/29.8 & 0.9/6.1/23.5 & 1.4/6.7/16.3 & 1.1/6.4/19.9 \\
GROOT-4 (teacher) & 3.2/17.4/43.7 & 2.3/12.0/30.5 & 2.1/10.1/20.9 & 2.2/11.1/25.7 \\[3.5pt]
VS-4 (self) & 2.1/12.5/34.2 & 1.3/7.7/22.9 & 1.8/8.7/18.7 & 1.6/8.2/20.8 \\
VS-4 (teacher) & \textbf{4.5}/\textbf{21.4}/\textbf{45.7} & \textbf{2.6}/\textbf{13.2}/\textbf{32.6} & \textbf{2.6}/\textbf{11.4}/\textbf{22.2} & \textbf{2.6}/\textbf{12.3}/\textbf{27.4} \\
\hline
\end{tabular}
\end{table*}

%% file: tables/code_passk_all_table.tex
\begin{table*}[t]
\centering
\footnotesize
\setlength{\aboverulesep}{0pt}\setlength{\belowrulesep}{0pt}
\setlength{\extrarowheight}{2pt}
\caption{Qwen3-4B code pass@$k$ at sampling budgets 4 and 8, and for an equal mix of the \textsc{Groot}-4 and \textsc{VS}-4 datasets (MIX-8). We report pass@1/8/64; Macro-Avg is the mean of the two held-out benchmarks, LiveCodeBench and OJBench. Bold marks the best value in each column.}
\label{tab:code_passk_all}
\begin{tabular}{l@{\hskip 8pt}c|c@{\hskip 8pt}c@{\hskip 8pt}c}
\hline
Method & Cobalt & LiveCodeBench & OJBench & Macro-Avg \\
 & $p@1/8/64$ & $p@1/8/64$ & $p@1/8/64$ & $p@1/8/64$ \\
\hline
\multicolumn{2}{l|}{\textit{Full dataset}} \\
Base & 16.0/39.4/56.7 & \textbf{34.3}/44.0/50.3 & \textbf{9.3}/14.8/18.3 & \textbf{21.8}/29.4/34.3 \\[3.5pt]
IID-4 & 15.0/37.9/55.1 & 34.0/43.4/50.0 & 8.8/14.6/18.4 & 21.4/29.0/34.2 \\
IID-8 & 15.1/38.1/55.5 & 33.9/43.3/50.1 & 9.0/14.6/18.3 & 21.4/28.9/34.2 \\
IID-64 & 16.2/39.6/55.7 & 33.9/43.7/50.8 & 9.2/14.8/19.1 & 21.5/29.2/35.0 \\[3.5pt]
GROOT-4 & 16.6/42.7/59.3 & 32.4/47.3/56.5 & 8.9/18.1/25.2 & 20.6/32.7/40.9 \\
GROOT-8 & 17.4/43.0/\textbf{61.3} & 32.4/47.1/55.6 & 9.0/17.8/24.8 & 20.7/32.5/40.2 \\[3.5pt]
VS-4 & 17.8/43.1/59.5 & 32.8/48.2/\textbf{58.5} & 9.1/\textbf{18.2}/\textbf{25.5} & 20.9/\textbf{33.2}/\textbf{42.0} \\
VS-8 & \textbf{17.9}/43.0/59.9 & 33.0/\textbf{48.5}/57.4 & 9.2/17.3/24.8 & 21.1/32.9/41.1 \\[3.5pt]
MIX-8 & 17.7/\textbf{43.3}/59.8 & 32.4/47.5/57.5 & 8.9/17.3/\textbf{25.5} & 20.6/32.4/41.5 \\
\hline\hline
\multicolumn{2}{l@{\hskip 6pt}|@{\hskip 6pt}}{\textit{Frontier ($\leq 2/64$)}} & \multicolumn{3}{l}{} \\
Base & 0.5/3.6/21.9 & 0.2/1.3/7.4 & 0.1/0.5/2.9 & 0.2/0.9/5.2 \\[3.5pt]
IID-4 & 0.7/4.9/19.8 & 0.2/1.5/7.0 & 0.1/0.7/3.5 & 0.2/1.1/5.2 \\
IID-8 & 0.7/5.1/20.8 & 0.2/1.7/7.4 & 0.1/0.6/3.2 & 0.2/1.1/5.3 \\
IID-64 & 1.0/6.3/21.5 & 0.4/2.5/8.7 & 0.1/1.0/4.4 & 0.2/1.8/6.5 \\[3.5pt]
GROOT-4 & 3.5/14.1/30.5 & 1.5/8.0/19.5 & 1.2/5.3/11.5 & 1.4/6.7/15.5 \\
GROOT-8 & 3.9/14.3/\textbf{32.4} & 1.4/7.4/17.6 & 1.2/5.4/11.7 & 1.3/6.4/14.7 \\[3.5pt]
VS-4 & 4.1/14.3/30.2 & \textbf{2.0}/\textbf{9.6}/\textbf{23.1} & \textbf{1.5}/\textbf{5.6}/11.9 & \textbf{1.8}/\textbf{7.6}/\textbf{17.5} \\
VS-8 & \textbf{4.3}/\textbf{14.6}/31.1 & \textbf{2.0}/9.3/20.7 & 1.2/4.8/11.2 & 1.6/7.1/15.9 \\[3.5pt]
MIX-8 & 3.9/14.5/30.2 & 1.6/8.3/21.0 & 1.2/4.9/\textbf{12.1} & 1.4/6.6/16.6 \\
\hline
\end{tabular}
\end{table*}

%% file: tables/filter_all_table.tex
\begin{table*}[t]
\centering
\footnotesize
\setlength{\aboverulesep}{0pt}\setlength{\belowrulesep}{0pt}
\setlength{\extrarowheight}{1pt}
\caption{Qwen3-4B competitive programming: RFT (correct responses only), SFT (all responses) and ANTI (incorrect responses only) at sampling budgets 4 and 8, trained on the same sampled datasets; IID-64 at $T{=}1.5$ is the strongest IID dataset. We report pass@1/8/64; Macro-Avg is the mean of the two held-out benchmarks, LiveCodeBench and OJBench. Bold marks the best value in each column.}
\label{tab:filters_all}
\begin{tabular}{l@{\hskip 8pt}c|c@{\hskip 8pt}c@{\hskip 8pt}c}
\hline
Method & Cobalt & LiveCodeBench & OJBench & Macro-Avg \\
 & $p@1/8/64$ & $p@1/8/64$ & $p@1/8/64$ & $p@1/8/64$ \\
\hline
\multicolumn{2}{l@{\hskip 6pt}|@{\hskip 6pt}}{\textit{Full dataset}} & \multicolumn{3}{l}{} \\
Base & 16.0/39.4/56.7 & 34.3/44.0/50.3 & 9.3/14.8/18.3 & 21.8/29.4/34.3 \\[3.5pt]
IID-4 RFT & 15.0/37.9/55.1 & 34.0/43.4/50.0 & 8.8/14.6/18.4 & 21.4/29.0/34.2 \\
IID-4 SFT & 15.3/38.6/55.0 & 33.8/43.7/50.9 & 9.3/14.9/18.8 & 21.5/29.3/34.9 \\
IID-4 ANTI & 15.5/38.7/55.1 & 34.1/43.5/49.9 & 9.0/14.6/18.9 & 21.6/29.1/34.4 \\
IID-64 RFT ($T{=}1.5$) & 17.1/41.1/57.0 & \textbf{34.4}/44.8/51.9 & \textbf{9.4}/15.4/20.1 & \textbf{21.9}/30.1/36.0 \\
IID-64 SFT ($T{=}1.5$) & 15.9/39.6/56.7 & 34.2/44.2/51.3 & 8.9/14.5/18.6 & 21.6/29.4/35.0 \\
IID-64 ANTI ($T{=}1.5$) & 15.6/40.0/57.0 & \textbf{34.4}/44.8/51.5 & 9.1/15.3/19.7 & 21.8/30.0/35.6 \\[3.5pt]
GROOT-4 RFT & 16.6/42.7/59.3 & 32.4/47.3/56.5 & 8.9/18.1/25.2 & 20.6/32.7/40.9 \\
GROOT-4 SFT & 15.6/40.7/57.8 & 31.1/45.8/55.0 & 8.5/16.2/22.0 & 19.8/31.0/38.5 \\
GROOT-4 ANTI & 15.4/40.9/57.1 & 31.2/46.1/55.1 & 8.6/16.3/22.3 & 19.9/31.2/38.7 \\
GROOT-8 RFT & 17.4/43.0/\textbf{61.3} & 32.4/47.1/55.6 & 9.0/17.8/24.8 & 20.7/32.5/40.2 \\
GROOT-8 SFT & 15.7/40.5/57.7 & 30.5/46.1/55.4 & 8.4/16.3/23.2 & 19.4/31.2/39.3 \\
GROOT-8 ANTI & 15.6/40.8/57.2 & 31.5/46.3/54.4 & 8.4/16.3/22.9 & 19.9/31.3/38.6 \\[3.5pt]
VS-4 RFT & 17.8/43.1/59.5 & 32.8/48.2/\textbf{58.5} & 9.1/\textbf{18.2}/\textbf{25.5} & 20.9/\textbf{33.2}/\textbf{42.0} \\
VS-4 SFT & 14.8/39.3/56.2 & 30.6/46.2/53.9 & 8.4/16.3/23.5 & 19.5/31.2/38.7 \\
VS-4 ANTI & 14.9/39.9/58.3 & 30.8/46.3/56.4 & 8.6/16.4/22.7 & 19.7/31.3/39.5 \\
VS-8 RFT & \textbf{17.9}/43.0/59.9 & 33.0/\textbf{48.5}/57.4 & 9.2/17.3/24.8 & 21.1/32.9/41.1 \\
VS-8 SFT & 14.7/39.3/57.7 & 30.8/46.4/56.5 & 8.4/16.5/24.8 & 19.6/31.4/40.6 \\
VS-8 ANTI & 15.7/41.1/58.3 & 31.7/47.0/54.8 & 8.7/16.9/24.7 & 20.2/31.9/39.8 \\[3.5pt]
MIX-8 RFT & 17.7/\textbf{43.3}/59.8 & 32.4/47.5/57.5 & 8.9/17.3/\textbf{25.5} & 20.6/32.4/41.5 \\
MIX-8 SFT & 15.5/41.0/59.4 & 31.4/46.7/55.3 & 8.4/16.3/23.5 & 19.9/31.5/39.4 \\
MIX-8 ANTI & 15.5/40.8/57.8 & 31.1/46.8/57.2 & 8.5/16.3/23.5 & 19.8/31.5/40.4 \\
\hline\hline
\multicolumn{2}{l@{\hskip 6pt}|@{\hskip 6pt}}{\textit{Frontier ($\leq 2/64$)}} & \multicolumn{3}{l}{} \\
Base & 0.5/3.6/21.9 & 0.2/1.3/7.4 & 0.1/0.5/2.9 & 0.2/0.9/5.2 \\[3.5pt]
IID-4 RFT & 0.7/4.9/19.8 & 0.2/1.5/7.0 & 0.1/0.7/3.5 & 0.2/1.1/5.2 \\
IID-4 SFT & 0.6/4.4/19.3 & 0.3/2.2/8.9 & 0.1/0.8/3.8 & 0.2/1.5/6.3 \\
IID-4 ANTI & 0.7/4.9/20.2 & 0.3/2.0/7.1 & 0.1/0.6/3.9 & 0.2/1.3/5.5 \\
IID-64 RFT ($T{=}1.5$) & 1.5/7.4/23.0 & 0.4/2.8/10.7 & 0.2/1.4/5.2 & 0.3/2.1/7.9 \\
IID-64 SFT ($T{=}1.5$) & 1.7/7.5/22.7 & 0.4/2.7/9.6 & 0.1/0.7/3.4 & 0.2/1.7/6.5 \\
IID-64 ANTI ($T{=}1.5$) & 1.5/7.5/23.3 & 0.4/2.9/10.2 & 0.2/1.1/4.6 & 0.3/2.0/7.4 \\[3.5pt]
GROOT-4 RFT & 3.5/14.1/30.5 & 1.5/8.0/19.5 & 1.2/5.3/11.5 & 1.4/6.7/15.5 \\
GROOT-4 SFT & 3.1/11.8/27.4 & 1.1/6.0/16.7 & 1.0/3.7/8.0 & 1.1/4.8/12.3 \\
GROOT-4 ANTI & 3.1/11.8/26.4 & 1.2/6.3/17.1 & 1.0/3.7/8.4 & 1.1/5.0/12.8 \\
GROOT-8 RFT & 3.9/14.3/\textbf{32.4} & 1.4/7.4/17.6 & 1.2/5.4/11.7 & 1.3/6.4/14.7 \\
GROOT-8 SFT & 3.4/12.9/27.9 & 1.4/7.1/17.5 & 1.0/3.9/9.1 & 1.2/5.5/13.3 \\
GROOT-8 ANTI & 3.2/12.2/26.7 & 1.2/6.4/15.6 & 0.9/3.9/9.1 & 1.1/5.2/12.3 \\[3.5pt]
VS-4 RFT & 4.1/14.3/30.2 & \textbf{2.0}/\textbf{9.6}/\textbf{23.1} & \textbf{1.5}/\textbf{5.6}/11.9 & \textbf{1.8}/\textbf{7.6}/\textbf{17.5} \\
VS-4 SFT & 3.1/11.2/25.2 & 1.2/6.9/15.0 & 0.9/3.8/10.0 & 1.1/5.3/12.5 \\
VS-4 ANTI & 3.3/12.5/29.1 & 1.2/7.0/20.3 & 0.9/3.8/9.2 & 1.1/5.4/14.8 \\
VS-8 RFT & \textbf{4.3}/\textbf{14.6}/31.1 & \textbf{2.0}/9.3/20.7 & 1.2/4.8/11.2 & 1.6/7.1/15.9 \\
VS-8 SFT & 3.0/11.0/27.0 & 1.3/7.7/20.0 & 0.9/4.0/10.9 & 1.1/5.8/15.4 \\
VS-8 ANTI & 3.6/12.7/28.8 & 1.6/7.4/16.7 & 1.0/3.9/10.9 & 1.3/5.7/13.8 \\[3.5pt]
MIX-8 RFT & 3.9/14.5/30.2 & 1.6/8.3/21.0 & 1.2/4.9/\textbf{12.1} & 1.4/6.6/16.6 \\
MIX-8 SFT & 3.3/12.4/30.8 & 1.6/7.7/17.2 & 0.8/3.9/10.0 & 1.2/5.8/13.6 \\
MIX-8 ANTI & 3.3/12.7/28.1 & 1.4/7.6/20.7 & 1.0/3.9/10.2 & 1.2/5.8/15.4 \\
\hline
\end{tabular}
\end{table*}

%% file: tables/hightemp_table.tex
\begin{table*}[t]
\centering
\footnotesize
\setlength{\aboverulesep}{0pt}\setlength{\belowrulesep}{0pt}
\setlength{\extrarowheight}{2pt}
\caption{Competitive programming. Effect of the sampling temperature used to create the Qwen3-4B training data, on the IID models and on the structured \textsc{Groot}-4 and \textsc{VS}-4 datasets. We report pass@1/8/64; Macro-Avg is the mean of the two held-out benchmarks, LiveCodeBench and OJBench. Bold marks the best value in each column.}
\label{tab:hightemp}
\begin{tabular}{l@{\hskip 8pt}c|c@{\hskip 8pt}c@{\hskip 8pt}c}
\hline
Method & Cobalt & LiveCodeBench & OJBench & Macro-Avg \\
 & $p@1/8/64$ & $p@1/8/64$ & $p@1/8/64$ & $p@1/8/64$ \\
\hline
\multicolumn{2}{l@{\hskip 6pt}|@{\hskip 6pt}}{\textit{Full dataset}} & \multicolumn{3}{l}{} \\
Base & 16.0/39.4/56.7 & 34.3/44.0/50.3 & 9.3/14.8/18.3 & 21.8/29.4/34.3 \\[3.5pt]
IID-4 & 15.0/37.9/55.1 & 34.0/43.4/50.0 & 8.8/14.6/18.4 & 21.4/29.0/34.2 \\
IID-4 ($T{=}1.5$) & 17.2/41.1/56.8 & \textbf{34.7}/44.9/52.1 & \textbf{9.6}/15.2/20.1 & \textbf{22.2}/30.0/36.1 \\
IID-8 & 15.1/38.1/55.5 & 33.9/43.3/50.1 & 9.0/14.6/18.3 & 21.4/28.9/34.2 \\
IID-8 ($T{=}1.5$) & 17.3/41.7/58.0 & \textbf{34.7}/45.4/53.7 & 9.4/15.1/19.7 & 22.1/30.2/36.7 \\
IID-64 & 16.2/39.6/55.7 & 33.9/43.7/50.8 & 9.2/14.8/19.1 & 21.5/29.2/35.0 \\
IID-64 ($T{=}1.5$) & 17.1/41.1/57.0 & 34.4/44.8/51.9 & 9.4/15.4/20.1 & 21.9/30.1/36.0 \\[3.5pt]
GROOT-4 & 16.6/42.7/59.3 & 32.4/47.3/56.5 & 8.9/18.1/25.2 & 20.6/32.7/40.9 \\
GROOT-4 (planner $T{=}1.0$) & 17.0/43.0/61.1 & 32.8/47.8/57.3 & 8.9/18.0/25.8 & 20.8/32.9/41.5 \\
GROOT-4 (planner $T{=}1.5$) & 18.6/43.4/\textbf{61.7} & 33.3/\textbf{48.4}/56.6 & 8.8/17.8/25.7 & 21.0/33.1/41.1 \\[3.5pt]
VS-4 & 17.8/43.1/59.5 & 32.8/48.2/\textbf{58.5} & 9.1/18.2/25.5 & 20.9/33.2/\textbf{42.0} \\
VS-4 (planner $T{=}1.0$) & 17.5/42.6/59.4 & 32.9/47.7/56.7 & 9.4/18.4/26.1 & 21.1/33.0/41.4 \\
VS-4 (planner $T{=}1.5$) & \textbf{18.8}/\textbf{43.8}/61.5 & 33.5/\textbf{48.4}/57.8 & \textbf{9.6}/\textbf{18.6}/\textbf{26.2} & 21.6/\textbf{33.5}/\textbf{42.0} \\
\hline\hline
\multicolumn{2}{l@{\hskip 6pt}|@{\hskip 6pt}}{\textit{Frontier ($\leq 2/64$)}} & \multicolumn{3}{l}{} \\
Base & 0.5/3.6/21.9 & 0.2/1.3/7.4 & 0.1/0.5/2.9 & 0.2/0.9/5.2 \\[3.5pt]
IID-4 & 0.7/4.9/19.8 & 0.2/1.5/7.0 & 0.1/0.7/3.5 & 0.2/1.1/5.2 \\
IID-4 ($T{=}1.5$) & 1.0/6.5/22.4 & 0.5/3.2/10.8 & 0.1/1.0/5.2 & 0.3/2.1/8.0 \\
IID-8 & 0.7/5.1/20.8 & 0.2/1.7/7.4 & 0.1/0.6/3.2 & 0.2/1.1/5.3 \\
IID-8 ($T{=}1.5$) & 1.2/7.1/24.6 & 0.6/4.0/13.8 & 0.1/1.0/4.5 & 0.3/2.5/9.2 \\
IID-64 & 1.0/6.3/21.5 & 0.4/2.5/8.7 & 0.1/1.0/4.4 & 0.2/1.8/6.5 \\
IID-64 ($T{=}1.5$) & 1.5/7.4/23.0 & 0.4/2.8/10.7 & 0.2/1.4/5.2 & 0.3/2.1/7.9 \\[3.5pt]
GROOT-4 & 3.5/14.1/30.5 & 1.5/8.0/19.5 & 1.2/5.3/11.5 & 1.4/6.7/15.5 \\
GROOT-4 (planner $T{=}1.0$) & 3.7/14.5/32.5 & 1.8/8.7/20.6 & 1.2/5.3/12.4 & 1.5/7.0/16.5 \\
GROOT-4 (planner $T{=}1.5$) & 4.2/15.1/\textbf{34.5} & 1.9/9.4/20.5 & 1.2/5.4/12.4 & 1.5/7.4/16.4 \\[3.5pt]
VS-4 & 4.1/14.3/30.2 & 2.0/\textbf{9.6}/\textbf{23.1} & \textbf{1.5}/5.6/11.9 & 1.8/\textbf{7.6}/\textbf{17.5} \\
VS-4 (planner $T{=}1.0$) & 3.9/13.8/31.0 & \textbf{2.2}/8.8/20.5 & 1.4/5.7/12.7 & 1.8/7.2/16.6 \\
VS-4 (planner $T{=}1.5$) & \textbf{4.5}/\textbf{15.7}/33.7 & \textbf{2.2}/9.1/21.5 & \textbf{1.5}/\textbf{6.0}/\textbf{12.9} & \textbf{1.9}/7.5/17.2 \\
\hline
\end{tabular}
\end{table*}

%% file: figures/fig_temp.tex
\begin{figure}[t]
\centering
\includegraphics[width=\linewidth]{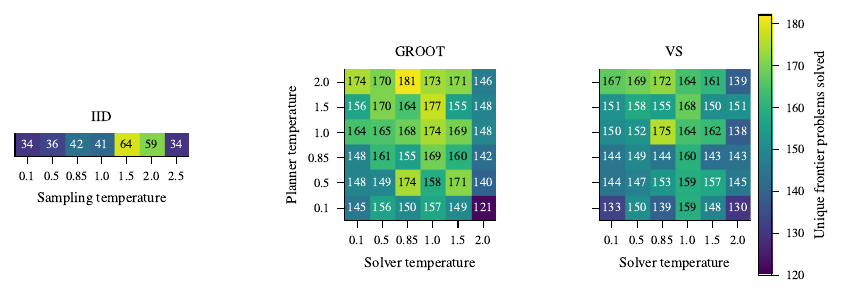}
\vspace{-.8cm}
\caption{Unique problems solved by Qwen3-4B on the 1,833-problem Cobalt training frontier at a sampling budget
of four. Left: IID sampling by temperature (mean of three runs). Right: \textsc{Groot} and
\textsc{VS} by planner and solver temperature (mean over three runs per cell).}
\label{fig:temp_grid}
\end{figure}

%% file: tables/ncp_t15_trunc_pair.tex
\begin{table}[t]
\centering
\begin{minipage}[t]{0.49\linewidth}
\vspace{0pt}
\centering
\caption{Qwen3-4B NCP coverage@16 (\%) of test sections at three perplexity-improvement thresholds, for models trained on plans sampled at $T{=}1.5$ against the standard $T{=}0.85$ models. The suffix is the sampling budget: 16 candidate plans per training section. Mean over three training runs. Bold marks the best value in each column.}
\label{tab:ncp_t15}
\input{tables/ncp_t15_body}
\end{minipage}\hfill
\begin{minipage}[t]{0.49\linewidth}
\vspace{0pt}
\centering
\caption{Truncation rates at 8k and 16k max-generated-token lengths, for Qwen-based models on Cobalt training data.}
\label{tab:teacher_truncation}
\input{tables/teacher_truncation_body}
\end{minipage}
\end{table}

%% file: tables/ncp_t15_body.tex
\footnotesize\setlength{\tabcolsep}{2pt}
\begin{tabular}{lrrr}
\toprule
Method & $5\%$ & $10\%$ & $15\%$ \\
\midrule
IID-16 & 24.40 & 7.94 & 1.60 \\
IID-16 ($T{=}1.5$) & 29.19 & 10.09 & 2.11 \\
\addlinespace[3.5pt]
\textsc{Groot}-16 & 33.84 & 12.36 & 2.76 \\
\textsc{Groot}-16 ($T{=}1.5$) & 37.36 & 14.01 & 3.13 \\
\addlinespace[3.5pt]
\textsc{VS}-16 & 37.50 & 13.74 & 3.11 \\
\textsc{VS}-16 ($T{=}1.5$) & \textbf{40.46} & \textbf{15.10} & \textbf{3.52} \\
\bottomrule
\end{tabular}

%% file: tables/teacher_truncation_body.tex
\footnotesize\setlength{\tabcolsep}{2pt}
\setlength{\aboverulesep}{0pt}\setlength{\belowrulesep}{0pt}
\setlength{\extrarowheight}{2pt}
\begin{tabular}{lcc}
\toprule
Model & 8k & 16k \\
\midrule
Base & 3.4\% & 0.2\% \\
\addlinespace[3.5pt]
IID-4 (self) & 3.2\% & 0.2\% \\
IID-4 (teacher) & 18.2\% & 7.1\% \\
\addlinespace[3.5pt]
\textsc{Groot}-4 (self) & 28.2\% & 15.4\% \\
\textsc{Groot}-4 (teacher) & 47.0\% & 32.4\% \\
\addlinespace[3.5pt]
\textsc{VS}-4 (self) & 33.2\% & 19.2\% \\
\textsc{VS}-4 (teacher) & 41.9\% & 30.6\% \\
\bottomrule
\end{tabular}

%% file: tables/teacher_16k_table.tex
\begin{table*}[t]
\centering
\footnotesize
\setlength{\aboverulesep}{0pt}\setlength{\belowrulesep}{0pt}
\setlength{\extrarowheight}{2pt}
\caption{Pass@k from teacher- versus self-generated training data. Teacher datasets are constructed with Qwen3-235B-A22B-Instruct for both the approaches and solutions. The standard self-generating procedure uses the 4B student (marked self). In both cases the trained student is Qwen3-4B-Instruct. Evaluated with a 16k-token generation limit. We report pass@1\,/\,pass@8\,/\,pass@64; Macro-Avg is the mean of the two held-out benchmarks, LiveCodeBench and OJBench. Bold marks the best value in each column.}
\label{tab:teacher16k}
\begin{tabular}{l@{\hskip 8pt}c|c@{\hskip 8pt}c@{\hskip 8pt}c}
\hline
Method & Cobalt & LiveCodeBench & OJBench & Macro-Avg \\
 & $p@1/8/64$ & $p@1/8/64$ & $p@1/8/64$ & $p@1/8/64$ \\
\hline
\multicolumn{2}{l@{\hskip 6pt}|@{\hskip 6pt}}{\textit{Full dataset}} & \multicolumn{3}{l}{} \\
IID-4 (self) & 15.1/38.3/55.5 & 34.1/43.8/50.6 & 8.8/14.3/17.8 & 21.5/29.0/34.2 \\
IID-4 (teacher) & 16.9/43.8/63.9 & 32.9/45.9/54.9 & 7.8/15.0/21.8 & 20.3/30.4/38.4 \\[3.5pt]
IID-8 (self) & 15.2/38.3/55.9 & 34.0/43.7/50.8 & 9.0/14.3/17.6 & 21.5/29.0/34.2 \\
IID-8 (teacher) & 17.9/45.5/65.2 & 32.7/46.2/55.6 & 8.0/15.2/21.7 & 20.4/30.7/38.6 \\[3.5pt]
GROOT-4 (self) & 19.8/47.1/64.9 & 34.2/49.9/59.6 & 9.9/20.0/\textbf{27.8} & 22.1/35.0/43.7 \\
GROOT-4 (teacher) & 16.6/44.6/64.7 & 28.9/47.4/57.1 & 7.9/18.6/27.4 & 18.4/33.0/42.2 \\[3.5pt]
VS-4 (self) & \textbf{21.4}/\textbf{48.4}/\textbf{65.7} & \textbf{35.0}/\textbf{51.5}/\textbf{62.8} & \textbf{10.3}/\textbf{20.2}/27.4 & \textbf{22.6}/\textbf{35.9}/\textbf{45.1} \\
VS-4 (teacher) & 15.0/43.3/64.4 & 28.3/46.3/57.5 & 7.3/17.6/26.2 & 17.8/31.9/41.9 \\
\hline\hline
\multicolumn{2}{l@{\hskip 6pt}|@{\hskip 6pt}}{\textit{Frontier ($\leq 2/64$)}} & \multicolumn{3}{l}{} \\
IID-4 (self) & 0.7/5.2/20.5 & 0.3/1.9/8.2 & 0.1/0.7/3.3 & 0.2/1.3/5.8 \\
IID-4 (teacher) & 4.6/16.7/39.3 & 1.1/6.9/19.1 & 0.3/2.1/7.6 & 0.7/4.5/13.4 \\[3.5pt]
IID-8 (self) & 0.7/5.3/21.6 & 0.3/2.0/8.7 & 0.1/0.6/2.8 & 0.2/1.3/5.8 \\
IID-8 (teacher) & 4.9/17.0/39.0 & 1.1/6.7/18.1 & 0.3/2.3/7.7 & 0.7/4.5/12.9 \\[3.5pt]
GROOT-4 (self) & 4.8/18.5/39.5 & 2.4/11.3/25.0 & 1.8/7.3/\textbf{15.2} & 2.1/9.3/20.1 \\
GROOT-4 (teacher) & 5.0/19.4/\textbf{40.3} & 1.8/9.3/20.5 & 1.4/6.7/15.0 & 1.6/8.0/17.8 \\[3.5pt]
VS-4 (self) & \textbf{5.5}/\textbf{19.6}/\textbf{40.3} & \textbf{3.1}/\textbf{13.6}/\textbf{30.8} & \textbf{2.1}/\textbf{7.6}/14.7 & \textbf{2.6}/\textbf{10.6}/\textbf{22.8} \\
VS-4 (teacher) & 4.6/18.6/39.4 & 1.8/9.4/21.6 & 1.1/6.1/13.8 & 1.5/7.8/17.7 \\
\hline
\end{tabular}
\end{table*}

%% file: tables/teacher_8k_table.tex
\begin{table*}[t]
\centering
\footnotesize
\setlength{\aboverulesep}{0pt}\setlength{\belowrulesep}{0pt}
\setlength{\extrarowheight}{2pt}
\caption{Pass@$k$ from teacher- versus self-generated training data, evaluated with an 8k-token generation limit; Table~\ref{tab:teacher16k} gives the same comparison at 16k. Teacher datasets are constructed with Qwen3-235B-A22B-Instruct for both the approaches and the solutions, while the standard self-generating procedure uses the 4B student (marked self). In both cases the trained student is Qwen3-4B-Instruct. We report pass@1/8/64 on the full test sets and on the frontier subsets; Macro-Avg is the mean of the two held-out benchmarks, LiveCodeBench and OJBench. Bold marks the best value in each column.}
\label{tab:teacher8k}
\begin{tabular}{l@{\hskip 8pt}c|c@{\hskip 8pt}c@{\hskip 8pt}c}
\hline
Method & Cobalt & LiveCodeBench & OJBench & Macro-Avg \\
 & $p@1/8/64$ & $p@1/8/64$ & $p@1/8/64$ & $p@1/8/64$ \\
\hline
\multicolumn{2}{l@{\hskip 6pt}|@{\hskip 6pt}}{\textit{Full dataset}} & \multicolumn{3}{l}{} \\
IID-4 (self) & 15.0/37.9/55.1 & \textbf{34.0}/43.4/50.0 & 8.8/14.6/18.4 & \textbf{21.4}/29.0/34.2 \\
IID-4 (teacher) & 16.0/42.0/62.3 & 32.4/45.1/54.3 & 7.5/14.4/21.1 & 19.9/29.8/37.7 \\[3.5pt]
IID-8 (self) & 15.1/38.1/55.5 & 33.9/43.3/50.1 & 9.0/14.6/18.3 & \textbf{21.4}/28.9/34.2 \\
IID-8 (teacher) & 17.0/\textbf{43.4}/\textbf{63.0} & 32.1/45.2/53.5 & 7.7/14.7/20.9 & 19.9/30.0/37.2 \\[3.5pt]
GROOT-4 (self) & 16.6/42.7/59.3 & 32.4/47.3/56.5 & 8.9/18.1/25.2 & 20.6/32.7/40.9 \\
GROOT-4 (teacher) & 11.1/32.8/53.3 & 25.6/40.3/52.3 & 5.7/12.7/20.4 & 15.7/26.5/36.3 \\[3.5pt]
VS-4 (self) & \textbf{17.8}/43.1/59.5 & 32.8/\textbf{48.2}/\textbf{58.5} & \textbf{9.1}/\textbf{18.2}/\textbf{25.5} & 20.9/\textbf{33.2}/\textbf{42.0} \\
VS-4 (teacher) & 10.4/32.3/51.6 & 25.2/39.5/51.8 & 5.6/12.6/20.7 & 15.4/26.1/36.2 \\
\hline\hline
\multicolumn{2}{l@{\hskip 6pt}|@{\hskip 6pt}}{\textit{Frontier ($\leq 2/64$)}} & \multicolumn{3}{l}{} \\
IID-4 (self) & 0.7/4.9/19.8 & 0.2/1.5/7.0 & 0.1/0.7/3.5 & 0.2/1.1/5.2 \\
IID-4 (teacher) & 4.3/15.1/\textbf{36.5} & 1.0/6.2/18.1 & 0.2/1.5/6.7 & 0.6/3.9/12.4 \\[3.5pt]
IID-8 (self) & 0.7/5.1/20.8 & 0.2/1.7/7.4 & 0.1/0.6/3.2 & 0.2/1.1/5.3 \\
IID-8 (teacher) & \textbf{4.5}/\textbf{15.3}/35.8 & 1.0/5.8/15.0 & 0.3/1.9/6.6 & 0.7/3.8/10.8 \\[3.5pt]
GROOT-4 (self) & 3.5/14.1/30.5 & 1.5/8.0/19.5 & 1.2/5.3/11.5 & 1.4/6.7/15.5 \\
GROOT-4 (teacher) & 3.3/12.4/26.6 & 0.8/4.7/14.4 & 0.5/2.7/8.0 & 0.7/3.7/11.2 \\[3.5pt]
VS-4 (self) & 4.1/14.3/30.2 & \textbf{2.0}/\textbf{9.6}/\textbf{23.1} & \textbf{1.5}/\textbf{5.6}/\textbf{11.9} & \textbf{1.8}/\textbf{7.6}/\textbf{17.5} \\
VS-4 (teacher) & 3.0/11.6/23.8 & 0.8/4.5/15.2 & 0.4/2.5/8.3 & 0.6/3.5/11.8 \\
\hline
\end{tabular}
\end{table*}

%% file: tables/postrl_all_table.tex
    \begin{table*}[t]
\centering
\footnotesize
\setlength{\aboverulesep}{0pt}\setlength{\belowrulesep}{0pt}
\setlength{\extrarowheight}{1pt}
\caption{Qwen3-4B performance before and after RL, from every RFT initialization: the base model, IID and strategic datasets at budgets 4 and 8, the high-temperature ($T{=}1.5$) IID variants, and MIX-8. Each variant's first row is the student before RL and the second after RL (mean over three RL runs). Results are given on the full test sets and on the frontier subsets. We report pass@1/8/64; Macro-Avg is the mean of the two held-out benchmarks, LiveCodeBench and OJBench. Bold marks the best value in each column.}
\label{tab:postrl_all}
\begin{tabular}{l@{\hskip 8pt}c|c@{\hskip 8pt}c@{\hskip 8pt}c}
\hline
Method & Cobalt & LiveCodeBench & OJBench & Macro-Avg \\
 & $p@1/8/64$ & $p@1/8/64$ & $p@1/8/64$ & $p@1/8/64$ \\
\hline
\multicolumn{2}{l@{\hskip 6pt}|@{\hskip 6pt}}{\textit{Full dataset}} & \multicolumn{3}{l}{} \\
Base & 16.0/39.4/56.7 & 34.3/44.0/50.3 & 9.3/14.8/18.3 & 21.8/29.4/34.3 \\
\quad +RL & 19.5/43.5/59.1 & 33.6/45.3/53.5 & 9.2/15.6/21.8 & 21.3/30.5/37.7 \\[2pt]
IID-4 & 15.0/37.9/55.1 & 34.0/43.4/50.0 & 8.8/14.6/18.4 & 21.4/29.0/34.2 \\
\quad +RL & 21.8/46.6/63.6 & \textbf{35.6}/46.8/54.1 & 9.3/15.6/20.9 & \textbf{22.5}/31.2/37.5 \\[2pt]
IID-8 & 15.1/38.1/55.5 & 33.9/43.3/50.1 & 9.0/14.6/18.3 & 21.4/28.9/34.2 \\
\quad +RL & 20.9/44.3/58.8 & 34.5/44.6/50.9 & 8.8/14.6/18.8 & 21.7/29.6/34.8 \\[2pt]
IID-4 ($T{=}1.5$) & 17.2/41.1/56.8 & 34.7/44.9/52.1 & 9.6/15.2/20.1 & 22.2/30.0/36.1 \\
\quad +RL & \textbf{22.9}/46.7/62.8 & 35.3/46.1/53.0 & 9.4/16.2/21.0 & 22.3/31.1/37.0 \\[2pt]
IID-8 ($T{=}1.5$) & 17.3/41.7/58.0 & 34.7/45.4/53.7 & 9.4/15.1/19.7 & 22.1/30.2/36.7 \\
\quad +RL & 21.0/45.5/62.5 & 35.1/47.3/55.1 & 9.4/16.2/22.1 & 22.3/31.8/38.6 \\[2pt]
IID-64 ($T{=}1.5$) & 17.1/41.1/57.0 & 34.4/44.8/51.9 & 9.4/15.4/20.1 & 21.9/30.1/36.0 \\
\quad +RL & 21.2/46.6/64.0 & 34.7/46.7/54.6 & 9.1/16.0/21.8 & 21.9/31.4/38.2 \\[2pt]
GROOT-4 & 16.6/42.7/59.3 & 32.4/47.3/56.5 & 8.9/18.1/25.2 & 20.6/32.7/40.9 \\
\quad +RL & 21.0/\textbf{47.6}/\textbf{65.2} & 34.4/48.7/\textbf{58.1} & \textbf{9.9}/\textbf{19.4}/\textbf{27.1} & 22.1/\textbf{34.0}/\textbf{42.6} \\[2pt]
GROOT-8 & 17.4/43.0/61.3 & 32.4/47.1/55.6 & 9.0/17.8/24.8 & 20.7/32.5/40.2 \\
\quad +RL & 19.7/47.0/64.5 & 33.2/47.5/56.3 & 9.2/18.1/24.7 & 21.2/32.8/40.5 \\[2pt]
VS-4 & 17.8/43.1/59.5 & 32.8/48.2/58.5 & 9.1/18.2/25.5 & 20.9/33.2/42.0 \\
\quad +RL & 21.6/47.5/64.0 & 35.2/\textbf{48.8}/57.3 & 9.8/18.3/24.9 & \textbf{22.5}/33.6/41.2 \\[2pt]
VS-8 & 17.9/43.0/59.9 & 33.0/48.5/57.4 & 9.2/17.3/24.8 & 21.1/32.9/41.1 \\
\quad +RL & 18.5/45.3/63.0 & 32.7/47.6/57.5 & 9.2/18.2/25.8 & 20.9/32.9/41.7 \\[2pt]
MIX-8 & 17.7/43.3/59.8 & 32.4/47.5/57.5 & 8.9/17.3/25.5 & 20.6/32.4/41.5 \\
\quad +RL & 20.2/46.5/63.5 & 33.8/48.2/57.7 & 9.6/18.0/25.4 & 21.7/33.1/41.5 \\
\hline\hline
\multicolumn{2}{l@{\hskip 6pt}|@{\hskip 6pt}}{\textit{Frontier ($\leq 2/64$)}} & \multicolumn{3}{l}{} \\
Base & 0.5/3.6/21.9 & 0.2/1.3/7.4 & 0.1/0.5/2.9 & 0.2/0.9/5.2 \\
\quad +RL & 4.5/13.9/29.3 & 1.1/5.9/15.3 & 0.4/2.4/7.5 & 0.8/4.1/11.4 \\[2pt]
IID-4 & 0.7/4.9/19.8 & 0.2/1.5/7.0 & 0.1/0.7/3.5 & 0.2/1.1/5.2 \\
\quad +RL & 5.5/17.3/36.6 & 1.1/6.0/15.4 & 0.4/2.0/6.4 & 0.7/4.0/10.9 \\[2pt]
IID-8 & 0.7/5.1/20.8 & 0.2/1.7/7.4 & 0.1/0.6/3.2 & 0.2/1.1/5.3 \\
\quad +RL & 5.5/15.6/29.6 & 1.3/5.4/11.8 & 0.2/1.4/4.8 & 0.8/3.4/8.3 \\[2pt]
IID-4 ($T{=}1.5$) & 1.0/6.5/22.4 & 0.5/3.2/10.8 & 0.1/1.0/5.2 & 0.3/2.1/8.0 \\
\quad +RL & 7.1/19.5/36.8 & 1.0/5.5/13.8 & 0.7/2.9/6.7 & 0.9/4.2/10.2 \\[2pt]
IID-8 ($T{=}1.5$) & 1.2/7.1/24.6 & 0.6/4.0/13.8 & 0.1/1.0/4.5 & 0.3/2.5/9.2 \\
\quad +RL & 4.7/15.1/33.8 & 1.5/7.2/16.9 & 0.5/2.8/7.7 & 1.0/5.0/12.3 \\[2pt]
IID-64 ($T{=}1.5$) & 1.5/7.4/23.0 & 0.4/2.8/10.7 & 0.2/1.4/5.2 & 0.3/2.1/7.9 \\
\quad +RL & 5.7/17.1/36.5 & 1.1/6.5/16.4 & 0.5/2.6/7.5 & 0.8/4.5/11.9 \\[2pt]
GROOT-4 & 3.5/14.1/30.5 & 1.5/8.0/19.5 & 1.2/5.3/11.5 & 1.4/6.7/15.5 \\
\quad +RL & \textbf{7.2}/\textbf{20.9}/\textbf{39.0} & \textbf{2.2}/\textbf{9.5}/\textbf{22.2} & \textbf{1.8}/\textbf{6.8}/\textbf{13.9} & \textbf{2.0}/\textbf{8.1}/\textbf{18.0} \\[2pt]
GROOT-8 & 3.9/14.3/32.4 & 1.4/7.4/17.6 & 1.2/5.4/11.7 & 1.3/6.4/14.7 \\
\quad +RL & 5.9/19.5/37.9 & 1.6/8.0/19.7 & 1.3/5.5/11.1 & 1.4/6.8/15.4 \\[2pt]
VS-4 & 4.1/14.3/30.2 & 2.0/9.6/23.1 & 1.5/5.6/11.9 & 1.8/7.6/17.5 \\
\quad +RL & 6.2/19.4/37.8 & 2.1/8.9/20.5 & 1.4/5.6/11.5 & 1.7/7.2/16.0 \\[2pt]
VS-8 & 4.3/14.6/31.1 & 2.0/9.3/20.7 & 1.2/4.8/11.2 & 1.6/7.1/15.9 \\
\quad +RL & 5.7/18.4/35.9 & 2.0/9.3/21.6 & 1.2/5.6/12.3 & 1.7/7.5/17.0 \\[2pt]
MIX-8 & 3.9/14.5/30.2 & 1.6/8.3/21.0 & 1.2/4.9/12.1 & 1.4/6.6/16.6 \\
\quad +RL & 5.8/19.0/36.5 & 2.1/9.2/21.6 & 1.5/5.7/12.0 & 1.8/7.4/16.8 \\
\hline
\end{tabular}
\end{table*}

%% file: tables/postrl_sft_anti_table.tex
\begin{table*}[t]
\centering
\footnotesize
\setlength{\aboverulesep}{0pt}\setlength{\belowrulesep}{0pt}
\setlength{\extrarowheight}{2pt}
\caption{Qwen3-4B performance before and after RL from the budget-4 SFT and ANTI initializations, the counterpart of Table~\ref{tab:postrl_all} for the other two correctness filters. Each variant's first row is the student before RL and the second after RL (mean over three RL runs). Results are given on the full test sets and on the frontier subsets. We report pass@1/8/64; Macro-Avg is the mean of the two held-out benchmarks, LiveCodeBench and OJBench. Bold marks the best value in each column.}
\label{tab:postrl_sft_anti}
\begin{tabular}{l@{\hskip 8pt}c|c@{\hskip 8pt}c@{\hskip 8pt}c}
\hline
Method & Cobalt & LiveCodeBench & OJBench & Macro-Avg \\
 & $p@1/8/64$ & $p@1/8/64$ & $p@1/8/64$ & $p@1/8/64$ \\
\hline
\multicolumn{2}{l@{\hskip 6pt}|@{\hskip 6pt}}{\textit{Full dataset}} & \multicolumn{3}{l}{} \\
IID-4 SFT & 15.3/38.6/55.0 & 33.8/43.7/50.9 & 9.3/14.9/18.8 & 21.5/29.3/34.9 \\
\quad +RL & \textbf{21.5}/\textbf{45.9}/60.9 & \textbf{36.0}/46.5/53.6 & \textbf{9.6}/16.0/21.9 & \textbf{22.8}/31.3/37.7 \\[3.5pt]
IID-4 ANTI & 15.5/38.7/55.1 & 34.1/43.5/49.9 & 9.0/14.6/18.9 & 21.6/29.1/34.4 \\
\quad +RL & 20.0/43.1/58.7 & 35.4/45.7/52.0 & 9.0/14.6/19.3 & 22.3/30.1/35.6 \\[3.5pt]
GROOT-4 SFT & 15.6/40.7/57.8 & 31.1/45.8/55.0 & 8.5/16.2/22.0 & 19.8/31.0/38.5 \\
\quad +RL & 18.8/45.2/\textbf{63.4} & 32.8/\textbf{47.4}/\textbf{56.7} & 8.6/16.4/23.0 & 20.7/31.8/\textbf{39.8} \\[3.5pt]
GROOT-4 ANTI & 15.4/40.9/57.1 & 31.2/46.1/55.1 & 8.6/16.3/22.3 & 19.9/31.2/38.7 \\
\quad +RL & 19.0/45.0/63.0 & 33.0/47.3/56.4 & 9.0/\textbf{16.8}/22.9 & 21.0/\textbf{32.1}/39.6 \\[3.5pt]
VS-4 SFT & 14.8/39.3/56.2 & 30.6/46.2/53.9 & 8.4/16.3/23.5 & 19.5/31.2/38.7 \\
\quad +RL & 15.4/40.7/59.7 & 31.9/46.6/54.7 & 8.6/16.6/\textbf{24.0} & 20.2/31.6/39.4 \\[3.5pt]
VS-4 ANTI & 14.9/39.9/58.3 & 30.8/46.3/56.4 & 8.6/16.4/22.7 & 19.7/31.3/39.5 \\
\quad +RL & 17.8/43.9/62.1 & 32.7/47.0/55.2 & 8.6/16.1/23.0 & 20.7/31.6/39.1 \\
\hline\hline
\multicolumn{2}{l@{\hskip 6pt}|@{\hskip 6pt}}{\textit{Frontier ($\leq 2/64$)}} & \multicolumn{3}{l}{} \\
IID-4 SFT & 0.6/4.4/19.3 & 0.3/2.2/8.9 & 0.1/0.8/3.8 & 0.2/1.5/6.3 \\
\quad +RL & 4.2/14.8/30.9 & 1.0/5.3/14.4 & 0.4/2.2/7.2 & 0.6/3.8/10.8 \\[3.5pt]
IID-4 ANTI & 0.7/4.9/20.2 & 0.3/2.0/7.1 & 0.1/0.6/3.9 & 0.2/1.3/5.5 \\
\quad +RL & 3.6/12.4/28.0 & 0.9/5.1/12.2 & 0.2/1.5/5.1 & 0.6/3.3/8.6 \\[3.5pt]
GROOT-4 SFT & 3.1/11.8/27.4 & 1.1/6.0/16.7 & 1.0/3.7/8.0 & 1.1/4.8/12.3 \\
\quad +RL & 5.1/17.0/35.7 & 1.7/\textbf{8.1}/\textbf{19.6} & 0.9/3.6/9.1 & 1.3/5.9/\textbf{14.4} \\[3.5pt]
GROOT-4 ANTI & 3.1/11.8/26.4 & 1.2/6.3/17.1 & 1.0/3.7/8.4 & 1.1/5.0/12.8 \\
\quad +RL & \textbf{5.8}/\textbf{18.7}/\textbf{36.3} & 1.6/8.0/19.0 & 1.0/4.3/9.4 & 1.3/\textbf{6.1}/14.2 \\[3.5pt]
VS-4 SFT & 3.1/11.2/25.2 & 1.2/6.9/15.0 & 0.9/3.8/10.0 & 1.1/5.3/12.5 \\
\quad +RL & 3.5/12.9/30.9 & 1.5/7.0/16.0 & \textbf{1.1}/\textbf{4.4}/\textbf{10.5} & 1.3/5.7/13.2 \\[3.5pt]
VS-4 ANTI & 3.3/12.5/29.1 & 1.2/7.0/20.3 & 0.9/3.8/9.2 & 1.1/5.4/14.8 \\
\quad +RL & 4.7/17.1/35.4 & \textbf{1.9}/7.9/17.6 & 0.9/3.5/9.2 & \textbf{1.4}/5.7/13.4 \\
\hline
\end{tabular}
\end{table*}

%% file: tables/native_cobalt_table.tex
\begin{table*}[t]
\centering
\footnotesize
\caption{Qwen3-4B performance on the full Cobalt test set, split by the benchmark's own difficulty tiers rather than by model-relative frontier difficulty; $n$ is the number of problems in each tier. We report pass@1/8/64. Bold marks the best value in each column.}
\label{tab:native_cobalt}
\resizebox{\linewidth}{!}{%
\begin{tabular}{l@{\hskip 12pt}c@{\hskip 10pt}c@{\hskip 10pt}c@{\hskip 10pt}c@{\hskip 10pt}c}
\toprule
Method & Easy ($n{=}104$) & Medium ($n{=}31$) & Med-hard ($n{=}47$) & Hard ($n{=}61$) & V-hard ($n{=}13$) \\
 & $p@1/8/64$ & $p@1/8/64$ & $p@1/8/64$ & $p@1/8/64$ & $p@1/8/64$ \\
\midrule
Base & 19.0/41.4/54.8 & 14.5/40.9/67.7 & 14.2/38.0/61.7 & \textbf{14.8}/38.4/55.7 & \textbf{7.2}/15.3/15.4 \\
\addlinespace[3.5pt]
IID-4 & 18.5/41.3/54.6 & 13.4/35.9/63.1 & 13.2/37.9/62.3 & 13.5/36.6/51.2 & 6.2/15.8/22.1 \\
IID-64 & 20.2/43.8/57.6 & 14.1/38.9/59.3 & 14.1/39.3/61.7 & 14.7/\textbf{38.9}/53.7 & 6.1/15.1/19.2 \\
IID-64 ($T{=}1.5$) & \textbf{22.1}/\textbf{46.6}/\textbf{58.9} & 14.3/39.3/62.2 & 14.4/37.9/59.5 & 14.5/38.8/53.3 & 6.9/16.1/21.2 \\
GROOT-4 & 20.0/43.8/57.5 & 21.8/55.3/\textbf{75.0} & 15.3/43.1/62.3 & 12.0/37.2/54.7 & 5.0/16.1/26.0 \\
VS-4 & 21.6/44.7/57.4 & \textbf{25.5}/\textbf{56.9}/68.5 & \textbf{16.9}/\textbf{44.0}/\textbf{62.7} & 12.6/37.9/\textbf{60.1} & 5.3/\textbf{17.7}/\textbf{34.2} \\
\bottomrule
\end{tabular}}
\end{table*}

%% file: tables/native_lcb_table.tex
\begin{table*}[t]
\centering
\footnotesize
\caption{Qwen3-4B performance on the full LiveCodeBench test set, split by the benchmark's own difficulty tiers rather than by model-relative frontier difficulty; $n$ is the number of problems in each tier. We report pass@1/8/64. Bold marks the best value in each column.}
\label{tab:native_lcb}
\begin{tabular}{l@{\hskip 12pt}c@{\hskip 10pt}c@{\hskip 10pt}c}
\toprule
Method & Easy ($n{=}43$) & Medium ($n{=}52$) & Hard ($n{=}80$) \\
 & $p@1/8/64$ & $p@1/8/64$ & $p@1/8/64$ \\
\midrule
Base & 86.2/95.4/97.7 & 32.3/47.2/57.7 & \textbf{7.8}/14.3/20.0 \\
\addlinespace[3.5pt]
IID-4 & 85.4/95.0/97.6 & 32.1/46.3/56.9 & 7.6/13.8/19.9 \\
IID-64 & 85.8/95.2/97.4 & 31.8/47.3/59.3 & 7.5/13.6/20.3 \\
IID-64 ($T{=}1.5$) & \textbf{86.4}/95.6/97.6 & \textbf{32.4}/48.2/60.0 & 7.6/15.3/22.0 \\
GROOT-4 & 84.1/\textbf{96.4}/\textbf{98.8} & 28.9/51.5/65.4 & 6.9/18.1/28.0 \\
VS-4 & 85.3/95.8/\textbf{98.8} & 29.0/\textbf{53.6}/\textbf{68.0} & 7.0/\textbf{19.2}/\textbf{30.6} \\
\bottomrule
\end{tabular}
\end{table*}

%% file: tables/native_ojbench_table.tex
\begin{table*}[t]
\centering
\footnotesize
\caption{Qwen3-4B performance on the full OJBench test set, split by the benchmark's own difficulty tiers rather than by model-relative frontier difficulty; $n$ is the number of problems in each tier. We report pass@1/8/64. Bold marks the best value in each column.}
\label{tab:native_ojbench}
\begin{tabular}{l@{\hskip 12pt}c@{\hskip 10pt}c@{\hskip 10pt}c}
\toprule
Method & Easy ($n{=}36$) & Medium ($n{=}79$) & Hard ($n{=}117$) \\
 & $p@1/8/64$ & $p@1/8/64$ & $p@1/8/64$ \\
\midrule
Base & 45.7/63.5/70.7 & 5.8/12.1/18.5 & 0.4/1.6/2.0 \\
\addlinespace[3.5pt]
IID-4 & 43.8/62.5/68.0 & 5.4/12.2/20.6 & 0.4/1.5/1.7 \\
IID-64 & 45.3/62.5/69.0 & 5.7/12.7/22.1 & 0.4/1.5/1.7 \\
IID-64 ($T{=}1.5$) & \textbf{45.9}/64.5/75.7 & 6.0/13.4/21.5 & \textbf{0.5}/\textbf{1.7}/2.1 \\
GROOT-4 & 41.7/70.2/\textbf{90.2} & \textbf{7.0}/19.6/29.3 & 0.2/1.0/\textbf{2.4} \\
VS-4 & 43.1/\textbf{71.0}/87.0 & \textbf{7.0}/\textbf{20.1}/\textbf{31.8} & 0.1/0.7/2.3 \\
\bottomrule
\end{tabular}
\end{table*}

%% file: tables/rsa_full_table.tex
\begin{table}[t]
\centering
\footnotesize\setlength{\tabcolsep}{4pt}
\caption{Qwen3-4B RSA pass@1 on the full test sets, before~$\rightarrow$~after ten aggregation iterations (population size 16), from the best-validation checkpoints; Table~\ref{tab:rsa} gives the same comparison on the frontier subsets. Bold marks the best final value in each column.}
\label{tab:rsa_full}
\begin{tabular}{lccc}
\toprule
Method & Cobalt & LiveCodeBench & OJBench \\
\midrule
Base & 15.9 $\rightarrow$ 29.6 & 34.2 $\rightarrow$ 39.9 & \hphantom{0}9.1 $\rightarrow$ 12.6 \\
IID-4 & 15.5 $\rightarrow$ 27.7 & 34.6 $\rightarrow$ 40.4 & \hphantom{0}9.1 $\rightarrow$ 12.5 \\
IID-64 ($T{=}1.5$) & 17.4 $\rightarrow$ 33.9 & 34.7 $\rightarrow$ 42.5 & \hphantom{0}9.2 $\rightarrow$ 13.4 \\
\addlinespace[2pt]\hdashline\addlinespace[2pt]
\textsc{Groot}-4 & 16.8 $\rightarrow$ \textbf{37.0} & 33.9 $\rightarrow$ 44.5 & \hphantom{0}9.3 $\rightarrow$ \textbf{14.1} \\
\textsc{VS}-4 & 17.9 $\rightarrow$ 35.2 & 33.4 $\rightarrow$ \textbf{45.1} & \hphantom{0}9.3 $\rightarrow$ \textbf{14.1} \\
\bottomrule
\end{tabular}
\end{table}

%% file: tables/native_rsa_cobalt_table.tex
\begin{table}[t]
\centering
\footnotesize\setlength{\tabcolsep}{4pt}
\newcolumntype{Z}{r@{\,$\rightarrow$\,}l}
\caption{Qwen3-4B RSA pass@1 on Cobalt, before~$\rightarrow$~after ten aggregation iterations (population size 16), split by the benchmark's own difficulty tiers. Bold marks the best final value in each column.}
\label{tab:native_rsa_cobalt}
\begin{tabular}{lZZZZZ}
\toprule
Method & \multicolumn{2}{c}{Easy} & \multicolumn{2}{c}{Medium} & \multicolumn{2}{c}{Med-hard} & \multicolumn{2}{c}{Hard} & \multicolumn{2}{c}{V-hard} \\
\midrule
Base & 17.9 & 35.9 & 16.0 & 27.3 & 15.1 & 27.5 & 15.4 & 25.6 & 6.2 & \textbf{12.8} \\
\addlinespace[3.5pt]
IID-4 & 19.1 & 34.5 & 14.0 & 21.2 & 13.1 & 28.4 & 14.4 & 22.7 & 5.6 & 10.1 \\
IID-64 & 19.9 & 38.7 & 17.0 & 21.3 & 13.8 & 31.3 & 14.7 & 26.2 & 6.7 & 9.1 \\
IID-64 ($T{=}1.5$) & 22.5 & \textbf{44.1} & 16.1 & 28.1 & 14.2 & 29.6 & 15.1 & 25.4 & 5.9 & 10.7 \\
GROOT-4 & 19.5 & 42.6 & 21.0 & 39.5 & 16.8 & \textbf{37.5} & 12.9 & \textbf{30.3} & 5.6 & 12.3 \\
VS-4 & 22.2 & 42.0 & 23.1 & \textbf{44.0} & 17.0 & 34.6 & 12.8 & 24.2 & 4.6 & 10.9 \\
\bottomrule
\end{tabular}
\end{table}

%% file: tables/native_rsa_pair.tex
\begin{table}[t]
\centering
\newcolumntype{Z}{r@{\,$\rightarrow$\,}l}
\begin{minipage}[t]{0.49\linewidth}
\vspace{0pt}
\centering
\caption{Qwen3-4B RSA pass@1 on LiveCodeBench, before~$\rightarrow$~after ten aggregation iterations (population size 16), split by the benchmark's own difficulty tiers. Bold marks the best final value in each column.}
\label{tab:native_rsa_lcb}
\input{tables/native_rsa_lcb_body}
\end{minipage}\hfill
\begin{minipage}[t]{0.49\linewidth}
\vspace{0pt}
\centering
\caption{Qwen3-4B RSA pass@1 on OJBench, before~$\rightarrow$~after ten aggregation iterations (population size 16), split by the benchmark's own difficulty tiers. Bold marks the best final value in each column.}
\label{tab:native_rsa_ojbench}
\input{tables/native_rsa_ojbench_body}
\end{minipage}
\end{table}

%% file: tables/native_rsa_lcb_body.tex
\footnotesize\setlength{\tabcolsep}{2pt}
\begin{tabular}{lZZZ}
\toprule
Method & \multicolumn{2}{c}{Easy} & \multicolumn{2}{c}{Medium} & \multicolumn{2}{c}{Hard} \\
\midrule
Base & 86.8 & 90.6 & 32.1 & 42.7 & 7.3 & 10.9 \\
\addlinespace[3.5pt]
IID-4 & 86.5 & 92.9 & 32.8 & 42.6 & 7.8 & 10.7 \\
IID-64 & 86.2 & 90.7 & 32.8 & 40.3 & 7.5 & 10.8 \\
IID-64 ($T{=}1.5$) & 85.6 & 91.6 & 33.0 & 45.9 & 8.5 & 13.8 \\
GROOT-4 & 85.6 & 95.6 & 31.2 & \textbf{48.3} & 7.8 & 14.5 \\
VS-4 & 85.5 & \textbf{97.0} & 30.0 & \textbf{48.3} & 7.6 & \textbf{15.1} \\
\bottomrule
\end{tabular}

%% file: tables/native_rsa_ojbench_body.tex
\footnotesize\setlength{\tabcolsep}{2pt}
\begin{tabular}{lZZZ}
\toprule
Method & \multicolumn{2}{c}{Easy} & \multicolumn{2}{c}{Medium} & \multicolumn{2}{c}{Hard} \\
\midrule
Base & 45.5 & 56.5 & 5.6 & 9.3 & 0.3 & 1.3 \\
\addlinespace[3.5pt]
IID-4 & 44.1 & 56.5 & 6.0 & 8.8 & 0.4 & 1.4 \\
IID-64 & 46.2 & \textbf{57.5} & 6.1 & 9.6 & 0.5 & 0.5 \\
IID-64 ($T{=}1.5$) & 45.6 & 56.0 & 5.7 & 11.4 & 0.4 & \textbf{1.6} \\
GROOT-4 & 42.1 & 55.2 & 7.9 & 15.6 & 0.2 & 0.5 \\
VS-4 & 42.6 & 54.6 & 7.8 & \textbf{16.0} & 0.1 & 0.3 \\
\bottomrule
\end{tabular}

%% file: figures/fig_ncp_curves.tex
\begin{figure}[t]
\centering
\includegraphics[width=\linewidth]{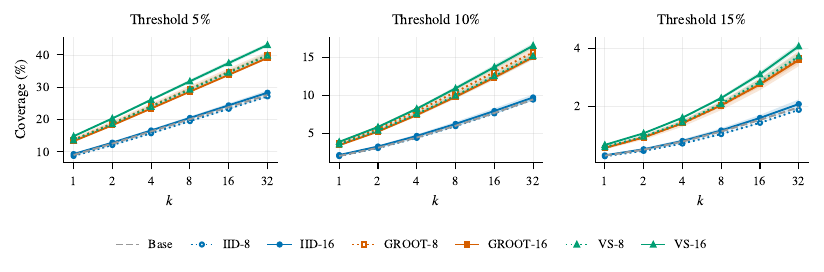}
\caption{Qwen3-4B NCP coverage of test sections at three perplexity-improvement thresholds as sampled plans $k$ increases, for the base model and
trained models. Bands are $\pm$ one standard error of the mean across the three training runs.}
\label{fig:ncp_curves}
\end{figure}

%% file: tables/ncp_v2_table.tex
\begin{table}[t]
\centering
\footnotesize
\caption{Qwen3-4B NCP coverage@32 (\%) of test sections at three perplexity-improvement thresholds, for Qwen3-4B-Instruct trained on plans clearing a 20\% threshold. The suffix is the sampling budget: 8 or 16 candidate plans per training section. Mean over three training runs (the base model is a single model). Bold marks the best value in each column.}
\label{tab:ncp}
\begin{tabular}{lrrr}
\toprule
 & \multicolumn{3}{c}{Threshold} \\
\cmidrule(lr){2-4}
Model & 5\% & 10\% & 15\% \\
\midrule
Base & 27.6 & 9.5 & 2.0 \\
\addlinespace
IID-8 & 27.2 & 9.5 & 1.9 \\
IID-16 & 28.3 & 9.7 & 2.1 \\
\addlinespace
\textsc{Groot}-8 & 39.9 & 15.7 & 3.7 \\
\textsc{Groot}-16 & 39.1 & 15.0 & 3.6 \\
\addlinespace
\textsc{VS}-8 & 39.8 & 15.2 & 3.7 \\
\textsc{VS}-16 & \textbf{43.1} & \textbf{16.6} & \textbf{4.1} \\
\bottomrule
\end{tabular}
\end{table}

%% file: tables/ncpstudents_table.tex
\begin{table}[t]
\centering
\caption{Qwen3-4B NCP coverage@1/8/32 (\%) of test sections for students trained under three
filters: RFT keeps only samples over 20\% improvement, SFT draws the same $d$ number of samples at
random, and ANTI takes the $d$ worst samples. Bold marks the best value in each column.}
\label{tab:ncpstudents}
\footnotesize
\begin{tabular}{lrrr}
\toprule
 & \multicolumn{3}{c}{Threshold} \\
\cmidrule(lr){2-4}
Model & 5\% & 10\% & 15\% \\
 & $c@1/8/32$ & $c@1/8/32$ & $c@1/8/32$ \\
\midrule
IID-16 RFT & 8.9/19.7/27.1 & 2.0/5.8/9.1 & 0.3/1.1/2.0 \\
IID-16 SFT & 8.9/19.7/27.2 & 2.1/6.0/9.1 & 0.3/1.1/2.0 \\
IID-16 ANTI & 9.0/19.9/28.0 & 2.0/6.1/9.7 & 0.3/1.1/2.0 \\
\addlinespace[2pt]\hdashline\addlinespace[2pt]
\textsc{Groot}-16 RFT & 13.6/29.1/39.5 & 3.5/10.0/15.4 & 0.6/2.1/3.9 \\
\textsc{Groot}-16 SFT & 12.5/27.5/38.0 & 3.2/9.1/14.1 & 0.5/1.8/3.2 \\
\textsc{Groot}-16 ANTI & 9.9/22.5/31.5 & 2.3/6.8/11.1 & 0.4/1.3/2.2 \\
\addlinespace[2pt]\hdashline\addlinespace[2pt]
\textsc{VS}-16 RFT & \textbf{14.6}/\textbf{31.2}/\textbf{42.2} & \textbf{3.9}/\textbf{10.8}/\textbf{16.4} & \textbf{0.7}/\textbf{2.3}/\textbf{3.9} \\
\textsc{VS}-16 SFT & 13.5/29.2/40.0 & 3.4/9.9/15.3 & 0.6/1.9/3.5 \\
\textsc{VS}-16 ANTI & 12.0/26.5/36.3 & 3.0/8.6/13.4 & 0.5/1.8/3.0 \\
\bottomrule
\end{tabular}
\end{table}

%% file: tables/ncprl_ksweep_table.tex
\begin{table}[t]
\centering
\caption{Qwen3-4B NCP coverage@1/8/32 (\%) before and after RL from each RFT initialization: the
base model, IID, and strategic datasets with budget 16. Bold marks the best value in each column.}
\label{tab:ncprl}
\footnotesize
\begin{tabular}{lrrr}
\toprule
 & \multicolumn{3}{c}{Threshold} \\
\cmidrule(lr){2-4}
Model & 5\% & 10\% & 15\% \\
 & $c@1/8/32$ & $c@1/8/32$ & $c@1/8/32$ \\
\midrule
Base & 9.0/20.1/27.6 & 2.0/6.0/9.5 & 0.3/1.1/2.0 \\
\quad +RL & 5.4/13.3/19.3 & 1.1/3.2/5.0 & 0.2/0.6/1.2 \\
\addlinespace[2pt]\hdashline\addlinespace[2pt]
IID-16 & 9.3/20.5/28.3 & 2.1/6.2/9.7 & 0.3/1.2/2.1 \\
\quad +RL & 7.7/17.9/24.9 & 1.6/4.8/7.8 & 0.2/0.9/1.6 \\
\addlinespace[2pt]\hdashline\addlinespace[2pt]
\textsc{Groot}-16 & 13.3/28.6/39.1 & 3.4/9.8/15.0 & 0.6/2.0/3.6 \\
\quad +RL & \textbf{25.1}/\textbf{45.7}/\textbf{57.3} & \textbf{8.4}/\textbf{18.8}/\textbf{26.3} & \textbf{1.9}/\textbf{5.2}/\textbf{8.1} \\
\addlinespace[2pt]\hdashline\addlinespace[2pt]
\textsc{VS}-16 & 14.8/31.8/43.1 & 3.9/10.9/16.6 & 0.7/2.3/4.1 \\
\quad +RL & 9.1/20.1/27.1 & 2.1/5.5/8.7 & 0.4/1.1/1.9 \\
\bottomrule
\end{tabular}
\end{table}

%% file: tables/ncprl_foils_table.tex
\begin{table}[t]
\centering
\caption{Qwen3-4B NCP coverage@1/8/32 (\%) of test sections after RL, requiring the arm's own plan to clear its threshold while both foil plans stay below the foil threshold. Row groups are the own threshold. Bold marks the best value in each column.}
\label{tab:ncprl_foils}
\footnotesize
\begin{tabular}{lrrr}
\toprule
 & \multicolumn{3}{c}{Foil threshold} \\
\cmidrule(lr){2-4}
Model & 5\% & 10\% & 15\% \\
 & $c@1/8/32$ & $c@1/8/32$ & $c@1/8/32$ \\
\midrule
\multicolumn{4}{l}{\textit{Own threshold 5\%}} \\
Base & 2.7/7.6/11.2 & 4.5/11.1/16.2 & 5.2/12.6/18.2 \\
IID-16 & 3.4/9.3/13.9 & 6.0/14.6/21.1 & 7.2/16.8/23.5 \\
\textsc{Groot}-16 & 5.1/13.7/\textbf{20.4} & 14.3/30.9/\textbf{41.2} & 21.5/40.8/\textbf{51.9} \\
\textsc{VS}-16 & 4.3/11.1/16.2 & 7.2/16.4/23.2 & 8.5/18.9/25.7 \\
\midrule
\multicolumn{4}{l}{\textit{Own threshold 10\%}} \\
Base & 0.6/1.7/2.9 & 1.0/2.7/4.3 & 1.1/3.1/4.9 \\
IID-16 & 0.7/2.4/4.2 & 1.3/3.9/6.5 & 1.5/4.5/7.4 \\
\textsc{Groot}-16 & 1.6/5.0/\textbf{8.1} & 4.8/11.5/\textbf{17.0} & 7.1/16.5/\textbf{23.5} \\
\textsc{VS}-16 & 1.1/3.1/4.8 & 1.8/4.6/7.1 & 2.0/5.2/8.2 \\
\midrule
\multicolumn{4}{l}{\textit{Own threshold 15\%}} \\
Base & 0.1/0.3/0.5 & 0.1/0.6/1.0 & 0.2/0.6/1.1 \\
IID-16 & 0.1/0.4/0.9 & 0.2/0.8/1.4 & 0.2/0.9/1.6 \\
\textsc{Groot}-16 & 0.3/1.0/\textbf{2.0} & 1.1/3.2/\textbf{5.1} & 1.7/4.6/\textbf{7.1} \\
\textsc{VS}-16 & 0.2/0.7/1.2 & 0.3/1.0/1.7 & 0.4/1.1/1.8 \\
\bottomrule
\end{tabular}
\end{table}

%% file: tables/hparams_table.tex
\begin{table}[t]
\centering
\small
\caption{Sampling, training, evaluation, RSA, and RL settings, used for every dataset and both base models (Qwen3-4B and Nemotron-3-Nano-4B) unless otherwise noted.}
\label{tab:hparams}
\begin{tabular}{p{0.15\linewidth}p{0.45\linewidth}p{0.3\linewidth}}
\toprule
 & Setting & Value \\
\midrule
Sampling (code) & default budget per problem & 4 ($1\times4$) \\
 & planner $T$ / top-$p$ / max tokens & 0.45 / 0.95 / 8192 \\
 & solver and IID $T$ / top-$p$ / max tokens & 0.85 / 0.95 / 8192 \\
 & high-temperature IID & $T=1.5$ \\
\addlinespace
Sampling (NCP) & default budget per section & 16 ($2\times8$) \\
 & planner and solver $T$ & 0.85 \\
\addlinespace
SFT / RFT & LoRA rank / $\alpha$ / targets & 128 / 32 / all linear layers \\
 & Nemotron LoRA targets & q, k, v, o, up, down, in projections \\
  & learning rate / schedule / KL & $10^{-6}$ / constant / none \\
 & episodes & 6 (code) / 2 (NCP) \\
 & epochs / sequences per step / max length & 4 / 48 / 8192 \\
 & NCP: sequences per step / max length & 16 / 51200 \\
 & checkpoint selection & lowest validation loss, every quarter epoch \\
\addlinespace
Evaluation & samples per problem & 128 \\
 & $T$ / top-$p$ / max tokens & 0.85 / 0.95 / 8192 (16k: 16384) \\
\addlinespace
RSA & population $N$ / aggregation size $K$ / steps $T$ & 16 / 4 / 10 (\citet{venkatraman2026recursive}'s defaults) \\
 & $T$ / max tokens / context & 1.0 / 8192 / 65536 \\
\addlinespace
RL (MaxRL for Code, GRPO for NCP) & prompts per step / samples per prompt / batch & 128 / 8 / 128 \\
 & learning rate / KL / episodes & $10^{-6}$ constant / none / 6 \\
 & clip range / overlong buffer / penalty & 0.2--0.28 / 1024 tokens / 0.25 \\
 & rollout $T$ / max new tokens & 1.0 / 8192 \\
  & checkpoint selection & code: best validation pass@8, every 4 steps; NCP: validation coverage@8 at own $>$10\%, foils $<$5\%, every 8 steps \\
\bottomrule
\end{tabular}
\end{table}

%% file: sections/appendix_prompts.tex
\subsection{Prompts}
\label{app:prompts}
\providecolor{deeppurple}{HTML}{4B0082}

All prompts are a single user message with no system prompt; \{PROBLEM\}, \{APPROACH\},
\{STORY INFORMATION\}, and \{DIRECTION\} mark the inserted text. IID sampling uses the solver
prompt without the approach directions.

The RSA aggregation prompt below is used unchanged in both domains; \{CANDIDATES\} marks the
$K$ sampled members of the previous population, inserted verbatim and untruncated.

\begin{tcolorbox}[breakable, colback=white, colframe=deeppurple, colbacktitle=deeppurple, coltitle=white, halign title=flush center, title=\textbf{Code: \textsc{Groot} planner}]
\small
You will be given a programming problem.

\medskip\par
PROBLEM:\par\noindent \{PROBLEM\}

\medskip\par
First, write out a decision tree for solving this problem inside \textless{}tree\textgreater{}\textless{}/tree\textgreater{} tags. Here is an example of the expected shape, written for a different, unrelated problem (find the k-th smallest element in an array) - use the format but not the content of this example:

\medskip\par
A. Sort the whole array, index into the result\par\noindent \hspace*{1.5em}A1. Sorting order\par\noindent \hspace*{3.5em}A1a. Ascending, answer at index k-1\par\noindent \hspace*{3.5em}A1b. Descending, answer at index n-k\par\noindent B. Maintain a bounded heap while scanning\par\noindent \hspace*{1.5em}B1. What the heap holds\par\noindent \hspace*{3.5em}B1a. Max-heap of the k smallest seen so far - pop when size exceeds k\par\noindent \hspace*{3.5em}B1b. Min-heap of everything - pop k times at the end\par\noindent C. Quickselect - partition, recurse into one side\par\noindent \hspace*{1.5em}C1. Pivot choice\par\noindent \hspace*{3.5em}C1a. Random pivot\par\noindent \hspace*{3.5em}C1b. Median-of-three\par\noindent D. Binary search on the answer value\par\noindent \hspace*{1.5em}D1. The predicate: how many elements are \textless{}= mid\par\noindent \hspace*{3.5em}D1a. Recount with a linear scan per step\par\noindent \hspace*{3.5em}D1b. Sort once, count by bisection per step

\medskip\par
The top-level entries of this tree should be fundamentally different strategies that seem like reasonable high-level approaches to solve this problem. Keep the tree high-level and compact, only briefly describing key decisions that meaningfully change results and are also reasonable alternatives. Obvious choices with no alternatives can be skipped over to keep the trees concise. Every root-to-leaf path should read as a coherent, distinct solution plan.

\medskip\par
Then pick the 4 root-to-leaf paths through the tree that best combine coverage of different strategies with likelihood of success. The 4 paths do not need to share any part of the tree - four approaches from four entirely different subtrees, with no overlap at all, are valid and often ideal. Immediately before each \textless{}approach\textgreater{} block, on its own line outside the tags, write the path it takes in the style "Path: B -\textgreater{} B1a". Inside the tags give the full approach summary itself: name the strategy and its one decisive insight; if the path relies on a formula, closed form, or feasibility rule, check it against each provided example and only commit if it reproduces the expected outputs - do this check briefly and state only its conclusion, do not write out step-by-step computations; pin down any output-format or tie-break rule the examples imply; and otherwise leave implementation details open for the solver. Fully self-contained prose that does not mention the tree, the labels, or the other approaches. Each approach must be a prose description of the reasoning and the plan - do NOT include code or code blocks in the approaches.

\medskip\par
Format your entire response exactly as:

\medskip\par
\textless{}tree\textgreater{}\par\noindent (the tree)\par\noindent \textless{}/tree\textgreater{}

\medskip\par
Path: ...\par\noindent \textless{}approach\textgreater{}\par\noindent (the approach summary)\par\noindent \textless{}/approach\textgreater{}

\medskip\par
(three more \textless{}approach\textgreater{} blocks, 4 in total - no other text before, between, or after)

\medskip\par
Do all of your thinking inside the tree and the approach blocks - if you are unsure, commit to your best guess rather than reasoning outside them.
\end{tcolorbox}

\begin{tcolorbox}[breakable, colback=white, colframe=deeppurple, colbacktitle=deeppurple, coltitle=white, halign title=flush center, title=\textbf{Code: \textsc{VS} planner}]
\small
You will be given a programming problem.

\medskip\par
PROBLEM:\par\noindent \{PROBLEM\}

\medskip\par
Generate 4 high-level approaches to solving the provided problem, along with their probability (the estimated probability from 0.0 to 1.0 of this approach given the problem relative to the full distribution). Randomly sample the approaches from the full distribution.

\medskip\par
For each of the 4 approaches:

\medskip\par
Write each approach as a complete summary of the full reasoning that leads to it - not just the change. Walk from the problem's structure through the key decisions to a working solution, with this approach's distinctive decision (the change that makes it different) incorporated as the path you take. Make clear what that change is and why it works, and cover the core idea and the main steps. Give enough detail that a solver could reconstruct the entire line of reasoning and implement it without any further figuring-out. Aim for roughly 10-15 sentences.

\medskip\par
Each approach must be a prose description of the reasoning and the plan - do NOT include code or code blocks in the approaches.

\medskip\par
Put each approach inside its own \textless{}approach\textgreater{}\textless{}/approach\textgreater{} tags. Begin each approach with a line "Probability: \textless{}p\textgreater{}" giving its probability from 0.0 to 1.0, followed by the approach summary.
\end{tcolorbox}

\begin{tcolorbox}[breakable, colback=white, colframe=deeppurple, colbacktitle=deeppurple, coltitle=white, halign title=flush center, title=\textbf{Code: solver (hidden approach)}]
\small
You are tasked with solving the programming problem below, using the given approach as the hidden basis for your reasoning:

\medskip\par
PROBLEM:\par\noindent \{PROBLEM\}

\medskip\par
Below is the hidden approach you will use:

\medskip\par
\{APPROACH\}

\medskip\par
Your response should contain two things: concise reasoning about how to solve the problem and a complete program. The reasoning should start from the beginning, concisely deriving the hidden approach as if you arrived at it yourself from the problem. However, you are explicitly not allowed to reference the fact you were given an approach in any way, e.g. saying 'based on the hint' or 'following the given approach'. Instead, your reasoning and program must be presented as your own work and as if you came up with it on your own. Every word is being monitored, a reader should not be able to tell that you had seen a hint based on your reasoning or answer. Your final answer must closely align with the given approach, although small obvious errors can be fixed if you encounter them.

\medskip\par
After reasoning through the problem concisely in prose, write the complete program in a single closed ```python``` block.
\end{tcolorbox}

\begin{tcolorbox}[breakable, colback=white, colframe=deeppurple, colbacktitle=deeppurple, coltitle=white, halign title=flush center, title=\textbf{RSA: aggregation}]
\small
You are given a coding problem and several candidate solutions. Some candidates may be incorrect or contain errors. Aggregate the useful ideas and produce a single, high-quality solution. Reason carefully; if candidates disagree, choose the correct path. If all are incorrect, then attempt a different strategy. End with the final result in ```python``` fences.

\medskip\par
Problem:

\medskip\par
\{PROBLEM\}

\medskip\par
Candidate solutions (may contain mistakes):

\medskip\par
---- Solution 1 ----\par\noindent \{CANDIDATE\}

\medskip\par
---- Solution 2 ----\par\noindent \{CANDIDATE\}

\medskip\par
\textit{($K$ candidates in total, each inserted verbatim and untruncated.)}

\medskip\par
Now write a single improved solution. Provide clear reasoning and end with the final answer in ```python``` fences.
\end{tcolorbox}

When a population member is aggregated on its own ($K{=}1$) the wording changes to refer to a single
candidate and asks the model to refine that trajectory rather than combine several. Outside the code
domain the problem is named accordingly and the answer format becomes \verb|\boxed{}| for maths or
\verb|<answer>| tags for reasoning tasks; nothing else differs.

\begin{tcolorbox}[breakable, colback=white, colframe=deeppurple, colbacktitle=deeppurple, coltitle=white, halign title=flush center, title=\textbf{NCP: \textsc{Groot} planner (stage 1)}]
\small
You are an expert novelist's planning assistant. Produce author-facing planning documents, never drafted narration or exact dialogue.

\medskip\par
You are planning the next section of a novel in progress. You will be given the story so far, character sheets, the chapter you are in up to the exact point where this section begins, and a summary of what this section covers.

\medskip\par
\#\# What to produce

\medskip\par
You are working out how the section below could be written, for a planner who will take one of your directions and write the plan from it.

\medskip\par
\{STORY INFORMATION: the story so far, character sheets, the current chapter up to the point where the section begins, the per-section guidance, and the target length\}

\medskip\par
\# How to answer

\medskip\par
First, write out a decision tree of the ways this section could be written, inside \textless{}tree\textgreater{}\textless{}/tree\textgreater{} tags. Here is an example of the expected shape. It is written for a different, unrelated section, in which an estranged sibling returns to the family house before its sale. Take the format from it, not the content:

\medskip\par
A. The section is a confrontation over the sale itself\par\noindent \hspace*{1.5em}A1. Who forces the issue\par\noindent \hspace*{3.5em}A1a. The returning sibling opens the attack on arrival\par\noindent \hspace*{3.5em}A1b. The resident sibling gets there first, the papers already signed\par\noindent \hspace*{6.0em}A1b1. ...\par\noindent \hspace*{1.5em}A2. Where it lands\par\noindent \hspace*{3.5em}A2a. Open rupture, one of them leaves before the section ends\par\noindent \hspace*{3.5em}A2b. Forced truce, the argument buried under logistics\par\noindent B. The section is an excavation of the house, the sale barely spoken\par\noindent \hspace*{1.5em}B1. What the house yields\par\noindent \hspace*{3.5em}B1a. An object that reopens an old injury, a letter or a photograph\par\noindent \hspace*{3.5em}B1b. An absence, something expected is already gone\par\noindent \hspace*{1.5em}B2. Narrative distance\par\noindent \hspace*{3.5em}B2a. Close interiority, grief in present tense\par\noindent \hspace*{3.5em}B2b. Cool inventory taking, feeling held off the page\par\noindent \hspace*{1.5em}...\par\noindent C. The section is told sideways, through a third party\par\noindent \hspace*{1.5em}C1. ...\par\noindent D. ...

\medskip\par
Your real tree should be more detailed than this sketch, but keep each entry to a brief phrase. The tree is a map, not an essay. The top level entries should be fundamentally different ways the section could go, different in what happens, who drives it, or how it is told. Under each one, include only the decisions that actually arise on that way of going, the places where a writer could plausibly commit to more than one thing, and skip choices that have no real alternative. Every root to leaf path should read as a coherent, distinct approach for the section, and none may contradict the summary of this passage.

\medskip\par
Then pick the 8 root to leaf paths that best combine coverage of fundamentally different ways with plausibility given the story information. The paths do not need to share any part of the tree. 8 approaches from 8 different subtrees are valid and often ideal. Immediately before each \textless{}approach\textgreater{} block, on its own line outside the tags, write the path it takes in the style "Path: B -\textgreater{} B1a". Inside the tags, write the approach itself.

\medskip\par
An approach is a brief for a planner, not a plan. For each approach, give: what it commits to and what it rules out; the exact phrases from the story information it rests on, quoted; what a writer following it would still have to settle; and what would show it to be the wrong reading. Ground every approach in the story information rather than in invention - the earlier chapters and character sheets hold the unfinished business this passage is likely to use. Do not mention the tree, the labels, or the other approaches.

\medskip\par
Do not write the plan itself, and do not draft narration or dialogue. Format your entire response exactly as:

\medskip\par
\textless{}tree\textgreater{}\par\noindent (the tree)\par\noindent \textless{}/tree\textgreater{}

\medskip\par
Path: ...\par\noindent \textless{}approach\textgreater{}\par\noindent (the approach)\par\noindent \textless{}/approach\textgreater{}

\medskip\par
(7 more Path lines and \textless{}approach\textgreater{} blocks, 8 in total, with no other text before, between, or after)
\end{tcolorbox}

\begin{tcolorbox}[breakable, colback=white, colframe=deeppurple, colbacktitle=deeppurple, coltitle=white, halign title=flush center, title=\textbf{NCP: \textsc{VS} planner (stage 1)}]
\small
You are an expert novelist's planning assistant. Produce author-facing planning documents, never drafted narration or exact dialogue.

\medskip\par
You are planning the next section of a novel in progress. You will be given the story so far, character sheets, the chapter you are in up to the exact point where this section begins, and a summary of what this section covers.

\medskip\par
\#\# What to produce

\medskip\par
You are working out how the section below could be written, for a planner who will take one of your directions and write the plan from it.

\medskip\par
\{STORY INFORMATION: the story so far, character sheets, the current chapter up to the point where the section begins, the per-section guidance, and the target length\}

\medskip\par
\# How to answer

\medskip\par
Sample 8 different ways this section could go, along with their probability (the estimated probability from 0.0 to 1.0 of this approach given the story information relative to the full distribution), spanning the range of what could plausibly happen rather than several readings of the obvious one. These are approaches for someone else to plan from, not plans.

\medskip\par
An approach is a brief for a planner, not a plan. For each approach, give: what it commits to and what it rules out; the exact phrases from the story information it rests on, quoted; what a writer following it would still have to settle; and what would show it to be the wrong reading. Ground every approach in the story information rather than in invention - the earlier chapters and character sheets hold the unfinished business this passage is likely to use.

\medskip\par
The approaches must differ in what happens, who drives it, or how it is told - not cosmetically - and none may contradict the summary of this passage.

\medskip\par
Do not write the plan itself, and do not draft narration or dialogue. Put each approach in its own \textless{}approach\textgreater{}...\textless{}/approach\textgreater{} block, 8 in total. Begin each approach with a line "Probability: \textless{}p\textgreater{}" giving its probability from 0.0 to 1.0, followed by the approach itself.
\end{tcolorbox}

\begin{tcolorbox}[breakable, colback=white, colframe=deeppurple, colbacktitle=deeppurple, coltitle=white, halign title=flush center, title=\textbf{NCP: solver (stage 2, hidden approach)}]
\small
You are an expert novelist's planning assistant. Produce author-facing planning documents, never drafted narration or exact dialogue.

\medskip\par
You are planning the next section of a novel in progress. You will be given the story so far, character sheets, the chapter you are in up to the exact point where this section begins, and a summary of what this section covers.

\medskip\par
\#\# What to produce

\medskip\par
You are planning the section below, using the approach given here as the hidden basis for your reasoning.

\medskip\par
APPROACH\par\noindent \{APPROACH\}

\medskip\par
Reach this approach's conclusion by your own reading of the story information: check the phrases it quotes against what is actually there, settle what it leaves open, and correct anything in it that the story information contradicts. Keep the approach it is going in rather than substituting one of your own.

\medskip\par
Your whole response must stand on its own as planning done from the story information. A reader given only the story information and your response must not be able to tell that a approach was supplied: never mention it, quote it as someone else's, or say that a claim checked out. Reason in the first person about the section itself.

\medskip\par
Write each plan as an organized paragraph of 100 to 300 words covering how the section should be written: the events in order from the exact continuation point to the stated endpoint, who drives them and how they behave, what the dialogue is doing and what ground it covers, what the reader learns or is kept from, and the pace.

\medskip\par
Draw on the whole story information, not only the prose immediately before this section. The earlier chapters and the character sheets establish objects, people and unresolved business that this passage is likely to reach for.

\medskip\par
Name people, places and things specifically rather than writing "the forest", "someone" or "the object". Name the ones this passage will actually use: the people already present or named in the guidance, the place the prose has just reached, the objects and unfinished business the preceding pages left standing. Do not list other characters or events from elsewhere in the book merely to be specific.

\medskip\par
Give the register concretely: the kind of diction, the sentence rhythm, whether the dialogue is clipped or discursive. Do not name an abstract mood and leave it at that.

\medskip\par
Focus the plan on new material: what this passage adds that the reader has not already been given. Do not describe anything already in the supplied prose, and do not spend words on events that will not happen.

\medskip\par
Preserve the story's established facts, the required development, who speaks, the order, and the endpoint.

\medskip\par
\{STORY INFORMATION: the story so far, character sheets, the current chapter up to the point where the section begins, the per-section guidance, and the target length\}

\medskip\par
\# How to answer

\medskip\par
Two things, in this order: your reasoning, then the plan.

\medskip\par
First, think in writing, working from the story information above. Cover what the preceding pages left unfinished, who is present and what each of them wants at this exact moment, which specific people, places and objects already in play this passage will reach for, and what the summary of this passage requires. Quote the exact phrases you are relying on, so the reasoning is checkable. Weigh at least one alternative and say why you are not taking it. At least 150 words of this, as ordinary prose.

\medskip\par
Then give the plan inside a single \textless{}writing\_plan\textgreater{}...your plan...\textless{}/writing\_plan\textgreater{} block. The reasoning goes outside the block; the plan alone goes inside it.

\medskip\par
The plan is a brief for the writer, not the passage itself. Write it as an account of what the section does: the events in order, who drives them and how they behave, what the dialogue covers, what the reader learns or is kept from, the pace, and the register. Saying where the scene is set, what is in the room, and what the weather or light is doing is exactly what a plan is for. Writing the sentences a reader would read is not: no drafted narration, and no lines of dialogue. Refer to the characters in the third person and say what they do, rather than staging it on the page.

\medskip\par
The approximate word count given above is the length of the finished passage a writer will produce from this plan. It is not a target for the plan itself.
\end{tcolorbox}

%% file: sections/appendix_examples.tex
\subsection{Examples}
\label{app:examples}
\providecolor{exgrey}{HTML}{6B7280}
\providecolor{exiid}{HTML}{0072B2}
\providecolor{exgroot}{HTML}{D55E00}
\providecolor{exvs}{HTML}{009E73}
\providecolor{annot}{HTML}{8A6512}
\lstset{basicstyle=\ttfamily\scriptsize, breaklines=true, columns=fullflexible, keepspaces=true, xleftmargin=0pt, frame=none, aboveskip=2pt, belowskip=2pt}

One mined problem from each domain. Everything inside the boxes is the model's own output except the text in \textit{\color{annot}(italic parentheses)}, which is our annotation.

\subsubsection*{Competitive programming}

\begin{tcolorbox}[breakable, colback=white, colframe=exgrey, colbacktitle=exgrey,
  coltitle=white, halign title=flush center, title=\textbf{Problem: Cobalt 14164}]
\footnotesize
In Takahashi Kingdom, there is a east-west railroad and N cities along it, numbered 1, 2, 3, ..., N from west to east.
A company called AtCoder Express possesses M trains, and the train i runs from City L\_i to City R\_i (it is possible that L\_i = R\_i).
Takahashi the king is interested in the following Q matters:
 - The number of the trains that runs strictly within the section from City p\_i to City q\_i, that is, the number of trains j such that p\_i $\leq$ L\_j and R\_j $\leq$ q\_i.
Although he is genius, this is too much data to process by himself. Find the answer for each of these Q queries to help him.
\end{tcolorbox}

\begin{tcolorbox}[breakable, colback=white, colframe=exgroot, colbacktitle=exgroot,
  coltitle=white, halign title=flush center, title=\textbf{\textsc{Groot}}]
\footnotesize
\textit{\color{annot}The tree:}
\medskip\par
\begin{lstlisting}
A. Direct query processing with precomputation
   A1. Precompute all train intervals and store in a list
   A2. For each query, iterate over all trains and check if L_j >= p_i and R_j <= q_i
B. Coordinate compression with 2D range queries
   B1. Map city coordinates to compressed indices
   B2. Use offline queries with sweep line and segment trees
C. Offline processing with sorting and event-based queries
   C1. Sort trains by left endpoint and queries by right endpoint
   C2. Process events in order using a data structure to count valid trains
D. Binary search on answer with feasibility checking
   D1. For each query, binary search on number of trains
   D2. Check feasibility by counting trains within [p_i, q_i] using interval overlap
\end{lstlisting}
\medskip\par
\textbf{Path: A $\rightarrow$ A1} \textit{\color{annot}(correct)}
\par\nopagebreak
Directly process each query by iterating through all trains and checking whether each train's interval [L\_j, R\_j] lies entirely within the query interval [p\_i, q\_i]. Since the constraints allow up to 200,000 trains and 100,000 queries, and each train check is O(1), the total complexity is O(M * Q), which may be too slow for worst-case inputs. However, given the small value of N ($\leq$ 500), and that each train interval is bounded by cities, this straightforward approach is feasible only if M and Q are small. But in worst-case M=200,000 and Q=100,000, this would result in 20 billion operations, which is too slow. Thus, this approach is only viable if optimized or if constraints were tighter. However, the sample input (small M and Q) suggests it might work in practice, but it fails in worst-case scalability.
\medskip\par
\textbf{Path: C $\rightarrow$ C1} \textit{\color{annot}(correct)}
\par\nopagebreak
Sort all trains by their left endpoint L\_i. For each query, sort queries by their right endpoint q\_i. Then process queries in increasing order of q\_i, and for each query, add all trains whose left endpoint is $\leq$ current query's q\_i and whose right endpoint is $\geq$ current query's p\_i. Use a data structure (like a balanced BST or sorted list) to maintain active trains and efficiently count those with R\_j $\leq$ q\_i. This approach uses offline processing and event-based filtering, reducing the complexity to O((M + Q) log M) with proper implementation. The key insight is that as we process queries in order of increasing q\_i, we can add trains incrementally and maintain only those that are relevant, reducing redundant checks.
\medskip\par
\textbf{Path: B $\rightarrow$ B1} \textit{\color{annot}(fails)}
\par\nopagebreak
Apply coordinate compression to map city positions to a smaller index space, then use a 2D range query structure (like a segment tree with coordinate compression) to answer each query. Each train is represented as a range [L\_i, R\_i], and each query [p\_i, q\_i] asks for the number of trains fully contained in that range. The feasibility of this approach relies on the fact that N is small ($\leq$ 500), so coordinate compression is trivial. However, building a 2D segment tree for up to 200,000 intervals and 100,000 queries is overkill and complex. Moreover, the range queries here are not standard 2D range sum queries; they are "count of intervals fully inside a query interval". This approach is overly complex for the given constraints and does not offer a clear advantage over simpler methods.
\medskip\par
\textbf{Path: D $\rightarrow$ D2} \textit{\color{annot}(correct)}
\par\nopagebreak
For each query, use binary search on the number of trains that lie within [p\_i, q\_i], but this is not directly applicable because the answer is not a value to be searched over. Instead, feasibility checking would involve counting how many trains satisfy p\_i $\leq$ L\_j and R\_j $\leq$ q\_i. This is a direct count, not a search. Thus, binary searching over the answer does not make sense here. The feasibility function would be trivial: for a given candidate count, check if that many trains satisfy the condition. But since the count is not a continuous variable and the condition is monotonic in no meaningful way, this approach is invalid and misapplies binary search. Therefore, it fails to produce a correct solution.
\medskip\par
\textit{\color{annot}The program for the first correct path:}
\medskip\par
\begin{lstlisting}
def main():
    import sys
    input = sys.stdin.read
    data = input().split()
    
    # Read N, M, Q
    idx = 0
    N = int(data[idx])
    M = int(data[idx+1])
    Q = int(data[idx+2])
    idx += 3
    
    # Read M trains
    trains = []
    for _ in range(M):
        l = int(data[idx])
        r = int(data[idx+1])
        idx += 2
        trains.append((l, r))
    
    # Read Q queries
    queries = []
    for _ in range(Q):
        p = int(data[idx])
        q = int(data[idx+1])
        idx += 2
        queries.append((p, q, _))  # store original index
    
    # Sort trains by left endpoint in descending order
    trains.sort(key=lambda x: x[0], reverse=True)
    
    # Sort queries by p in descending order, preserving original index
    queries.sort(key=lambda x: x[0], reverse=True)
    
    # Fenwick tree (Binary Indexed Tree) for range sum queries
    class FenwickTree:
        def __init__(self, size):
            self.n = size
            self.tree = [0] * (self.n + 1)
        
        def update(self, i, delta):
            while i <= self.n:
                self.tree[i] += delta
                i += i & (-i)
        
        def query(self, i):
            s = 0
            while i > 0:
                s += self.tree[i]
                i -= i & (-i)
            return s
    
    fenw = FenwickTree(N)
    
    # Process queries in descending order of p
    results = [0] * Q
    train_ptr = 0
    
    for p, q, orig_idx in queries:
        # Add all trains with L >= p (and not yet added)
        while train_ptr < M and trains[train_ptr][0] >= p:
            l, r = trains[train_ptr]
            fenw.update(r, 1)
            train_ptr += 1
        
        # Query the number of trains with R <= q
        results[orig_idx] = fenw.query(q)
    
    # Output results in original query order
    for res in results:
        print(res)

if __name__ == "__main__":
    main()
\end{lstlisting}
\medskip\par
\textit{\color{annot}3 out of 4 correct.}
\end{tcolorbox}

\begin{tcolorbox}[breakable, colback=white, colframe=exvs, colbacktitle=exvs,
  coltitle=white, halign title=flush center, title=\textbf{\textsc{VS}}]
\footnotesize
\textit{\color{annot}The approaches, with their stated probabilities:}
\medskip\par
\textbf{$p = 0.95$} \textit{\color{annot}(fails)}
\par\nopagebreak
The problem requires counting the number of trains that run entirely within a given interval [p\_i, q\_i], meaning each train j must satisfy p\_i $\leq$ L\_j and R\_j $\leq$ q\_i. A straightforward approach is to iterate over each query and, for each train, check whether it lies completely within the query interval. Given that Q can be up to 100,000 and M up to 200,000, a naive O(Q$\times$M) solution would be too slow in the worst case. However, since N is only up to 500, we can consider preprocessing the trains by their left and right endpoints. For each possible left endpoint, we can store all trains with that left endpoint and their right endpoints. Then, for each query [p\_i, q\_i], we can iterate through all left endpoints from p\_i to q\_i and count how many trains with that left endpoint have right endpoints $\leq$ q\_i. This reduces the number of checks per query to at most O(N), and with preprocessing, the total complexity becomes O(M$\times$N + Q$\times$N), which is acceptable given the small value of N. The key insight is that instead of checking every train for every query, we group trains by their left endpoint and use the bounded range of N to limit the number of endpoints we need to consider. This approach efficiently handles the constraints due to the small value of N.
\medskip\par
\textbf{$p = 0.10$} \textit{\color{annot}(correct)}
\par\nopagebreak
We can use coordinate compression and offline queries with a sweep line technique. The idea is to treat each train as a segment [L\_i, R\_i] and each query as a range [p\_i, q\_i]. We want to count how many segments are fully contained within each query. Instead of checking each train per query, we can process all queries and all segments together. We create events: for each train, we add a "start" event at L\_i and an "end" event at R\_i + 1. For each query, we want to count the number of segments that start at or after p\_i and end at or before q\_i. We can process all events in sorted order and maintain a data structure that tracks active segments. However, since we need to answer queries for specific intervals, we can use a sweep line that goes from left to right and maintains a list of active segments. For each query, we can determine the count by checking the state of the sweep line at position p\_i and then advancing to q\_i, counting how many segments are active throughout that interval. The challenge is that segments that start before p\_i and end after q\_i are irrelevant. This method requires sorting events and processing them in order, then using a Fenwick tree or segment tree to maintain counts. However, due to the small value of N, a simpler method is preferred. This approach is more complex and overkill for the constraints, so it is less likely to succeed in practice.
\medskip\par
\textbf{$p = 0.20$} \textit{\color{annot}(fails)}
\par\nopagebreak
We can precompute for each possible interval [l, r] (where 1 $\leq$ l $\leq$ r $\leq$ N) the number of trains that run entirely within it. Since N is at most 500, the total number of intervals is O(N$^2$) = about 125,000, which is manageable. For each interval [l, r], we iterate over all M trains and check if L\_i $\geq$ l and R\_i $\leq$ r. We store the count in a 2D array `count[l][r]`. Then, for each query [p\_i, q\_i], we simply look up `count[p\_i][q\_i]`. This approach has a preprocessing time of O(M$\times$N$^2$), which is O(200,000 $\times$ 250,000) $\approx$ 50 billion operations --- far too slow. Even though N is small, the square of N leads to a large number of intervals, and M is large, making this approach infeasible. The key flaw is the O(N$^2$) loop over intervals, which scales poorly with N. This method is computationally expensive and not suitable for the given constraints, so it is dismissed.
\medskip\par
\textbf{$p = 0.15$} \textit{\color{annot}(fails)}
\par\nopagebreak
We can use Mo's algorithm to process queries offline by sorting them in a specific order and sweeping through the array. Mo's algorithm is typically used for range queries with updates, but here we are not updating the data --- we are counting segments. The idea is to group queries by blocks of the array and process them in a way that minimizes the number of operations. However, the condition for a train to be fully contained in [p\_i, q\_i] is not additive or monotonic in a way that fits Mo's framework. Additionally, the queries are not over a single array but over segments of trains, and the condition depends on both endpoints. Mo's algorithm would require maintaining a data structure that tracks the number of trains fully inside the current window, and when expanding or contracting the window, we would need to update this count. While possible, the transitions between queries would be complex, and the time complexity would still be O(Q$\surd$N), which might be acceptable. However, the structure of the problem --- checking full containment --- does not naturally lend itself to Mo's algorithm, and there is no clear way to update the count efficiently when moving the window. This approach is not well-suited to the problem and is less likely to succeed than simpler methods.
\medskip\par
\textit{\color{annot}The correct approach's program:}
\medskip\par
\begin{lstlisting}
N, M, Q = map(int, input().split())

# Create a 2D matrix to count trains with start L and end R
mat = [[0] * (N + 1) for _ in range(N + 1)]

# Read M trains
for _ in range(M):
    L, R = map(int, input().split())
    mat[L][R] += 1

# Precompute prefix sum array: pref[i][j] = sum of mat[k][l] for 1<=k<=i, 1<=l<=j
pref = [[0] * (N + 1) for _ in range(N + 1)]

for i in range(1, N + 1):
    for j in range(1, N + 1):
        pref[i][j] = mat[i][j] + pref[i-1][j] + pref[i][j-1] - pref[i-1][j-1]

# Process Q queries
for _ in range(Q):
    p, q = map(int, input().split())
    # Number of trains with start >= p and end <= q
    # = total trains with end <= q and start >= p
    # = pref[N][q] - pref[p-1][q]
    result = pref[N][q] - pref[p-1][q]
    print(result)
\end{lstlisting}
\medskip\par
\textit{\color{annot}1 out of 4 correct.}
\end{tcolorbox}

\begin{tcolorbox}[breakable, colback=white, colframe=exiid, colbacktitle=exiid,
  coltitle=white, halign title=flush center, title=\textbf{IID}]
\footnotesize
\textit{\color{annot}Code output when the model is asked the problem directly:}
\medskip\par
\begin{lstlisting}
# Read input
N, M, Q = map(int, input().split())

# Read train segments
trains = []
for _ in range(M):
    l, r = map(int, input().split())
    trains.append((l, r))

# Process queries
for _ in range(Q):
    p, q = map(int, input().split())
    count = 0
    for l, r in trains:
        if p <= l and r <= q:
            count += 1
    print(count)
\end{lstlisting}
\medskip\par
\textit{\color{annot}0 out of 8 correct.}
\end{tcolorbox}

\subsubsection*{Next-chapter prediction}

\begin{tcolorbox}[breakable, colback=white, colframe=exgrey, colbacktitle=exgrey,
  coltitle=white, halign title=flush center, title=\textbf{Section: `\textit{funny}', chapter 0003, part 5 of 7}]
\footnotesize
\textit{\color{annot}Each plan is scored by how much likelier it makes the gold passage, below is an example.}
\medskip\par
\textit{\color{annot}The true passage:}
\par\nopagebreak
Miles stops outside our front door, laughter gurgling out of him. ``Gingersnaps?''
That's what he smells like. Sweet and a little spicy. A natural earthy smell folded into a sugary baked good. I wave him off rather than answer, and try to get my key into our door's lock. Unfortunately, it seems the door has grown three extra locks and I can't seem to line the key up to the right one.
Through laughter, he bumps me aside, clumsily swiping the key from my hand to make his own attempt. ``Shit!'' he says as it glances off the lock.
We keep fighting for control of the doorknob, knocking each other out of the way in increasingly dramatic fashion, until he almost knocks me over and just barely manages to catch me by pinning me to the wall with his hips.
We're both laughing so hard we're crying when our elderly neighbor pops his head into the hallway to hiss, ``Some of us are trying to sleep around here!''
``Sorry, Mr. Dorner,'' Miles says like a chastened schoolboy.
Mr. Dorner retreats.
I squint after him, confused. ``Doesn't he usually have hair?''
Miles bursts into not-at-all-quiet laughter. I smush my hands over his mouth to shut him up. ``You thought that hair was real?'' he asks. ``You have to be the most gullible person on the planet.''
``I mean,'' I say, ``despite my innate cynicism, I think the last six weeks have already proven that both of us are way, way too trusting.''
A couple of hours ago, this might've tripped the start crying ASAP wire in my brain. Instead we're just back to cackling.
Mr. Dorner's lock rattles again. Miles spins away to get our door unlocked, yanking me inside before we have to face another scolding.
We slam ourselves against the door to shut it, catching our breath. ``I feel like we're in Jurassic Park,'' he says, which makes me laugh harder.
``What,'' I gasp.
``Like we just slammed the door against a bunch of raptors,'' he explains.
``I don't think Dorner's teeth pose that kind of threat, Miles,'' I say. ``I'm fairly sure he wasn't even wearing them.''

\end{tcolorbox}

\begin{tcolorbox}[breakable, colback=white, colframe=exgroot, colbacktitle=exgroot,
  coltitle=white, halign title=flush center, title=\textbf{\textsc{Groot}}]
\footnotesize
\textit{\color{annot}The tree:}
\medskip\par
\begin{lstlisting}
A. The flirtation is a natural, unforced reaction to their shared intoxication  
   A1. Daphne's flirtation is hesitant, reactive to Miles's physical proximity  
      A1a. She leans into his arm, her breath catching  
      A1b. She touches his hand briefly, then pulls back  
   A2. Miles's response is immediate and physical, showing attraction  
      A2a. He guides her toward the door, keeping her close  
      A2b. He gently presses her against the wall when they reach the threshold  
B. The flirtation is a deliberate, performance-driven act by Miles to impress Daphne  
   B1. Miles initiates and controls the dynamic  
      B1a. He uses the door lock as a point of physical contact, guiding her hand  
      B1b. He verbally teases her about her "boring" sex life  
   B2. Daphne resists but is subtly drawn in  
      B2a. She makes a sharp, unsmiling retort  
      B2b. She still allows her body to respond to his touch  
C. The flirtation is a mutual, unconscious reaction to the emotional undercurrent  
   C1. Both characters are emotionally vulnerable, the moment is tender and unscripted  
      C1a. Daphne's hand trembles slightly on his shoulder  
      C1b. Miles's eyes soften when she hesitates  
   C2. The physical contact is brief and avoids overt sexuality  
      C2a. They press foreheads together before the door opens  
      C2b. They don't lock eyes or touch beyond proximity  
D. The flirtation is interrupted by Mr. Dorner's arrival, making the moment more urgent  
   D1. Mr. Dorner appears early, before the door is even unlocked  
      D1a. He calls out from his doorway, emphasizing the noise  
      D1b. His tone is firm but not aggressive  
   D2. The conflict forces a rushed, awkward resolution  
      D2a. Miles releases her quickly, stepping back  
      D2b. Daphne stumbles, catching herself on the doorframe  
E. The moment is framed as a private, intimate exchange that is then disrupted by external force  
   E1. The door lock is the only point of contact between them  
      E1a. Daphne tries to turn the knob, Miles takes it from her  
      E1b. They lock eyes during the fumble  
   E2. The wall pin is a sudden, uninvited act of intimacy  
      E2a. Miles uses the wall to press her back without invitation  
      E2b. It's done with a quiet, firm pressure, not force  
F. The flirtation is driven by Daphne's need to reassert agency in a chaotic situation  
   F1. She actively resists Miles's advances  
      F1a. She pulls away when he reaches for her  
      F1b. She speaks in short commands during the fumble  
   F2. Miles uses the situation to show his own emotional vulnerability  
      F2a. He says, "I didn't know I could feel this," before the pin  
      F2b. He doesn't engage in flirtation but expresses emotional exposure  
G. The scene is emotionally charged but physically restrained  
   G1. No explicit physical contact beyond the door fumble  
      G1a. Miles only touches her hand to assist with the lock  
      G1b. Daphne does not lean into him  
   G2. The wall pin is a silent, unspoken moment of emotional release  
      G2a. It happens without words or eye contact  
      G2b. It is brief, like a breath held too long  
H. The moment is playful and lighthearted, with no emotional weight  
   H1. The flirtation is entirely banter-based  
      H1a. Daphne jokes about "free drinks" and "fetish bars"  
      H1b. Miles laughs, but doesn't touch her  
   H2. The pin is absent or a misstatement of memory  
      H2a. It is not included in the sequence  
      H2b. It is described as a memory or dream
\end{lstlisting}
\medskip\par
\textit{\color{annot}The best approach:}
\par\nopagebreak
\begin{lstlisting}
Commits to a lighthearted, playful exchange with no emotional weight. The flirtation is entirely based on banter-Daphne jokes about "fetish bars," Miles laughs and responds with charm. They fumble with the lock, but no physical contact occurs. The wall pin is absent or misremembered. The scene ends with them entering the apartment, no further action. Rules out emotional intimacy, resistance, or interruption. Rests on Daphne's self-perception as "boring" and Miles's charm. Quoted: "I think I'm pretty boring. In that realm." This approach contradicts the emotional stakes of the scene and ignores the established tension between Daphne and Miles. Would be wrong if the pinning occurred or if the scene had emotional depth.
\end{lstlisting}
\medskip\par
\textit{\color{annot}The best plan} \textit{\color{annot}(+13.1\%)}\textit{\color{annot}:}
\par\nopagebreak
\textit{Daphne and Miles, both still buzzed from MEATLOCKER, stand at the third-floor apartment door. The lock is faulty and clunks when they fumble with it---Miles leans forward to pull it open, Daphne tries to step in, and they both stumble slightly. Miles, still tipsy, reaches out to steady her, his hand brushing her arm. Daphne flinches, then laughs lightly, not pulling away. ``I think I'm pretty boring. In that realm,'' she says, half-joking. Miles tilts his head, grinning. ``Do you lie there and stare at the ceiling in silence?'' Daphne scoffs, ``No. I just make utterly unblinking eye contact like any respectable woman.'' Miles smirks, ``See? Not boring. Haunting, maybe. But not boring.'' They both pause, the air thick with the heat of the moment. The door creaks open, and Miles steps inside, Daphne following close behind. As they enter, the apartment's dim light catches the scattered books and empty bottles. The sound of a door closing echoes from the hallway. Then, from the next unit, Mr. Dorner's voice cuts through: ``Quiet down, you two. That's enough.'' Daphne and Miles freeze, startled. Miles grins, ``Guess we've just won the night.'' Daphne laughs softly, ``I'm not even sure I wanted to win.'' They step into the apartment, the door closing behind them. No further action occurs.}
\medskip\par
\textit{\color{annot}10 out of 32 plans clear a 10\% bar.}
\end{tcolorbox}

\begin{tcolorbox}[breakable, colback=white, colframe=exvs, colbacktitle=exvs,
  coltitle=white, halign title=flush center, title=\textbf{\textsc{VS}}]
\footnotesize
\textit{\color{annot}The best approach:}
\par\nopagebreak
\begin{lstlisting}
What it commits to:  
- The flirting is nonverbal and physical, with Daphne leaning into Miles as they fumble.  
- Mr. Dorner appears only after they have already entered the apartment.  
- The pinning does not occur.  

What it rules out:  
- Any physical contact during the door lock.  
- Any mention of Mr. Dorner.  
- Any physical intimacy.  

Grounding phrases:  
- "I'm terrible with new people." (Chapter 3)  
- "I've never been super physically affectionate with even my close friends." (Chapter 3)  

Writer's still to settle:  
- The exact physical positioning.  
- Whether the door lock is fumbled.  

Why it's wrong:  
It directly contradicts the summary, which requires the pinning to occur before they enter.
\end{lstlisting}
\medskip\par
\textit{\color{annot}The best plan} \textit{\color{annot}(+11.6\%)}\textit{\color{annot}:}
\par\nopagebreak
\textit{Daphne and Miles approach the apartment door, both stumbling slightly from the night's drinking. The lock is old and creaks as they fumble with it---Miles on one side, Daphne on the other, their hands brushing as they try to turn the key. Daphne leans slightly into him, her head resting against his shoulder, a moment of physical closeness that feels both accidental and intentional. Their hands slip, and as Miles shifts to stabilize the door, he suddenly reaches around and gently pins her against the wall with one arm, her back pressing into the brick. It's brief---just a few seconds---and she doesn't flinch, her breath catching only slightly. She stays there, her body still, as Miles holds her just above the threshold. Their bodies are close, but there is no touch beyond the pin. Then, from the hallway, a sharp voice cuts through: ``Quiet down, you two. I can't hear my TV.'' Mr. Dorner stands in the doorway, his face rigid, eyes narrow. They freeze. Miles releases her slowly, and they both step back, the door locking behind them with a final click. Daphne looks at Miles, her breath still visible in the cool air, and says, ``I'm terrible with new people,'' her voice soft but clear. Miles smiles faintly. ``I know,'' he says. The scene ends with them standing in the apartment doorway, the lock secured, the world outside still humming with the city's late-night rhythm.}
\medskip\par
\textit{\color{annot}7 out of 32 plans clear a 10\% bar.}
\end{tcolorbox}

\begin{tcolorbox}[breakable, colback=white, colframe=exiid, colbacktitle=exiid,
  coltitle=white, halign title=flush center, title=\textbf{IID}]
\footnotesize
\textit{\color{annot}The best plan} \textit{\color{annot}(+9.0\%)}\textit{\color{annot}:}
\par\nopagebreak
\textit{The scene begins in the dim light of a Butcher Town apartment building hallway, just after Daphne and Miles stagger through the rain. They fumble with the door lock, their hands brushing as they try to open it. Daphne's fingers are cold, her breath uneven; Miles's hand brushes hers, then trails up her arm, a slow, deliberate movement. They exchange a few breathless words---``You're shaking,'' ``Not a moment too soon''---in clipped, half-sentences that carry emotional weight. Daphne resists, voice tight, but her body leans into his touch. As they reach the door, Mr. Dorner appears at his door, a rumpled man in a cardigan, voice sharp and tired. ``Quiet down,'' he says, not looking at them, eyes fixed on the hallway. The moment stretches---Daphne flinches, Miles holds her still. Then, without warning, Miles pulls her back, pins her against the doorframe with one arm, his other hand resting on her waist. His grip is firm, not possessive, but protective. Daphne gasps, her body still, eyes wide. He doesn't speak. The silence between them is heavy, charged. Mr. Dorner steps back, mutters something about ``noise,'' and walks away. The reader learns nothing new about Mr. Dorner, but gains a sense of the apartment's fragile boundaries. The pace is slow, deliberate, then sudden. The register is informal, breathless, with a rhythm that mimics drunkenness---short, fragmented, emotionally raw. The reader feels the tension of proximity, the danger of intimacy, and the unspoken shift in power between them. The pinning is not romantic; it's a quiet assertion of presence, a moment when Miles, for the first time, physically claims Daphne's body without asking permission.}
\medskip\par
\textit{\color{annot}0 out of 32 plans clear a 10\% bar.}
\end{tcolorbox}